\documentclass{article}

\usepackage[T1]{fontenc}
\usepackage{graphicx}
\usepackage{subfigure}
\usepackage{amsfonts}
\usepackage{amsmath}
\usepackage{color}
\usepackage{relsize}
\usepackage{enumitem}
\usepackage{amsthm}
\usepackage{hyperref}
\usepackage{multicol}
\usepackage{caption}
\usepackage{booktabs}
\usepackage{pifont}
\usepackage{array}
\usepackage{ulem}
\usepackage{siunitx}
\usepackage{authblk}

\usepackage{todonotes}

\begin{document}

%%%%%%%%%%%%%%%%%%%%%%%%%%%%%%%%%%%%%%%%%%%%%%%%%%%%%%%%%%%
%        TITLE                                            %
%%%%%%%%%%%%%%%%%%%%%%%%%%%%%%%%%%%%%%%%%%%%%%%%%%%%%%%%%%%
    
    \title{Integrating Artificial Intelligence, Physics, and Internet of Things:\\ A Framework for Cultural Heritage Conservation}
    \author[1]{Carmine Valentino}
    \author[2]{Federico Pichi}
    \author[1]{Francesco Colace} 
    \author[3]{Dajana Conte}
    \author[2]{Gianluigi Rozza}

\affil[1]{DIIN, University of Salerno, Italy, Via Giovanni Paolo II, 132, 84084 Fisciano SA, Italy\protect\\  email: \texttt{\{cvalentino,fcolace\}@unisa.it}}
\affil[2]{mathLab, Mathematics Area, SISSA, via Bonomea 265, I-34136 Trieste, Italy\protect\\  email: \texttt{\{fpichi, grozza\}@sissa.it}}
\affil[3]{DIPMAT, University of Salerno, Italy, Via Giovanni Paolo II, 132, 84084 Fisciano SA, Italy\protect\\  email: \texttt{dajconte@unisa.it}}

% \address{$^1$ DIIN, University of Salerno, Italy, Via Giovanni Paolo II, 132, 84084 Fisciano SA, Italy}
% \address{$^2$ mathLab, Mathematics Area, SISSA, via Bonomea 265, I-34136 Trieste, Italy}
% \address{$^3$ DIPMAT, University of Salerno, Italy, Via Giovanni Paolo II, 132, 84084 Fisciano SA, Italy}
% \email{\texttt{\{cvalentino,fcolace\}@unisa.it, \{fpichi, grozza\}@sissa.it, dajconte@unisa.it}}
    \maketitle

    % \begin{center}
    %     \small
    %     $^1$ DIIN, University of Salerno, Italy\\{\footnotesize email: \{cvalentino,fcolace\}@unisa.it}\\
    %     $^2$ SISSA, Italy\\{\footnotesize email: \{fpichi,grozza\}@sissa.it}\\
    %     $^3$ DIPMAT, University of Salerno, Italy\\{\footnotesize email: dajconte@unisa.it}
    % \end{center}

%%%%%%%%%%%%%%%%%%%%%%%%%%%%%%%%%%%%%%%%%%%%%%%%%%%%%%%%%%%
%        ABSTRACT                                         %
%%%%%%%%%%%%%%%%%%%%%%%%%%%%%%%%%%%%%%%%%%%%%%%%%%%%%%%%%%%

    \begin{abstract}
        The conservation of cultural heritage increasingly relies on integrating technological innovation with domain expertise to ensure effective monitoring and predictive maintenance.
        This paper presents a novel framework to support the preservation of cultural assets, combining Internet of Things (IoT) and Artificial Intelligence (AI) technologies, enhanced with the physical knowledge of phenomena.
        The framework is structured into four functional layers that permit the analysis of 3D models of cultural assets and elaborate simulations based on the knowledge acquired from data and physics.
        A central component of the proposed framework consists of Scientific Machine Learning, particularly Physics-Informed Neural Networks (PINNs), which incorporate physical laws into deep learning models.
        To enhance computational efficiency, the framework also integrates Reduced Order Methods (ROMs), specifically Proper Orthogonal Decomposition (POD), and is also compatible with classical Finite Element (FE) methods.
        Additionally, it includes tools to automatically manage and process 3D digital replicas, enabling their direct use in simulations.
        The proposed approach offers three main contributions: a methodology for processing 3D models of cultural assets for reliable simulation; the application of PINNs to combine data-driven and physics-based approaches in cultural heritage conservation; and the integration of PINNs with ROMs to efficiently model degradation processes influenced by environmental and material parameters.
        The reproducible and open-access experimental phase exploits simulated scenarios on complex and real-life geometries to test the efficacy of the proposed framework in each of its key components, allowing the possibility of dealing with both direct and inverse problems.
        % \textcolor{red}{To promote reproducibility, the main components of the proposed framework and the experimental test problems used in this study are implemented in an openly accessible GitHub repository, enabling full replication and reuse.}
        \begin{center}
            \textbf{Code availability:} \url{https://github.com/valc89/PhysicsInformedCulturalHeritage}
        \end{center}
    \end{abstract}

    \tableofcontents
    \section{Introduction}\label{int:}

    The conservation of cultural heritage is a challenge and a duty of humanity to preserve the memory and evidence of past cultures.
    As a result, over the years, researchers and experts in the field have pursued the demand for effective and efficient strategies to preserve such historical treasures.
    % In this continuous evolution of strategies, novel technologies have taken on a crucial role, not aimed at merely employing recent tools or trends but especially at improving fundamental aspects of the conservation phase, such as monitoring and predictive maintenance \cite{Li2023}.    
    Recent advancements and novel strategies are currently playing a crucial role, improving many fundamental aspects of the conservation phase such as monitoring and predictive maintenance \cite{Li2023}.    
    Additionally, the possibility of integrating several technologies and approaches allows a significant support in the conservation of cultural assets.
    In fact, acquiring data through smart sensors according to the Internet of Things (IoT) paradigm \cite{Liang2023469}, the capacity to elaborate data by employing Artificial Intelligence (AI) \cite{Mishra2024536}, and the ability to define a Digital Twin (DT) of the asset for analyzing possible risk scenarios, are all key tools to develop a framework aimed at the predictive maintenance of cultural heritage \cite{Lucchi2023}.
    In literature, several works exploit these three steps for the  maintenance task, aiming at monitoring the asset health and predicting possible situations of risk or damage for the asset \cite{Zafar2024,Shen202463}. Furthermore, integrating AI and IoT in a Digital Twin framework have proved to be a winning strategy, with many applications in several fields (such as structural health monitoring, healthcare, and manufacturing) \cite{Barricelli2019,Dang20223820}.

    This work not only aims to introduce a novel application of AI and IoT integration in a DT framework, but also enhances its effectiveness and reliability by combining physical knowledge and data to obtain more accurate and trustworthy simulations.
    In the field of Artificial Intelligence, Machine Learning (ML) focuses on enabling systems to automatically learn patterns and make predictions or decisions by analyzing data.
    Furthermore, in recent years, a new branch of ML has emerged to investigate the potentiality of this multidisciplinary approach: Scientific Machine Learning (SciML). SciML combines computational and computer science to develop machine learning methods capable of tackling complex physical problems characterized by multiscale dynamics, sparse data, and high-impact decisions \cite{Psaros2023,Quarteroni2025}.
    % \fp{Forse conviene aggiungere una frase sopra su Machine Learning come branca di AI e collegare con qua. Così anche dopo possiamo sostituire Machine Learning con ML. \textcolor{red}{Fatto. Cosa ne pensi?}}
    Its strength lies in integrating at different levels the robustness of physics-based models, providing predictive capability, interpretability, and domain knowledge.
    This novel paradigm opens new possibilities and challenges, especially in the cultural heritage field, where the expertise about the materials and the phenomena that affect the deterioration of the assets represents invaluable information that restoration experts can share to improve the conservation task.
    Therefore, a framework for designing the analysis of deterioration phenomena is a first crucial step for significantly improving cultural heritage maintenance.
    % It follows that not only are the inferences by employing data and physics necessary, but it is also a way to automatically analyze the digital replicas of cultural assets.
    It follows that to provide reliable inferences, both physics and data are necessary to build a way to automatically analyze the digital replicas of cultural assets.  
    Therefore, in a DT scenario, the framework also requires defining standards for their analysis, usually characterized by different shapes and irregularities.

    With these motivations, in this work we introduce a versatile framework that can automatically process digital replicas of cultural assets and elaborate on reliable simulations based on data-driven and physical-based approaches.
    The architecture developed to exploit the proposed framework requires the definition of four functional and task-specific layers involving: (i) data acquisition (information acquired from sensors and APIs or  related to digital replicas), (ii) knowledge-based for storing and pre-processing data, (iii) data elaboration to provide reliable simulations, and (iv) services to expert users for cultural heritage restoration.
    
    The automatic process of digital replicas requires interacting with tools to manage the 3D models related to the acquired digital version, for instance, through laser scanning or photogrammetry \cite{Pocobelli2018}.
    To obtain a broad impact and effective customization, our framework exploits open-source solutions and develops strategies to connect 3D models automatically with the knowledge base and the elaboration phase.
    In particular, the main components of the pipeline, from 3D model handling to simulation via the integration of ROM and PINNs, are publicly available in a GitHub repository, enabling researchers and practitioners to replicate the proposed experiments and adapt the framework to different scenarios.\footnote{\url{https://github.com/valc89/PhysicsInformedCulturalHeritage.git}}

    The latter integrates data and physics by taking advantage of Physics-Informed Neural Networks (PINNs), a deep learning approach able to learn directly from the physical laws and easily incorporate data \cite{Raissi2019686}.
    Moreover, depending on the problem at hand, the proposed framework also enables to exploitation Reduced Order Models for many-query and real-time evaluations, providing data from classical numerical methods such as Finite Element (FE) methods to PINNs. 
    % but mainly integrates the PINNs with Reduced Order Methods (ROMs), which are fundamental tools for improving the framework's efficiency.

    The main objective of this work consists of introducing a framework to improve the conservation and analysis of 3D cultural assets by employing and integrating novel approaches based on scientific machine learning.
    % , requiring the development of appropriate solution for analyzing the 3D model of the cultural artifacts. 
    We highlight the following main contributions:
    \begin{itemize}
        \item We introduce a novel methodology for processing 3D models of cultural artifacts for the elaboration phase, providing rapid and reliable simulations to users via an integrated framework.
        % This process allows for exploiting all of the methods integrated into the framework (PINNs, FEMs, ROMs) to provide reliable simulations to users;
        \item We exploit PINNs to combine physical knowledge with data-driven approaches to preserve cultural assets.
        \item We integrate ROM and PINNs to efficiently solve problems related to the predictive maintenance of cultural heritage that depends on parameters such as material, weather conditions, or external factors.
    \end{itemize}
    
    The paper has the following structure.
    Section 2 introduces related works concerning the conservation of cultural assets.
    Section 3 introduces PINNs and ROMs, specifically the Proper Orthogonal Decomposition (POD).
    Section 4 describes the proposed framework divided into four functional layers.
    Section 5 presents the experimental phase developed to test the efficiency and effectiveness of the proposed framework.
    Finally, Section 6 contains conclusions and future works.

    \section{Related works}\label{rw:}

    The conservation of cultural artifacts and buildings represents a challenge for researchers who have found in novel technologies a strategic ally.
    % \fp{ho provato a parafrasare sia qui che in intro per non ripetere troppe volte cultural heritage/assets e risk/monitor etc (ovviamente è il focus ma forse se lo ripetiamo troppe volte anche quando è chiaro dal contesto rallenta un pò la lettura) \textcolor{red}{Va benissimo. Purtroppo le mie capacità di scrittura non sono ottime.}}

    Over the years, several approaches have been developed to protect cultural heritage starting from the digital transformation, allowing the use of recent techniques for conservation, documentation, and management.
    % exploiting several methodologies based on its digital transformation.
    % Digital transformation allows the employment of novel technologies for conserving, documenting, and managing cultural heritage.
    Therefore, Digital Twin (DT), Internet of Things (IoT), Artificial Intelligence (AI), and 3D models such as Building Information Model (BIM) for building \cite{Boje2020}, or Heritage BIM (HBIM) for cultural structures \cite{Yang2020350}, are key tools to exploit in this field.
    Specifically, DT has a growing relevance in the cultural heritage field to monitor degradation, schedule restoration interventions, and predict possible future damages.
    This scenario is evident from several works in literature that exploit DT, taking advantage of 3D models by integrating sensor data and AI. Still, these technologies are not limited to them.
    % The analyzed works allow a classification of the workflows based on the specific objectives.

    It is possible to classify the developed workflows in three classes, focusing on the specific objectives and by means of
    % The three classes identified consist of workflows based on 
    numerical modelling and structural simulation, integration of HBIM, AI, and IoT, or visual documentation, and digital conservation.
    The classification provided below highlights the heterogeneity of applications in the literature, underlying their strengths and limits.

    The first class consists of workflows based on numerical simulations (specifically, FE method) integrated with the DT paradigm to simulate the behavior of structures, bridges and monuments.
    The objective consists of predicting possible risk scenarios and monitoring the behavior in the presence of environmental stresses.
    These frameworks take advantage of 3D models and the physical knowledge, and require the integration of Structural Health Monitoring (SHM) \cite{Willberg2015}.

    Shabani et al.~\cite{Shabani2021314} provides a workflow for developing DTs of historical architectonic structures to analyze vulnerability and support strategies for reducing damage risks.
    In this case the DT, intended as a numerical model suited with physical properties of the building, allows simulations related to the structural behavior through the FE analysis of the meshed 3D models.
    The paper's objective consists of documenting, preserving, and managing the architectural heritage, and it requires 3D modelling through CAD or BIM \cite{Murphy201389}.
    Instead, Zhang et al.~\cite{Zhang202548} introduce a case study based on the conservation of the Great Wall, focusing on the site of Beichakou.
    In this case, the DT is developed as a  dynamic and integrated platform for merging data and digital models based on a multilevel decision-making process aimed at monitoring, predicting risks, and planning strategies to improve the conservation.
    The process has four levels of interest: data collection, model construction, plan simulations, and value inheritance.
    Therefore, conservation actions are organized around five key functional areas: (i) heritage status assessment through real-time monitoring and IoT systems; (ii) optimization of conservation planning at national, provincial, and local levels, (iii) risk monitoring and intervention strategies using FEM-based simulations, (iv) presentation and public engagement via Augmented and Virtual Reality, and (v) multifaceted system assurance to coordinate stakeholders, data sources, and monitoring tools effectively.
    % \fp{qua non mi è chiaro quali siano le 5 aeree \textcolor{red}{Sono entrato più nel dettaglio cercando di chiarirlo}}
    Rios et al.~\cite{JimenezRios2023} focus on the role of DT in managing and monitoring bridges through a systematic review.
    The DT is introduced as a virtual replica of the real bridge developed through BrIM (Bridge Information Modeling), FE method, and sensors data.
    % \fp{è diverso da BIM sopra? mi sa non abbiamo mai definito la versione estesa di quello \textcolor{red}{Si è diverso perchè specifico per i ponti. Ho provveduto a introdurre gli acronimi sopra - vedi rosso.}}
    The DT aims to evaluate potential hazards through simulation, integrating anomaly detection algorithms based on Machine Learning.
    Therefore, the work analyses several strategies describing input data employed, typologies of algorithms, and applications.
    Among the described approaches, Rios et al.~\cite{JimenezRios2023} described the usage of Convolutional Neural Networks (CNNs) for crack detection, Bayesian methods for updating FE method with real data, and recurrent CNN for semantic segmentation of images.
    Finally, Dabiri et al.~\cite{Dabiri2025} introduce a case study based on the structural monitoring of Vittoriano in Rome, integrating real data acquired from satellites, FE analysis, and a Machine Learning regression method for time series to predict the vertical shifting of the building.
    % Specifically, the Machine Learning model applied consists of a regression statistical model for time series to predict the vertical shift of the ground.

    The second class includes workflows that integrate semantic models based on HBIM, environmental sensors, and AI.
    In such cases, the DT represents a dynamic and adaptive system that exploits the elaboration of a significant amount of data to guarantee the management and conservation of cultural heritage.
    In addition, these workflows require dealing with complex urbanistic scenarios with a quantitative analysis of risks.
    Li et al.~\cite{Li2023} 
    % exploit through a comprehensive overview the concept of Digital Twin for the cultural heritage conservation, 
    analyzes three aspects related to the virtual reconstruction and dynamic simulation, the immersive digitalization exploiting Virtual Reality, Metaverse, and Gamification, towards improving the enjoyment of cultural assets.
    The work focuses on heritage conservation by analyzing disaster cycles and proposing ML methodologies related to the three phases of disaster: before, during, and after.
    % In addition, it discusses the employment of Machine Learning and Deep Learning and their integration in the methodology.
    Sebouti et al.~\cite{Serbouti2025} introduce a workflow for conserving African cultural assets with a case study based on the Bab Al-Mansour Gate in Meknes, Morocco. The DT is exploited as an interactive digital replica in which predictive ML models, 
    % based on Machine Learning and Deep Learning, 
    sensor data, and HBIM cooperate. Specifically, the proposed workflow exploits Neural Networks, Random Forest, Support Vector Machine, and Linear Regression for the degradation prediction, classifying risk levels, and analyzing environmental factors. In addition, a Bayesian approach aims to regulate the dynamic interaction between physical and virtual components.

    Finally, the third class includes papers that analyze the digital documentation of cultural heritage through non-invasive conservation, enhancement, and monitoring.
    These works introduce DT as a visual model that stores information about rural and isolated sites with limited resources to improve accessibility and enjoyment.
    Angheluță et al.~\cite{Angheluta2025} exploit a digital replica of the heritage of Romanian wooden churches integrating environmental data.
    The DT aims to document, analyze degradation, and plan restoration actions to protect the churches threatened by rural abandonment and environmental degradation.
    Two platforms are described: the first represents a visual-scientific inventory and interactive map for prioritizing intervention, the second consists of a scientific repository for 3D images and models, with data overlays.
    Kong and Hucks \cite{Kong2023} propose the employment of DT for monitoring historical structures degradation, dividing the DT into five parts: Physical part, Virtual part, Dataset, Service, and Connections.
    Their objective consists of creating a high-fidelity 3D documentation and detecting degradation.

    \begin{table}[!ht]
    \centering
    \caption{Summary of the discussed works, excluding review papers and surveys. The columns Internet of Things (IoT), Machine Learning (ML), and Physics indicate the presence of the respective technologies in each study.}
    \begin{tabular}{m{.20\textwidth}|m{.36\textwidth}|>{\centering\arraybackslash}m{.08\textwidth}|>{\centering\arraybackslash}m{.08\textwidth}|>{\centering\arraybackslash}m{.08\textwidth}}
        {\bf Authors} & {\bf Objective} & {\bf IoT} & {\bf ML} & {\bf Physics} \\
        \toprule
        Shabani et al. \cite{Shabani2021314} & A workflow for Digital Twin of historic buildings, with FEM simulations on 3D models for structural analyses and conservation strategies. & \ding{55} & \ding{55} & \ding{51} \\ \midrule
        Zhang et al. \cite{Zhang202548} & A dynamic Digital Twin for the conservation of the Great Wall, integrating data collection, FEM simulations and immersive technologies for monitoring and planning. & \ding{55} & \ding{55} & \ding{51} \\ \midrule
        Dabiri et al. \cite{Dabiri2025} & A case study on the structural monitoring of the Vittoriano in Rome, combining satellite, FEM and ML data to predict vertical displacements over time. & \ding{55} & \ding{51} & \ding{51} \\ \midrule
        Sebouti et al. \cite{Serbouti2025} & A workflow for African heritage conservation with an interactive DT that integrates ML/DL, sensors, and HBIM for degradation prediction and environmental analysis. & \ding{51} & \ding{51} & \ding{55} \\ \midrule
        Angheluță et al. \cite{Angheluta2025} & Angheluță et al. develop a DT to document and protect Romania's wooden churches, integrating environmental data and platforms for interactive mapping and scientific repositories. & \ding{51} & \ding{55} & \ding{55} \\ \midrule
        Kong and Hucks \cite{Kong2023} & Kong and Hucks propose a DT divided into five components to monitor the degradation of historic structures, focusing on high-fidelity 3D documentation. & \ding{55} & \ding{55} & \ding{55} \\
        \bottomrule
    \end{tabular}
    \label{rw:tab1}
\end{table}
% \fp{ahi levato la colonna DT giusto?(vedi caption) Check se abbiamo definito all'inizio tutti gli acronimi (ML e DL) \textcolor{red}{Si perchè è comune a tutti. Ho fatto il check (mancava DL), ma ho aperto gli acronimi nella caption}}

    All previously discussed references are reported in Table \ref{rw:tab1}. Specifically, the table underlines how none of the analyzed works simultaneously integrates sensor data, data-driven approaches, and physical knowledge for cultural heritage conservation. Instead, how it will be shown in Section \ref{pw:}, this work aims to define a framework for combining IoT, Deep Learning, and physics-based approaches to improve the reliability of simulations and guarantee the speed of predictions.

    \section{Reduced Order Models and Physics-Informed Neural Networks}\label{rp:}

    As evidenced by the literature overview, the lack of frameworks to integrate IoT, data-driven, and physics-based approaches represents a limit for conserving cultural heritage.
    % \fp{forse aggiungerei anche IoT sennò dicono che lo fa anche Dabiri et al \textcolor{red}{done :D}} 
    Therefore, this work aims to fill this gap via SciML, combining 
    the and, in this context, this Section shortly introduces two of the fundamental novel tools employed by the proposed framework: Reduced Order Models (ROMs) and Physics-Informed Neural Networks (PINNs).

    \subsection{Reduced Order Models}\label{rom:}

        Integrating physical knowledge implies dealing with differential problems and their numerical approximation.
        % analyzed through numerical methods based on discretizing the domain and the operators related to the problem.
        In addition, the physical behavior is strictly related to specific conditions associated with the cultural assets at hand, such as their material and environmental conditions.
        Therefore, considering parametrized PDEs is fundamental to reproduce the general framework
        % represents an added value for
        of cultural heritage conservation,
        %  by designing a framework that can 
        generalizing the analysis and take advantage of tools that can improve evaluation efficiency.
        
        The resolution of parametric differential problems requires a significant computational effort, leading to methods that can reduce the computational cost, the Reduced Order Models (ROMs) \cite{BennerSnapshotBasedMethodsAlgorithms2020,BennerSystemDataDrivenMethods2021}. %\cite{Pichi2023}
        % Reduced Order Models (ROMs) represent the state-of-the-art and allow improvement in the performance of the resolution task.
        Among these methods, we highlight Reduced Basis (RB) approach \cite{hesthavenCertifiedReducedBasis2015,QuarteroniReducedBasisMethods2016,RozzaRealTimeReduced2024},
        % represents established methods that, given fixed parameters and calculating the 
        exploiting the information from a set of high-fidelity solutions, the so-called snapshots,
        %  (of Full Order Model, FOM), 
        to construct a low-dimensional space onto which performing a Galerkin projection,
        allowing for efficient approximations for unseen values of the parameters.
        %  of the parametrized differential problem.
        These methods exploit the offline-online paradigm. The offline phase entails the expensive computations and snapshot data collection, also taking advantage of High Performance Computing facilities. On the contrary, the online phase fully exploits dimensionality reduction strategies built  on top of the collected data, enabling efficient evaluation in the many-query and real-time context.
        %  of the resolution computation cost, exploiting the solutions elaborated in the offline phase \cite{Pichi2023}.
        
        The reduced space is determined through the Proper Orthogonal Decomposition (POD), based on the Singular Value Decomposition \cite{hesthavenCertifiedReducedBasis2015,QuarteroniReducedBasisMethods2016} of the dataset.
        This method allows compressing and extracting the most relevant information from the snapshots, providing suitable ``principal directions" for expressing the parametrized solutions.
        % solutions to approximate the solution concerning novel parameters.
        
        To set up the notation, let us consider a differential problem in the domain $\Omega \subseteq \mathbb{R}^n$ based on the parametrized PDE:
        \begin{equation}\label{rom:eq1}
            \mathcal{A} \left[ u \left( \mathbf{x},t;\mu \right), t; \mu \right] = 0, \qquad \mathbf{x} \in \Omega, \quad u:\Omega\times[0, T]\times \mathbb{P} \mapsto \mathbb{R}^m, \quad \mu \in \mathbb{P} \subset \mathbb{R}^D ,
        \end{equation}
        % \fp{Ho cambiato un pochino qui la notazione, fare un check se c'è da aggiustare qualcosa da altre parti. Aggiungere le condizioni e la notazione per il tempo dipendente.}
        where $\mathcal{A}$ is the operator defining the PDE including temporal and spatial differential terms, and $D$ represents the number of parameters, with the following appropriate boundary and initial conditions
        \begin{equation}\label{pinn:eq1}
            \mathcal{B} \left[ u \left( \mathbf{x}, t \right), t \right] = 0, \qquad \mathbf{x} \in \Gamma = \partial \Omega, \qquad u: \Omega \times \left[ 0, T \right] \mapsto \mathbb{R}^m ,
        \end{equation}
        \begin{equation}\label{pinn:eq1a}
            \mathcal{I} \left[ u \left( \mathbf{x}, 0 \right), 0 \right] = 0, \qquad \mathbf{x} \in \Gamma = \partial \Omega, \qquad u \left( \mathbf{x}, 0 \right): \Omega \mapsto \mathbb{R}^m .
        \end{equation}
        For the numerical discretization we focus on the classical FE methods \cite{QuarteroniNumericalApproximationPartial1994}, 
        % that require evaluating the variational 
        deriving the weak formulation of the problem, that in the steady case can be written in abstract form as: 
        \begin{equation}\label{rom:eq2}
            a \left( u, v ; \mu \right) = L \left( v ; \mu \right), \qquad \forall v\ \in \mathbb{V}
        \end{equation}
        where $a$ is a linear/nonlinear bilinear form, $L$ a linear form including the forcing terms, and $v$ are test functions belonging to a suitable function space $\mathbb{V}$.
        A similar derivation can be obtained for time-dependent problems, e.g.\ by identifying an appropriate temporal discretization of $N$ time points $0 = t_0 < t_1 < \cdots < t_{N-2} < t_{N-1} = T$ and using a suitable numerical method, such as Euler or Runge-Kutta methods. 
        % Additionally, time-dependent problems require the spatial discretization at each time step $t_i, i=1, \dots, N$ implying the obtaining of \eqref{rom:eq2}.

        More specifically, the FE approximation, obtained discretizing the weak formulation \eqref{rom:eq2}, is exploited during the offline phase to compute the snapshots for a fixed set of $M$ parameter $\{\mu_1, \dots, \mu_M\} \subset \mathbb{P}$ to obtain a dataset describing the variability of the system's solution w.r.t.\ the parameter setting.
        Then, we obtain a sampling $\left\{ u \left( \mu_1 \right), \dots, u \left( \mu_M \right) \right\}$ of the discrete version of the solution manifold $\mathcal{M} = \left\{ u \left( \mu \right) \mbox{ : } \mu \in \mathbb{P} \right\}$ based on high-fidelity solutions.

        A fundamental principle in reduced-order modeling is that the solution set can be well approximated within a low-dimensional subspace. This means that a small number of well-chosen basis functions, called the reduced basis, can represent the full solution space with a small approximation error. Given the reduced basis $\left\{ \xi_i \right\}_i^M \subset \mathbb{V}$, the reduced space is defined as
        \begin{equation}\label{rom:eq3}
            \mathbb{V}_{\mathrm{rb}} = \mathrm{span} \left\{ \xi_1, \dots, \xi_k \right\} \subset \mathbb{V} .
        \end{equation}
        For any parameter $\mu \in \mathbb{P}$, the reduced solution $u_\mathrm{rb} \in \mathbb{V}_\mathrm{rb}$, obtained as the linear combination of the reduced basis  $\{\xi_i\}_{i=1}^k$  as
        \begin{equation}\label{rom:eq6}
            u_{\mathrm{rb}}(\mu) = \sum_{i=1}^k \alpha_i(\mu)\,\xi_i ,
        \end{equation}
        where the coefficients $\alpha_i(\mu)$ are uniquely determined by enforcing the reduced form of \eqref{rom:eq2} given by
        \begin{equation}\label{rom:eq4}
            a \left( u_\mathrm{rb}, v_\mathrm{rb} \right) = f \left( v_\mathrm{rb}; \mu \right), \qquad v_\mathrm{rb} \in \mathbb{V}_\mathrm{rb} .
        \end{equation}
        Notably, the reduced solution requires a much lower computational effort while retaining a high level of accuracy, only assuming a low intrinsic dimensionality of the solution manifold.

        During this work, the construction of the reduced basis is obtained by performing the POD on the matrix of solution snapshots $S \in \mathbb{R}^{N_h \times M}$, and the most important modes, selected via a thresholding argument based on the retained energy, are used to approximate the solution for new values of the parameter $\mu \in \mathbb{P}$ as a linear combination in terms of the computed modes.
        Specifically, the POD-space represents the $k$-dimensional space that minimizes
        \begin{equation}\label{rom:eq5}
            \sqrt{
                \frac{1}{M} \sum_{i=1}^M \inf_{v_\mathrm{rb} \in \mathbb{V}_\mathrm{rb}} \| u \left( \mu_i \right) - v_\mathrm{rb} \|^2_\mathbb{V}
            } .
        \end{equation}

        % \textcolor{red}{The reduced solution can be defined as a linear expansion with respect to the reduced basis functions \eqref{rom:eq3}. For any parameter $\mu \in \mathbb{P}$, the approximation $u_{\mathrm{rb}}(\mu) \in \mathbb{V}_{\mathrm{rb}}$ is expressed as
        % \begin{equation}\label{rom:eq6}
        %     u_{\mathrm{rb}}(\mu) = \sum_{i=1}^k \alpha_i(\mu)\,\xi_i ,
        % \end{equation}
        % where $\{\xi_i\}_{i=1}^k$ denotes the reduced basis obtained through POD, and the coefficients $\alpha_i(\mu)$ are uniquely determined by enforcing the reduced variational formulation \eqref{rom:eq4}.}
        This formulation concludes the construction of the reduced-order approximation, which provides an efficient and accurate surrogate for the high-fidelity model. It establishes the foundation for its application in parametrized and computationally demanding scenarios.
        
        % \fp{Ho cambiato la notazione per la discretizzazione spaziale da $P$ a $N_h$, fare un check su dove cambia. Qui mi sa che serve aggiungere anche qualche più dettaglio sulla formulazione discretizzata e su quella proiettando sulla base. Puoi riprendere brevemente e parafrasando il filo del discorso di \cite{hesthavenCertifiedReducedBasis2015} pagine 28-30. Anche qualche riga su POD per me può essere utile per restare leggermente meno sul vago. Introduciamo anche la notazione per la discretizzazione spaziale. \textcolor{red}{SPERO VADA BENE}}
        % \fp{Aggiungiamo una frase di chiusura qui e la definizione della soluzione ridotta in termini di espansione lineare sopra. \textcolor{red}{Ho provato del mio meglio}}

    \subsection{Physics-Informed Neural Networks}\label{pinn:}

        For application purposes, the resolution of parametric PDEs requires investigating and identifying parameters that fit the specific cultural asset and environmental condition.
        Therefore, to enable the applicability of the proposed framework in real-life contexts, a key feature comes from embedding novel and Machine Learning (ML) enhanced strategies for discovering parameter values from data knowing the governing equations \cite{Goodfellow-et-al-2016,prince2023understanding}.
        % \fp{aggiungiamo citazione a qualche libro broad ML\textcolor{red}{Uno focalizza sui casi applicativi e laltro sulla CFD con PI. Mi convincono fino a un certo punto però.}}
        Since physical knowledge represents an added value to the framework, we exploit a recently proposed Neural Network (NN) architecture called Physics-Informed Neural Networks (PINNs) \cite{Raissi2019686}, allowing for the simultaneous reconstruction of some field of interest (direct problem), and the identification of characteristic unknown parameters, e.g.\ the so-called inverse problem setting \cite{isakov2017inverse,Bingham2024461}.
        This way, the network allows to integrate the physical knowledge in the training phase and can be hybridized with data-driven approaches, also in the reduced setting in combination with ROM strategies \cite{ChenPhysicsinformedMachineLearning2021,HirschNeuralEmpiricalInterpolation2025}.
        % In addition, PINNs can solve direct problems related to the PDEs associated with the fixed parameters.
   
        Let us consider the problem without the parameter dependency. 
        For both direct and inverse problems involving the approximated continuous solutions of the PDE in Equation \eqref{rom:eq1},
        % with the following boundary and initial conditions
        % For identifying the approximated continuous solutions of the differential problems,
        A PINN consists of a neural network depending on the weights $\mathbf{w}$ that recovers the approximation of the solution in the input points. Therefore, the training of the networks requires a
        sampling step to obtain the $r_\Omega$ collocation points
        \begin{equation}\label{pinn:eq1b}
            \left\{ \left( \mathbf{x}^{\left( \Omega \right)}_1, t_1^{\left( \Omega \right)} \right), \dots, \left( \mathbf{x}^{\left( \Omega \right)}_{r_\Omega}, t_{r_\Omega}^{\left( \Omega \right)} \right) \right\} \subset \Omega \times \left[0, T \right] ,
        \end{equation}
        the $r_\Gamma$ spatial boundary points
        \begin{equation}\label{pinn:eq1c}
            \left\{ \left( \mathbf{x}^{\left( \Gamma \right)}_1, t_1^{\left( \Gamma \right)} \right), \dots, \left( \mathbf{x}^{\left( \Gamma \right)}_{r_\Gamma}, t_{r_\Gamma}^{\left( \Gamma \right)} \right) \right\} \subset \Gamma \times \left[0, T \right],
        \end{equation}
        and $r_0$ points for the initial condition
        \begin{equation}\label{pinn:eq1d}
            \left\{ \mathbf{x}_1^0, \dots, \mathbf{x_{r_0}^0} \right\} \subset \Omega .
        \end{equation}
        Then, the training of the networks consists of minimizing the loss function given by
        \begin{equation}\label{pinn:eq2}
            \mathcal{L} \left( \mathbf{w} \right) = 
            \frac{1}{r_\Omega} \sum_{i=1}^{r_\Omega} \left\| \mathcal{A} \left[ u \left( \mathbf{x}^{\left( \Omega \right)}_i, t^{\left( \Omega \right)}_i \right), t^{\left( \Omega \right)}_i \right] \right\|^2 + 
            \frac{1}{r_\Gamma} \sum_{i=1}^{r_\Gamma} \left\| \mathcal{B} \left[ u \left( \mathbf{x}^{\left( \Gamma \right)}_i, t^{\left( \Gamma \right)}_i \right), t^{\left( \Gamma \right)}_i \right] \right\|^2 +
            \frac{1}{r_0} \sum_{i=1}^{r_0} \left\| \mathcal{I} \left[ u \left( \mathbf{x}^0_i, 0 \right), 0 \right] \right\|^2
        \end{equation}
        % \fp{così sembra che minimizziamo la pde al continuo, direi scriviamo che esprimiamo la NN come rete che ha dei pesi $s$ (forse meglio $w$ di weights?) e il problema di minimo è su quella variabile. \textcolor{red}{Provato a chiarire meglio.}}
        with respect the weights vector $\mathbf{w}$, including the physical knowledge by directly exploiting the differential problem, where  $\left\| \cdot \right\|$ represents a suitable norm.
        In case of a time independent differential problems, the loss \eqref{pinn:eq2} does not integrate the component associated with the initial condition.
        Solving inverse problems requires also the integration of physical or geometrical parameters in the vector of weights $\mathbf{w}$ defining the neural network. These weights are then optimized and updated during the training phase accordingly to the physical knowledge incorporated in the loss function \eqref{pinn:eq2}.
        Toward this goal, the neural network can easily integrate supervised terms, e.g.\ when sensor measurements are available or when imposing a known function as a boundary condition. It is worth noting that, by including additional data-informed components to the global loss function, the training procedure benefits from additional information, but in general the optimization task is more difficult due to the complex loss landscape, and weighted sum strategies could be applied.

        Indeed, already from their first introduction \cite{Lagaris1998987,Raissi2019686},
        % \fp{add ref, non solo karniadakis in realtà ce ne stava uno prima devo cercare\textcolor{red}{articolo del 98. Lo conosco bene :D, lo provai a implementare con MATLAB quando iniziai con le PINN :D}}, 
        PINNs have reached a significant interest in the community, with several recent improvements related to two main issues: the loss balancing and the causality.
        The first affects the learning efficacy since different terms in the loss function are associated with Neural Tangent Kernel (NTK) eigenvalues of different magnitudes \cite{Wang2022,HirschConvergenceSketchingBasedEfficient2025a}. Since the NTK spectrum determines how quickly each error component decreases, significant eigenvalue disparities lead to imbalanced convergence: some terms are minimized rapidly, while others stagnate. This stiffness in the training dynamics undermines the effectiveness of gradient descent, ultimately reducing the accuracy of the learned solution.
        % \fp{qua forse non si capisce benissimo, citerei il neural tangent kernel e la velocità di apprendimento di ciascuna componente del vettore dei pesi \textcolor{red}{Ho provato a chiare e citato Wang.}}
        The second issue is related to the imposition of the principle of causality in the learning of the approximate solution for time-dependent differential problems, meaning that local variations in the initial or boundary conditions of a spatio-temporal dynamical system influence its subsequent states over time \cite{Wang2024}.

        Several approaches dealt with the loss imbalance issue, aimed at improving the learning capability of NNs.
        Yu et al.~\cite{Yu2022} propose Gradient PINNs (GPINNs) based on integrating gradient information in the loss function to reduce loss fluctuation in the training phase.
        McClenny and Braga-Neto \cite{McClenny2023} introduce Self-Adaptive PINNs (SA-PINNs) employing additional self-adaptive weight in the loss function that are maximized at the training point where the loss is bigger.
        Zeng et al.~\cite{Zeng2023} proposed Competitive PINNs (CPINNs), which replace the classical squared-residual loss with a game-theoretic minimax formulation. In this approach, a discriminator network learns to detect the PDE and boundary violations made by the PINN, while the PINN is trained to minimize them, achieving higher accuracy and faster convergence across linear and nonlinear PDEs.
        Zou et al.~\cite{Zou2025} proposed an ensemble PINN framework to capture multiple solutions of nonlinear differential equations, a challenge where standard PINNs typically converge to a single mode. The approach systematically uncovers diverse stable and unstable solutions by exploiting random initialization and the deep ensemble method. Moreover, realistic PINN outputs can be used as initial guesses for conventional solvers (Finite Difference Methods, FEMs, Spectral Element Methods), establishing a general and efficient strategy for addressing solution multiplicity.
        % Then, the learning phase requires the update of self-adaptive weights and network parameters.
        Finally, Anagnostopoulos et al.~\cite{Anagnostopoulos2024} proposed Residual-Based Attention PINNs (RBA-PINNs) by exploiting an attention scheme coming from Transformer architectures \cite{Xu202312113} that evaluates the residual of the differential problem at each collocation point.%\fp{add cit \textcolor{red}{Survey su multimodal con Transformer}}

        Trying to respect the temporal causality has also led to different advancements of the original PINN approach.  Wang et al.~\cite{Wang2024} introduced Causal PINNs, a reformulation of the loss function with temporal weights that enforce the causal structure of time-dependent PDEs. By ensuring that residuals at later times are minimized only after earlier ones are resolved, this approach corrects the NTK-induced bias of standard PINNs.
        % \fp{qui forse non chiarissimo. \textcolor{red}{Provato a essere breve ma più chiaro}}
        Instead, in Valentino et al.~\cite{Valentino2025541}, the authors introduce a Step-by-Step Time Discrete PINNs based on the integration of an iterative scheme typical of classical numerical methods to obtain a semi-discretization of the time interval and enforce the learning from the initial condition of the problem.
        % \fp{per pinn forse possiamo citare un pò più cose \textcolor{red}{Aggiunto Competitive PINN e Ensemble PINN.}}
    
    \section{A comprehensive framework for cultural heritage}\label{pw:}

    This Section introduces the proposed framework for integrating Internet of Things (IoT), data-driven AI-based approaches, and physical knowledge to conserve cultural heritage.
    
\begin{figure}[!ht]
        \centering
        \includegraphics[width=\linewidth]{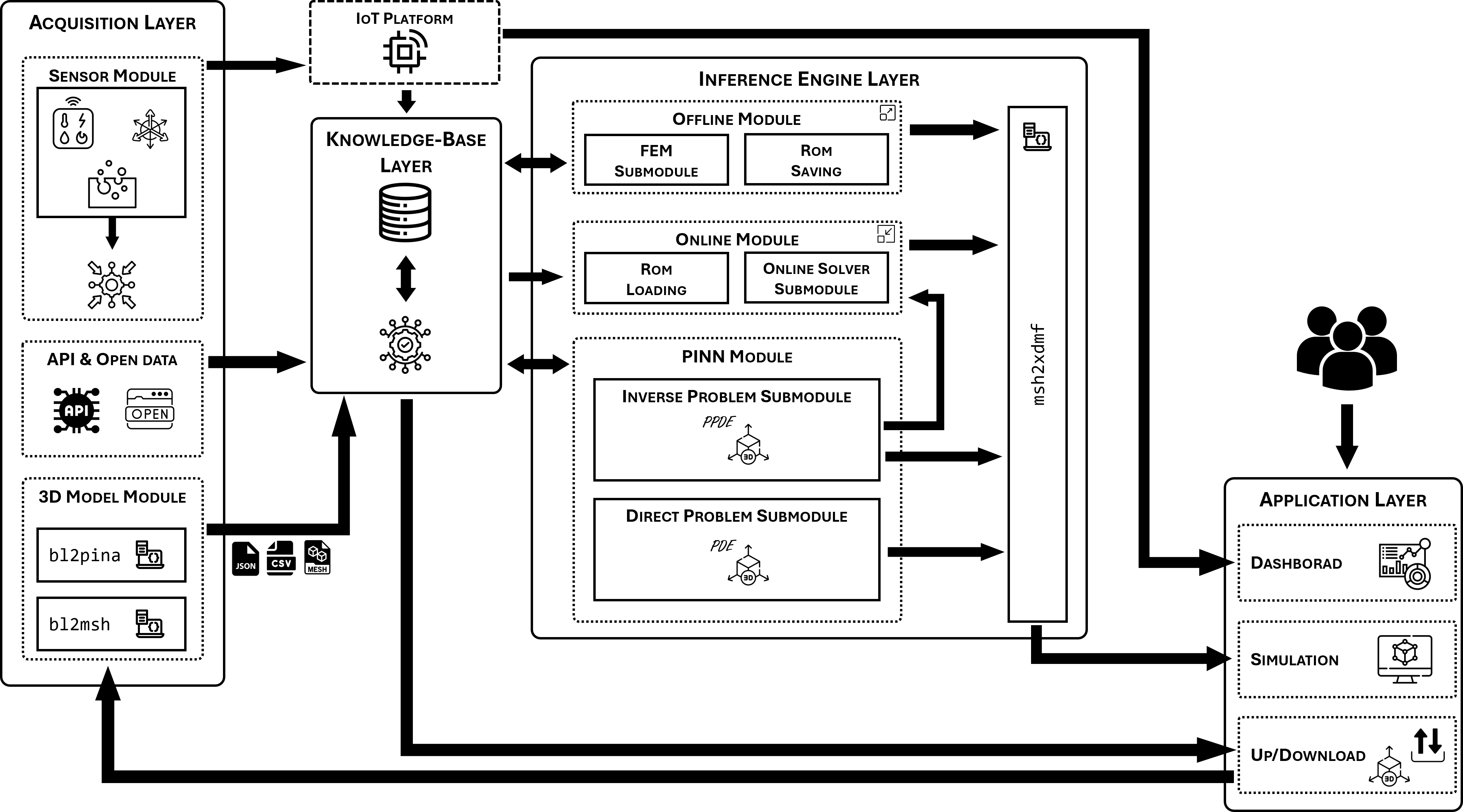}
        \caption{The architecture associated with the proposed framework consisting of four functional layers: the Acquisition Layer, the Knowledge-Base Layer, the Inference Engine Layer, and the Application Layer.}
        \label{pw:fig_1}
    \end{figure}

    Our comprehensive strategy is characterized by four fundamental requirements:
    \begin{itemize}
        \item the ability to acquire 3D models of cultural assets;
        \item the necessity to manage data from sensors and integrate them in the elaboration phase;
        \item an efficient offline-online ROM-like procedure to simulate the cultural asset;
        \item the possibility to train and load physics-aware neural network models for direct and inverse PDE problems.
    \end{itemize}
    To address the aforementioned points, our framework exploits a four-layer architecture described in Figure \ref{pw:fig_1}.
    
    % \fp{forse nell'immagine meglio invertire rom saving e fem submodule? cioè se leggo da sinistra a destra direi che  prima mi serve fem \textcolor{red}{Fatto :D}.}

    % \noindent\textbf{Acquisition Layer.} 
    \subsection{Acquisition Layer}
    This first layer focuses on managing 3D models and sensor data.
    It consists of three components: a module for the acquisition of data from sensors defined as the Sensor Module, a component for integrating data from external sources defined as API \& Open Data, and finally, a module to pre-process digital models of cultural assets defined as the 3D Model Module.
    The Sensor Module manages centralizers and sensors for the collection of data from cultural assets.
    Environmental conditions and monitored physical processes, such as corrosion, temperature monitoring, and pollution effects, are acquired and stored through sensors.
    This module is event-oriented and aims to manage the collection of data from sensors.
    The centralizers synchronize the acquired data and send it to the IoT platform in a key-value format. The IoT platform allows for data storage, visualization, and transfer through REST APIs \cite{Garg2019}.
    The component API \& Open Data also exploits REST API services to acquire data from external sources, aiming to integrate information related to the environmental conditions in which the analysis on the cultural asset is carried out.
    This component communicates directly with the Knowledge-Base Layer.
    Finally, the 3D Model Module manages the digital models uploaded by users, enabling the analysis of cultural assets.
    The module acquires the digital replicas of cultural assets as .blend files, which are processed to store in the Knowledge-Base Layer information related to the mesh elaborated on the cultural asset, the tables related to the random sampling of the assets to perform physics-based AI applications, and the $\textsf{xdmf}$ files that allow the visualization of simulations performed by the Inference Engine Layer.
    Therefore, this module prepares the data for applying hybrid approaches based on sensors, physical knowledge, and AI.
    The detailed behavior of this module is described in Subsection \ref{pre:}.

    % \noindent\textbf{Knowledge-Base Layer.} 
    \subsection{Knowledge-Base Layer} 
    The second layer of the framework represents its storage core.
    In this layer, the data from sensors arrive in the database through the IoT platform via REST API services.
    Data acquired from the API \& Open Data component are automatically exploited here, similar to the 3D Model Module, providing information related to the mesh, the sampling for the application of physics-based approaches, and the files needed to provide simulations to users.
    In addition, this layer also communicates with the Inference Engine Layer, which requires storage for training neural network models and ROMs.
    Moreover, this layer is linked with the Application Layer, allowing for the possibility to download data related to the 3D Model Module.
    Finally, this layer contains a pre-processing module that enables the control of anomalies related to missing data before storage.
    % \fp{anche qui come sopra possiamo specificare meglio. Inoltre se introduciamo PINNS e offline poi non si capisce perché quelle sono nella fase successiva.}

    % \noindent\textbf{Inference Engine Layer.} 
    \subsection{Inference Engine Layer}
    The third layer permits the elaboration of reliable simulations to users who are experts in the field of cultural heritage conservation.
    It consists of four different modules: the Offline Module, the Online Module, the PINN Module, and the module named $\textsf{msh2xdmf}$.
    The Offline Module performs the offline phase related to the application of ROMs, specifically the POD, and includes two submodules: the FEM Submodule and the ROM Saving.
    The FEM Submodule solves the differential problem to provide the Full Order solution and extract the reduced basis to be exploited during the online phase.
    The ROM Saving module interacts with the database of the Knowledge-Base Layer to store the solutions related to the selected snapshots, i.e., the fixed parameters of the parametric differential problem.
    The Online Module, instead, performs the online phase of the ROM procedure and consists of the ROM Loading and Online Solver Submodules.
    The ROM Loading allows access to the appropriate stored data for the reduced system.
    Instead, the Online Solver Submodule performs the online resolution of the problem once projected on a low-dimensional reduced space and, after its resolution, it brings back the solution lifting it onto the original space.
    The PINN Module allows the framework to integrate the physical knowledge of the phenomena with collected data through two different objectives: identifying parameters, solving inverse problems, or solving direct ones, thus consisting of the Inverse Problem and the Direct Problem Submodules.
    Specifically, they exploit the 3D model elaboration and collected data through the Acquisition Layer.
    In addition, the Inverse Problem Submodule communicates with the Online Module, providing the parameters of the differential problem that fits the data and the physics behind the specific analysis.
    Finally, the module named {\textsf msh2xdmf} exploits the 3D model and integrates the approximated solution identified by the other modules.
    This module provides the visualization of predictions obtained by integrating IoT, physical knowledge, and AI.

    % \noindent\textbf{Application Layer.} 
    \subsection{Application Layer} 
    This last layer allows the interaction between the framework and users for the cultural heritage conservation.
    It comprises the Dashboard Module for visualizing data, the Simulation Module to furnish the simulation through the elaboration of the Inference Engine Layer, and the UP/Download Module.
    This module permits users to upload 3D models related to the cultural assets and download data and models included in the Knowledge-Base Layer.
    
    % Finally, to better explain the framework's behavior, it is analysed in Section \ref{res:} concerning each numerical experiment performed.
    
    \subsection{3D Model Module: processing geometries to provide simulations}\label{pre:}

    The architecture described in Section \ref{pw:} and depicted in Figure \ref{pw:fig_1} is characterized by the possibility to work with very complex 3D geometries via the 3D Model Module of the Acquisition Layer.
    For this purpose, the architecture exploits the API provided by Blender\footnote{https://www.blender.org/about/}, an open-source software for creating and managing 3D content.
    The application of Blender regards multiple fields, such as three-dimensional modeling, animation, rendering, and physical simulation.
    Blender is a cross-platform software that allows access to the modeling phase both by experts and via Python scripts.
    In addition, as mentioned above, it is possible to exploit APIs via the $\textsf{bpy}$ library \cite{Sybren2025}, enabling a seamless and easy interaction with the geometries.

    \begin{figure}[!ht]
    \centering
    % Prima immagine
    \subfigure[Rock blender model]{
        \includegraphics[width=0.25\textwidth]{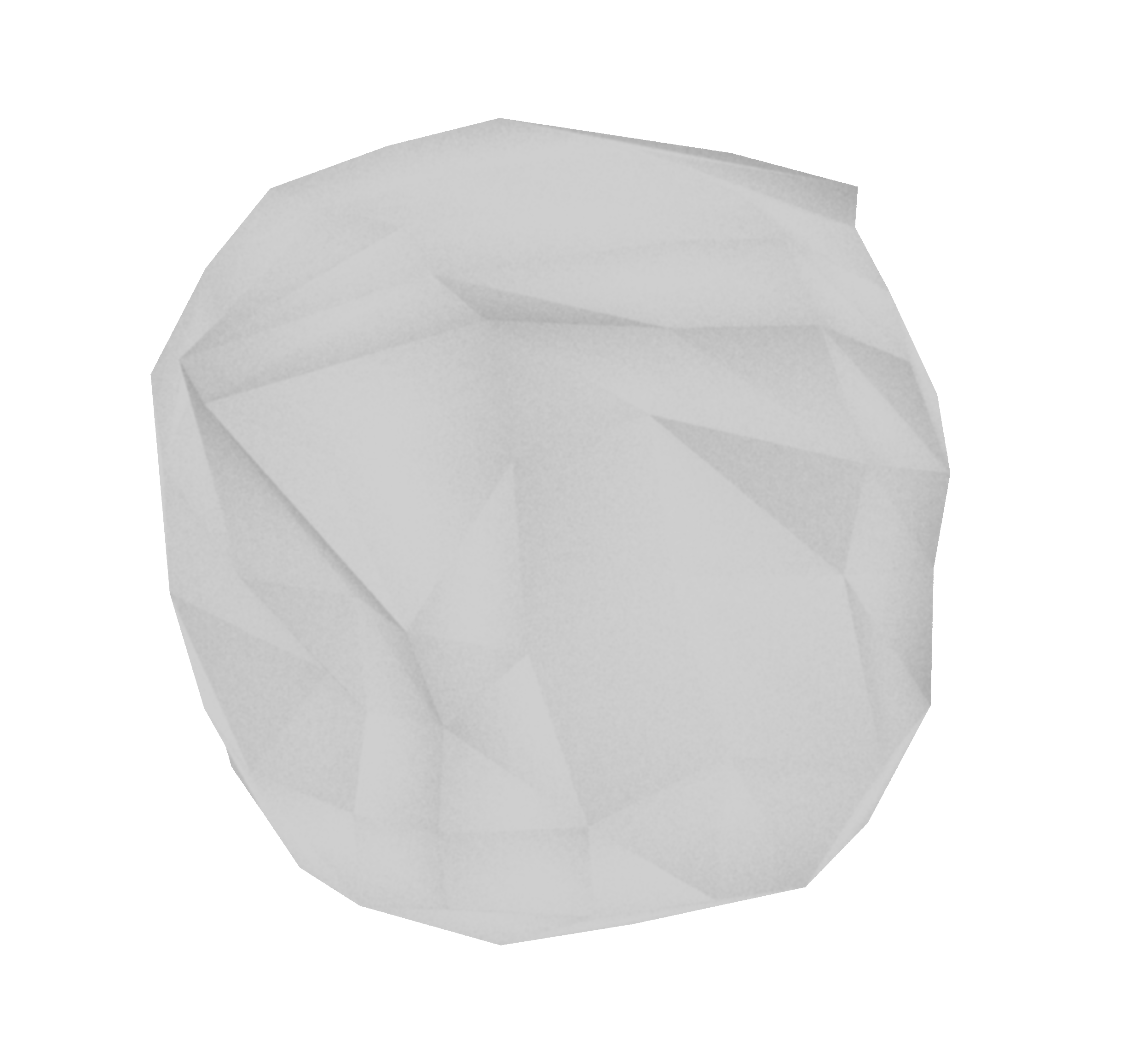}
        \label{pre:fig_1a}
    }
    % Seconda immagine
    \subfigure[Rock mesh]{
        \includegraphics[width=0.25\textwidth]{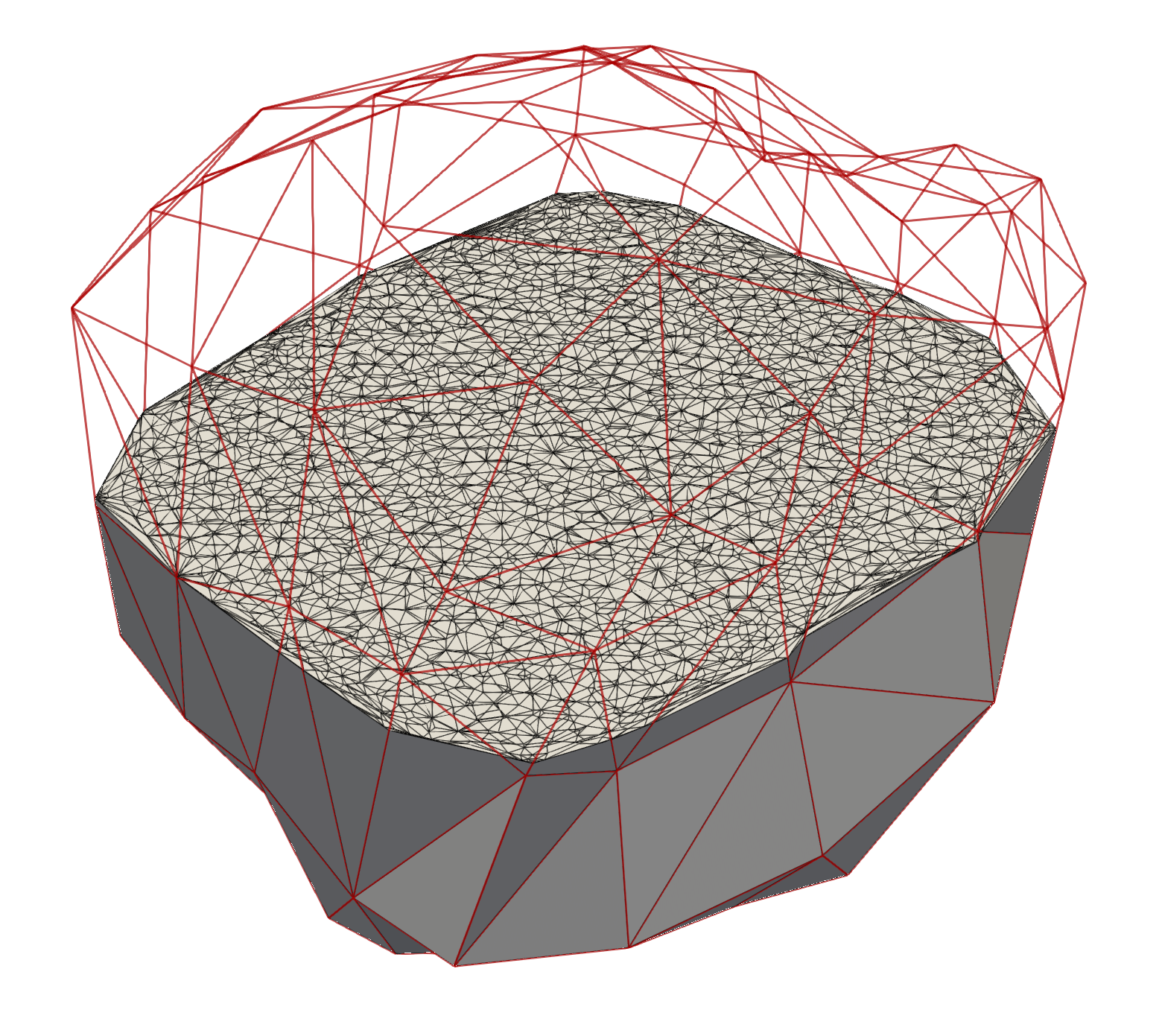}
        \label{pre:fig_1b}
    }
    % Terza immagine
    \subfigure[Rock $\mathsf{xdmf}$ file]{
        \includegraphics[width=0.25\textwidth]{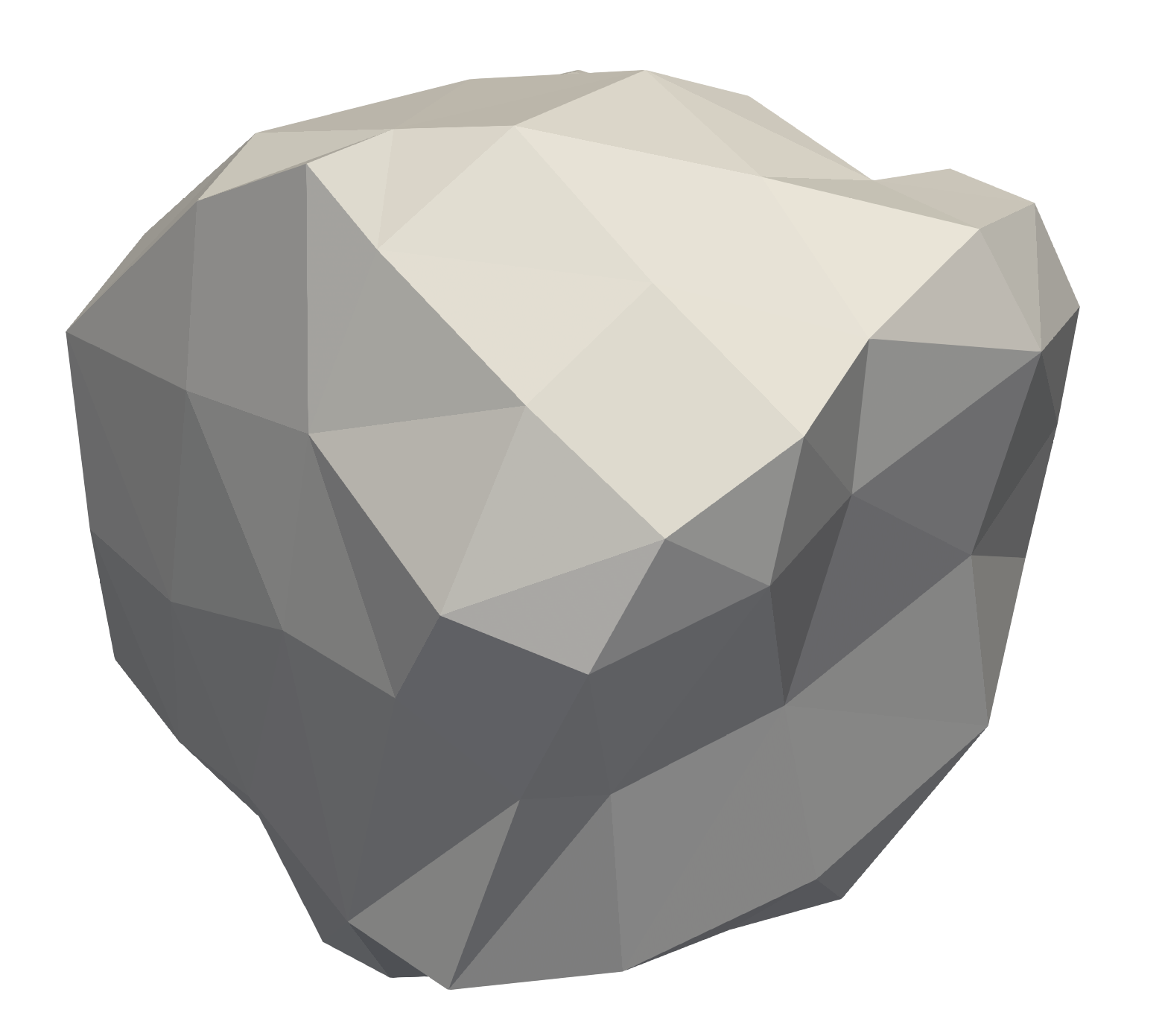}
        \label{pre:fig_1c}
    }
    \caption{Acquisition of the 3D model related to the rock represented in Figure. The 3D Model (a) is acquired and elaborated to provide to the Knowledge-Base Layer the list of collocation and boundary point for PINNs. In addition, the framework (b) elaborates the mesh and (c) prepares the visualization through an XDMF file.}
    \label{pre:fig_1}
\end{figure}

    \begin{figure}[!ht]
    \centering
    % Prima immagine
    \subfigure[Column blender model]{
        \includegraphics[width=0.19\textwidth]{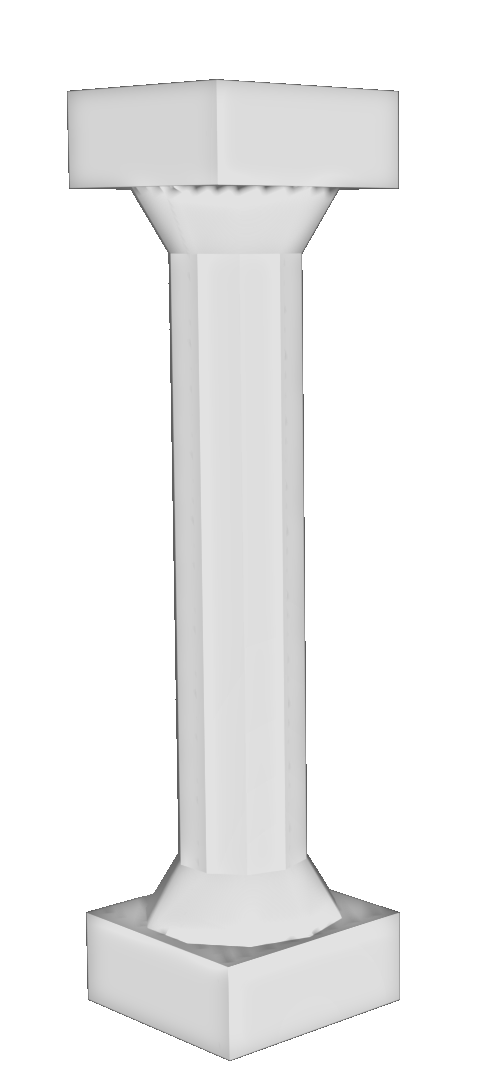}
        \label{pre:fig_2a}
    }
    \hspace{1.0cm}
    % Seconda immagine
    \subfigure[Column mesh]{
        \includegraphics[width=0.175\textwidth]{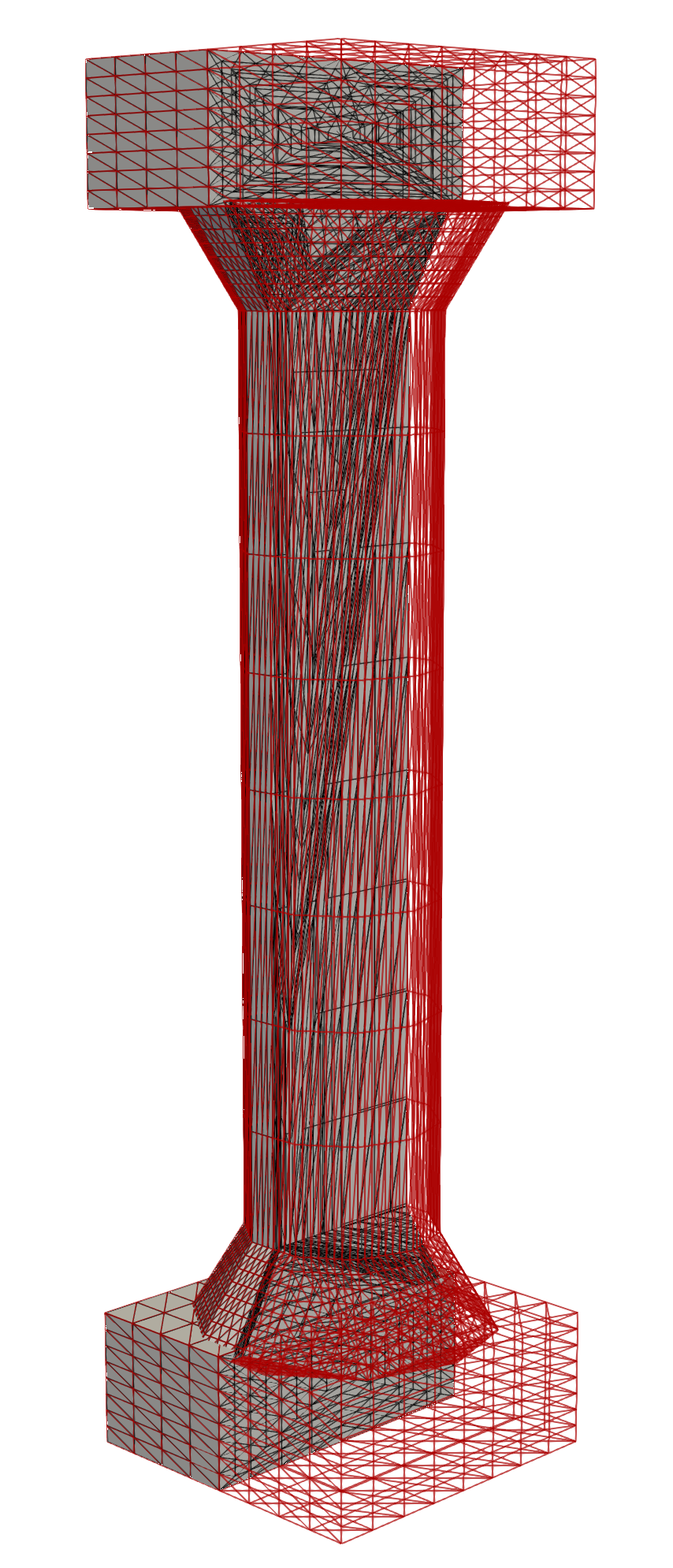}
        \label{pre:fig_2b}
    }
    \hspace{1.0cm}
    % Terza immagine
    \subfigure[Column $\mathsf{xdmf}$ file]{
        \includegraphics[width=0.175\textwidth]{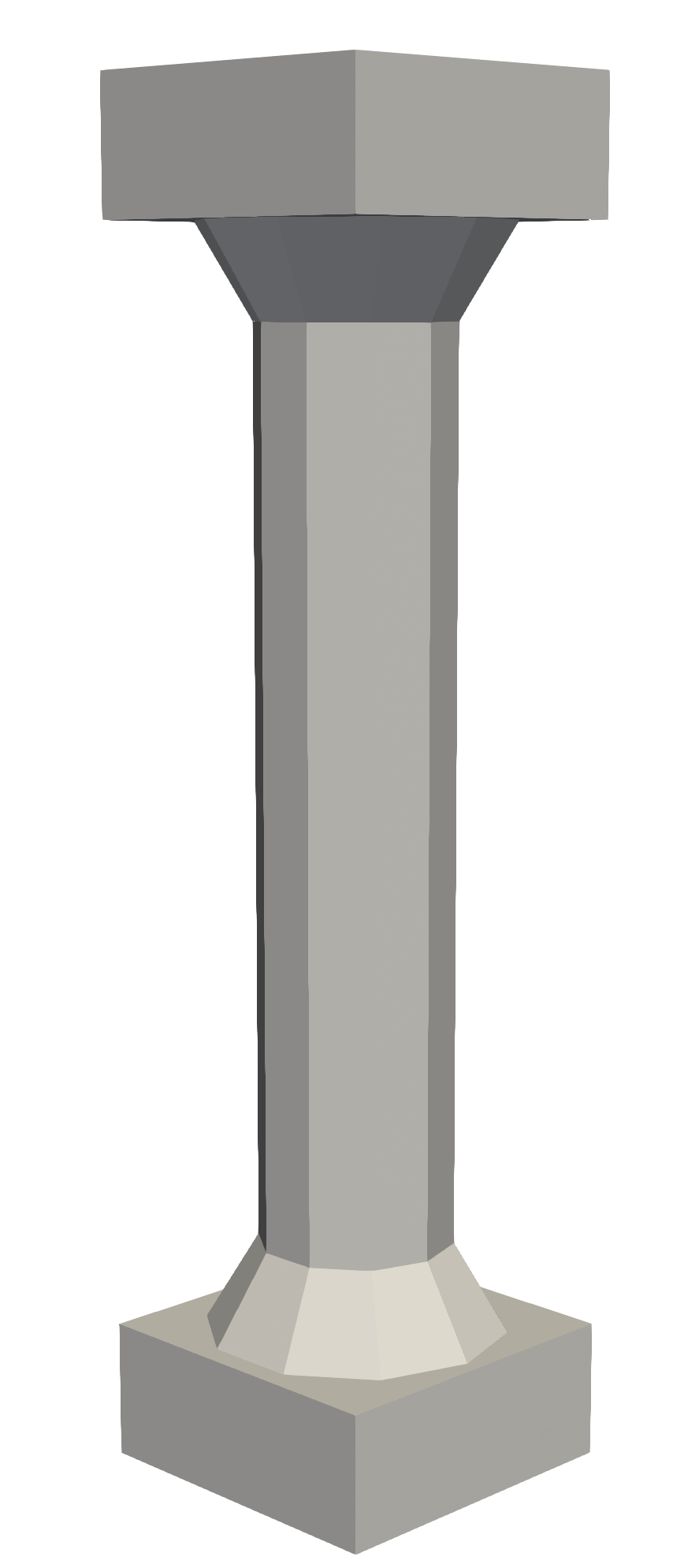}
        \label{pre:fig_2c}
    }
    \caption{Acquisition of the 3D model related to the column.}
    \label{pre:fig_2}
\end{figure}

    \begin{figure}[!ht]
    \centering
    % Prima immagine
    \subfigure[Temple blender model]{
        \includegraphics[width=0.3\textwidth]{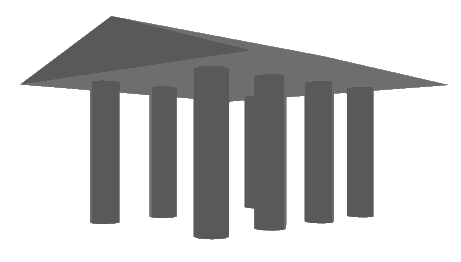}
        \label{pre:fig_3a}
    }
    % Seconda immagine
    \subfigure[Temple mesh]{
        \includegraphics[width=0.3\textwidth]{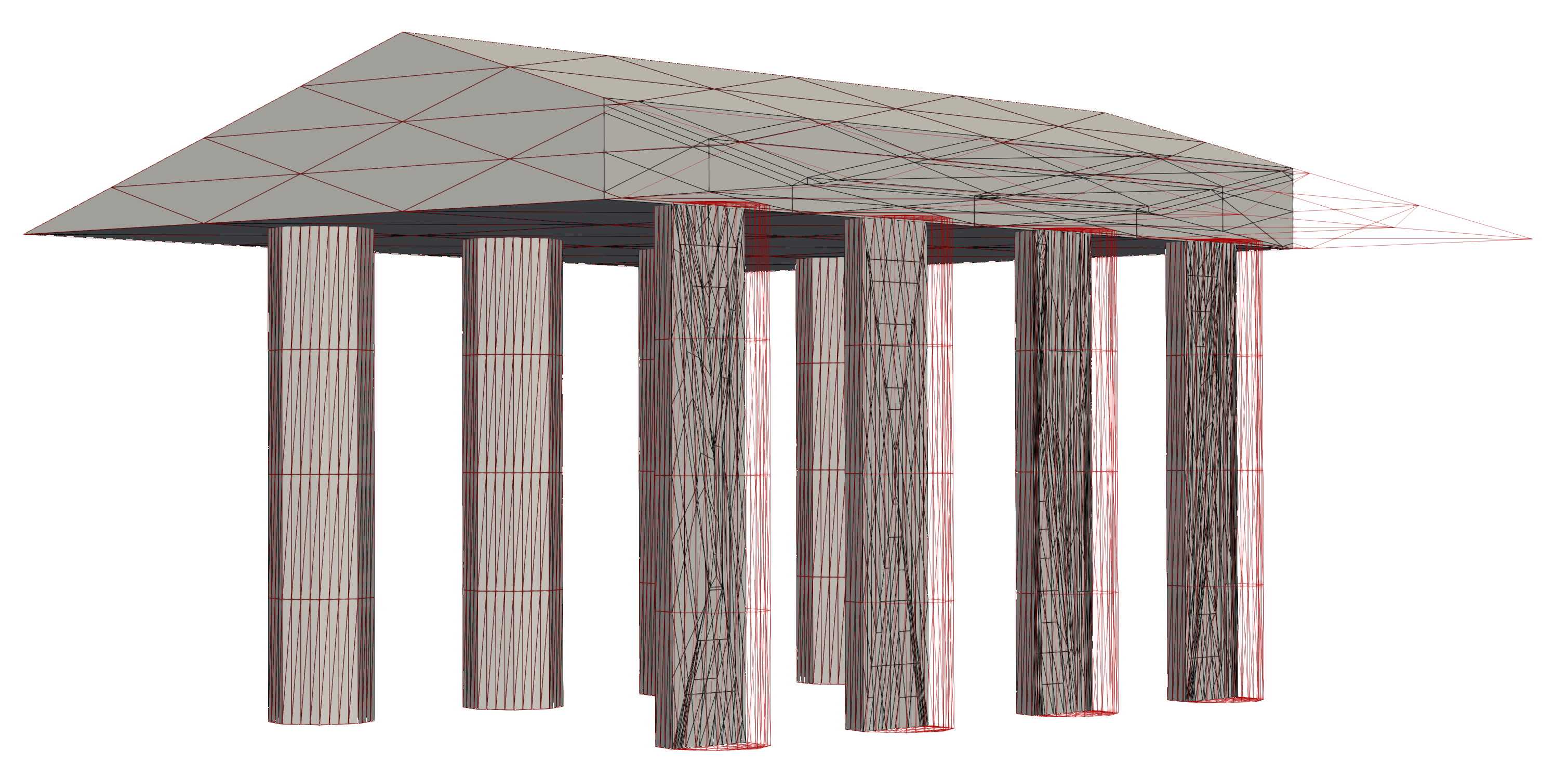}
        \label{pre:fig_3b}
    }
    % Terza immagine
    \subfigure[Temple $\mathsf{xdmf}$ file]{
        \includegraphics[width=0.3\textwidth]{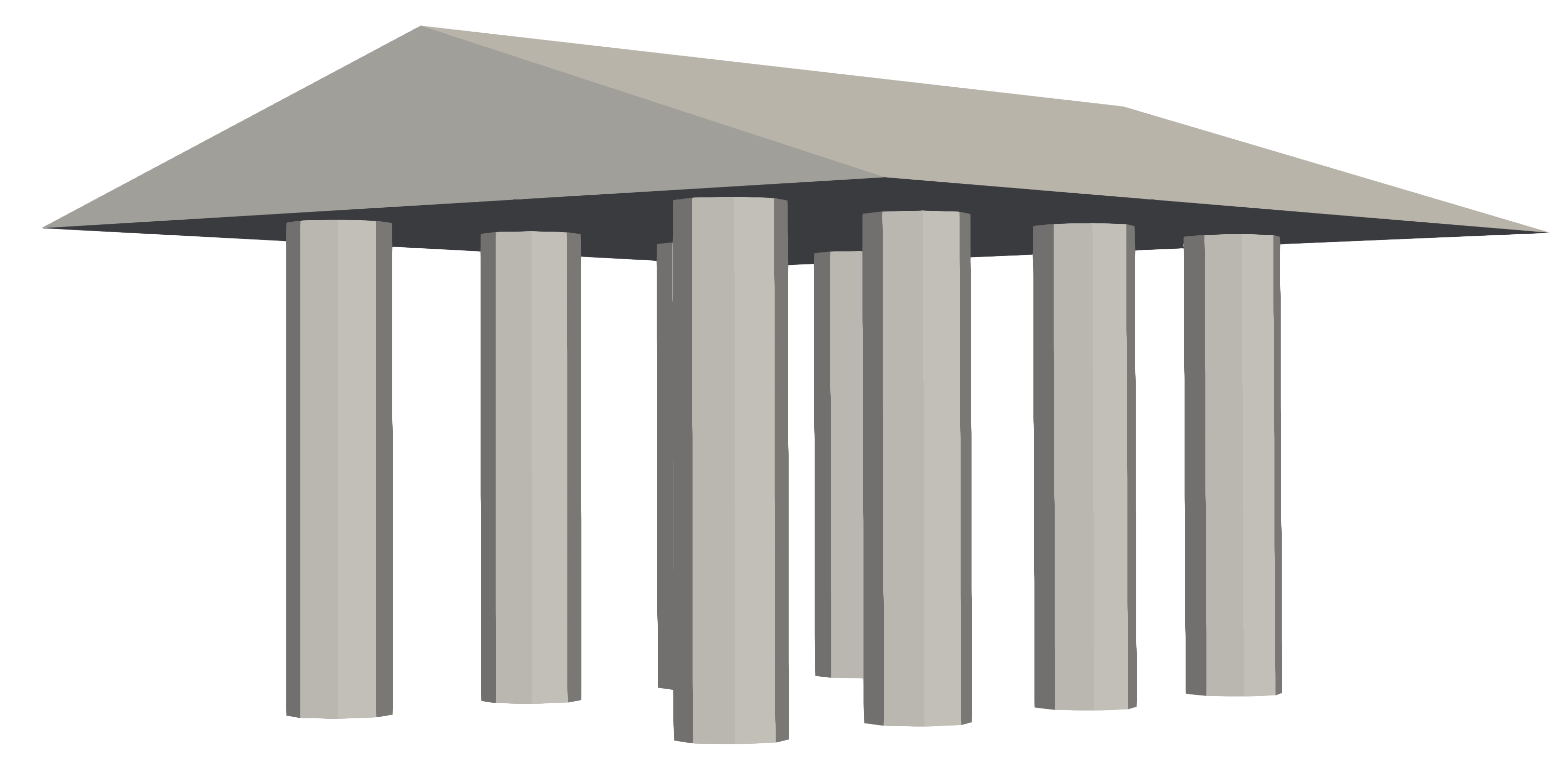}
        \label{pre:fig_3c}
    }
    \caption{Acquisition of the 3D model related to the temple.}
    \label{pre:fig_3}
\end{figure}
    % \fp{Queste tre le visualizzerei nella stessa prospettiva. Intendo la stessa per ogni figura singola. Inoltre, dici meglio la mesh così o esportata direttamente da paraview? ho aggiunto un esempio in images/rappresentazioni/rock}
    
    The architecture exploits Blender's API to acquire the information of interest after an appropriate pre-processing of the model.
    In fact, after a phase in which the model is triangulated, information about points, faces, scales, location, and the list of normals to the faces is acquired.
    The sub-modules of the architecture then use this information to integrate it into the PINA library \cite{coscia2023physics}, a Python package developed for PINNs and more broadly SciML, via the developed tool $\textsf{bl2pina}$ and by automatically generating the mesh of the domain for the differential problem at hand through $\textsf{bl2msh}$.

    In particular, the acquisition of the blender model exploits an organization of the model information according to semi-structured data in a key-value format that allows the model summary to be exported in JSON files.
    The key-value structure allows, on the one hand, integrating the model information both within the PINN and the ROM workflows.
    For the PINN workflow, the model includes a description of the domain's boundary, so it is easier to describe all processes requiring the acquisition of data both on the boundary and within the domain.
    The triangulation of the boundary is obtained via Blender's API, while PINA library allows the integration of the model for the definition of the domain for the PINN problem.
    However, the architecture integrates the services of PINA by defining integration and sampling strategies, also managing the cases in which the 3D model is composed by multiple 3D objects, by creating a list of key-values structures.

    Thus, the proposed architecture automatically generates meshes through the 3D models starting from such list of key-values structures and exploiting the information acquired from the Blender APIs.
    The blender API also allows the proposed framework to elaborate the model to sample the digital replica of the asset. The sampled points, exploited to obtain a simulation through PINNs, act as alternative mesh points and are stored in tabular format in the Knowledge-Base layer.
    % \fp{questa ultima frase non mi è chiarissima}

    To test of the efficacy of the developed strategies in integrating the Blender model for PINNs and ROMs purposes, we show in Figures the 3D models described in Figures \ref{pre:fig_1}, \ref{pre:fig_2}, and \ref{pre:fig_3}: a rock, a column, and a temple\footnote{The 3D models for the rock and the column have been acquired from: \url{https://free3d.com/3d-model/low-poly-rock-4631.html} and \url{https://free3d.com/3d-model/white-column-44873.html}, respectively. The 3D model of the temple has been obtained as a re-elaboration of the one available at: \url{https://free3d.com/3d-model/temple-57751.html}}.
    % \fp{il tempio l'avevi fatto tu no?}
    These structures show the performance of 3D Model Module for increasing level of difficulty in their acquisition.
    In fact, the rock solely consists of a unique 3D object, the column consists of three 3D objects (the basis, the central part, and the top component), while the temple consists of nine 3D objects (eight columns and the top part).
    Figures \ref{pre:fig_1b}, \ref{pre:fig_2b}, and \ref{pre:fig_3b} demonstrate the robustness of the architecture to automatically acquire the object from the Blender file and produce an accurate mesh for all benchmarks.

    Finally, managing 3D models also allows the architecture to generate $\textsf{xdmf}$ (eXtensible Data Model and Format) and HDF5 (Hierarchical Data Format version 5) files via the \textsf{msh2xdmf} module. 
    $\textsf{xdmf}$ and HDF5 files are often used together to represent and manage complex scientific data, usually coming from numerical simulation, to store large datasets, and within applications where a structured data representation is required.
    In particular, an HDF5 file is a high-performance binary format that allows large amounts of structured data to be stored in a hierarchical structure-based manner, with multi-language compatibility, high storage capacity, and the possibility of compression.
    On the other side, an $\textsf{xdmf}$ file is an XML-based metadata format designed to describe complex scientific data, and it is often used as an index pointing to an HDF5 file for the actual data.
    The idea is to separate metadata (description) from numerical data (content).
    Specifically, the $\mathsf{msh2xdmf}$ module enable the access to the visualization of the simulation elaborated by employing such architecture. 
    % Finally, Figures \ref{pre:fig_1c}, \ref{pre:fig_2c}, and \ref{pre:fig_3c} introduce the $\mathsf{msh2xdmf}$ files related to the rock, the column, and the temple.

    \section{Numerical Results}\label{res:}

    To validate the efficacy of the proposed workflow, we set up here an experimental phase deploying numerical simulations for simulated benchmarking scenarios involving physical problems defined on realistic domains of potential interest for the control and monitoring of cultural heritage assets. 
    % numerically evaluates the efficacy of the proposed framework based on the PINN Module.
    In particular, we test the performances of the integrated environment, from the data-acquisition to the numerical prediction, in both settings in which: (i) physics is known and enforced via PINN, and the goal is to identify the physical parameters by exploiting the Offline-Online Modules in combination with the Inverse Problem one, and (ii) we use the Direct Problem Submodule to obtain an efficient evaluation of the PDE for known parameter values. 
    % \fp{\textcolor{green}{qui forse aggiustiamo rispetto al contenuto se cambia, fare un check alla fine}} \fp{mi pare che l'inverse lo facciamo alla fine di ogni test}
    %  of both benchmark problems where the physics is known and simulated scenarios that integrate physical and multi-physical problems with data obtained from simulated sensors.
    % Therefore, the experimental section consists of two subsections.
    % The first tests the architecture related to the Offline-Online Modules of the architecture in Figure \ref{pw:fig_1} combined with the Inverse Problem Submodule of the PINN Module.
    % The third aims to exploit the Direct Problem Submodule of the PINN Module.
    In both cases, the objective consists of evaluating the accuracy of the simulations obtained to testify the effectiveness of the proposed architecture in supporting expert users for preventive maintenance of cultural assets.

    \subsection{Offline-Online modules with Inverse Problem Submodule}\label{fr:}

    The experimental evaluation of the architecture exploiting ROMs for the cultural heritage maintenance requires testing specific modules of all architecture's layers, as highlighted in Figure \ref{fr:fig_1}.

\begin{figure}[!ht]
    	\centering
    	\includegraphics[width=.9\linewidth]{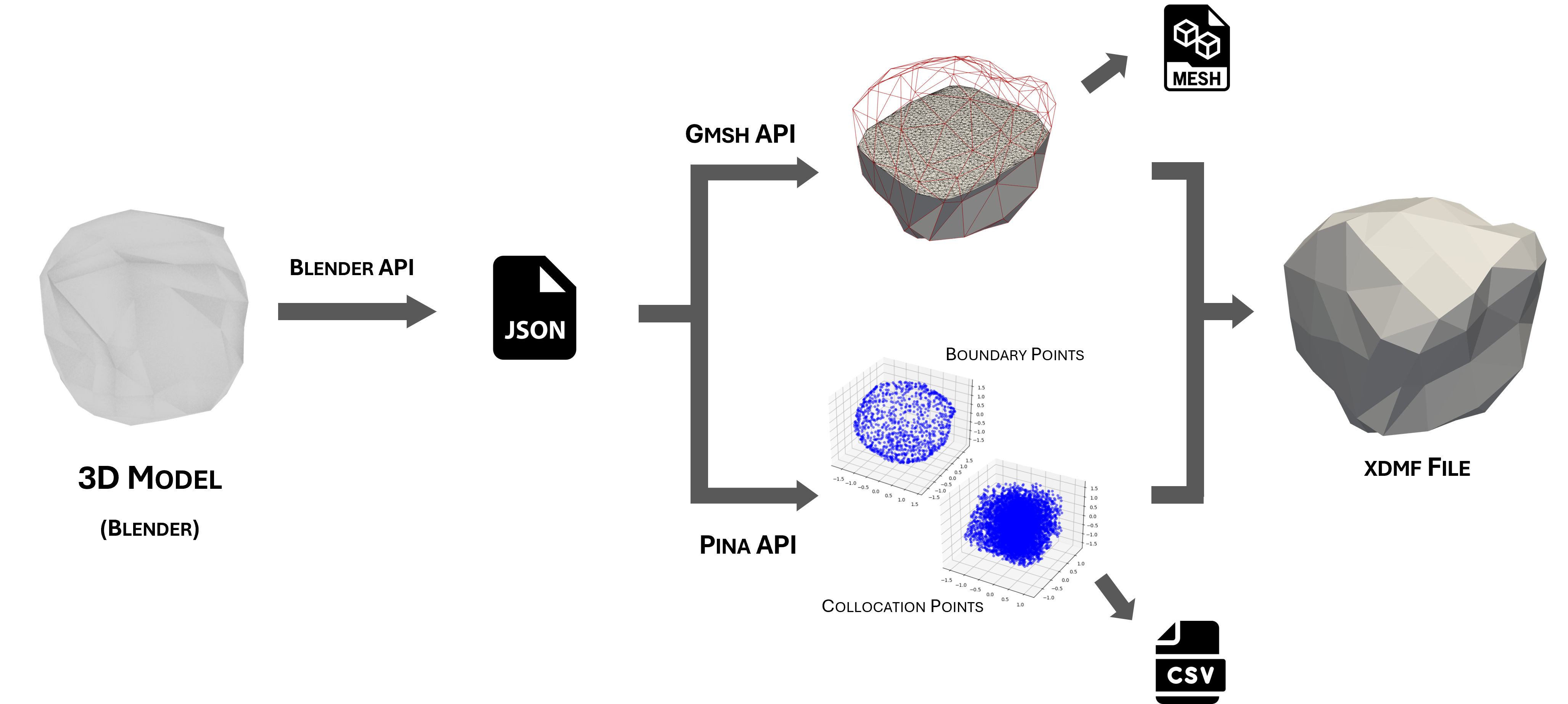}
    	\caption{Steps of the model elaboration performed by the 3D Model Module.}
    	\label{fr:fig_0}
    \end{figure}

\begin{figure}[ht]
        \centering
        \includegraphics[width=\linewidth]{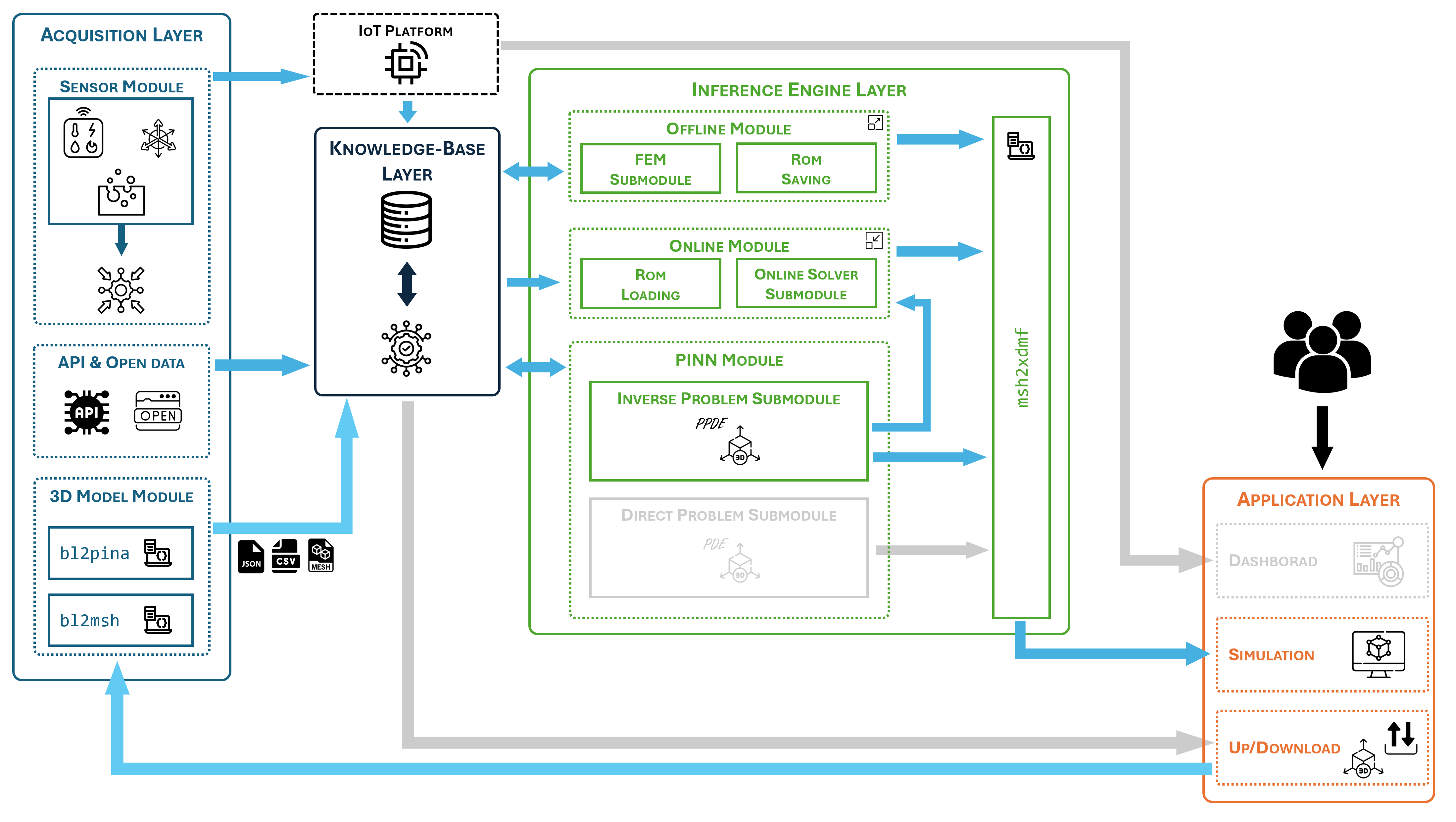}
        \caption{Highlighted are the active modules of the architecture to perform the numerical approximation of parametrized differential problems.}
        \label{fr:fig_1}
    \end{figure}

    The Blender model, loaded through the Up/Download Module, allows the acquisition of the digital replica of the cultural asset. Therefore, the 3D Model Module elaborates the Blender file to obtain a JSON file in which the main features of the domain are collected. This JSON file permits the automatic elaboration of the object's mesh, the random sampling of the domain, and the visualization and post-processing of the simulations. In particular, in Figure \ref{fr:fig_0} we depict the workflow of the 3D Model Module. The main features of the asset stored in the Blender file are collected through a JSON file, from which they are processed via the GMSH API \cite{Geuzaine20091309} to export the mesh of the model in an $\mathsf{msh}$ file, and the PINA API \cite{coscia2023physics} to acquire boundary and collocation sampling points. In addition, the framework produces the $\mathsf{xdmf}$ file in which the numerical solutions obtained through the Inference Engine Module are integrated to visualize the output of interest.
    % \fp{qui forse non chiarissima seconda parte e nella prima abbiamo definito chiaramente a quale ci riferiamo?}
    
    In addition, the Sensors module and the API \& Open Data Module, communicating with the IoT platform, allow the acquisition of data related to the phenomenon, providing collected data to the database in the Knowledge-Based Layer.
    % Specifically, the Sensors Module communicates with , which provides collected data to the database in the Knowledge-Based Layer.
    This data allows for solving inverse problems and identifying parameters related to the parametric field configurations as solution to PDEs.
    
    Therefore, the Inference Engine Layer can analyze the parametric PDE by activating the Offline Module, obtaining reliable reference solutions to apply the Proper Orthogonal Decomposition as a Reduced Order Model.
    Specifically, the user can select the POD energy tolerance through the Simulation Module to control its reliability, otherwise, a pre-defined tolerance of $1e-6$ is applied.
    Similarly, the Simulation Module allows choosing the level of exploration of the high fidelity model, i.e.\ the number of sampled snapshots, whose default value is equal to 100. We remark that this number has to be carefully chosen according to the complexity of the parametric space and the computational budget available during the offline phase.

    The Offline Module stores the information needed for the ROM Module in the database through the ROM Saving submodule.
    This step takes advantage of the {\textsf RBniCS} package for Python, allowing the integration of the {\textsf FEniCS} package for the PDE resolution via the Finite Element Method by exploiting the FEM Submodule and the application of the reduced strategies.
    % The resolution of the FEM method happens.

    Then, the Inverse Problem Submodule in the PINN Module take advantage of the physics underlying the analyzed phenomenon, integrating this with the collected data, to identify the parameters of the simulated PDE.
    In this phase, the high-level of customization of the framework allows the user to choose if apply the PINN approach on the mesh points' coordinates or if randomly select points in the domain defined by the 3D model.
    % This process aims to identify the parameters of the parametric PDE analyzed.
    Therefore, the user have access to both the solution provided by PINN and the solution obtained via the POD executed by the Online Solver Submodule of the Online Module.
    
    Finally, the {\textsf msh2xdmf} Module elaborates on the simulation and stores it in an $\textsf{xdmf}$ file built on the available mesh.

    To perform the experimental phase related to this architecture, we elaborated simulated data related to benchmark test problems.
    Moreover, we evaluated the Acquisition Layer's ability to analyze three different 3D models related to different structures.
    The test problems are described as follows.
    After introducing the parametric PDE under investigation, we describe the acquisition of the 3D model, the process to obtain simulated data, and the accuracy performance, by comparing the ROM procedure with full-order and analytical solutions.
    As concerns the reliability of the PINN in solving the problem, we postpone the discussion to subsection \ref{d:}

    \subsubsection{Test Problem 1: Poisson problem on a rock}\label{tp_1:}

        Let us consider a Poisson problem defined on a domain $\Omega \subseteq \mathbb{R}^3$ represented by the rock in Figure \ref{pre:fig_1}, described by the parametric elliptic PDE

        \begin{equation}\label{tp_1:eq1}
            \begin{cases}
                \Delta u \left(x, y, z \right)  = - \left( \alpha^2 + \beta^2 \right) \pi^2 \lambda x \cos{\left( \alpha \pi y \right)} \sin{\left( \beta \pi z \right)} &  \text{in}\ \Omega, \\
                u \left( x, y, z \right) = \lambda x \cos{\left( \alpha \pi y \right)} \sin{\left( \beta \pi z \right)} &  \text{on}\ \partial\Omega,
            \end{cases} 
        \end{equation}
        where the function $u: \Omega \cup \partial \Omega \mapsto \mathbb{R}$ represents the temperature distribution, $\lambda$ is the amplitude of the parametric forcing term, while $\alpha$ and $\beta$ control the spatial oscillations in $y$ and $z$ directions, respectively. The parametric analytical solution, constructed by the method of manufactured solutions, is given by $u \left( x, y, z \right) = \lambda x \cos{\left( \alpha \pi y \right)} \sin{\left( \beta \pi z \right)}$.
        % \begin{equation}\label{tp_1:eq2}
        %     u \left( x, y, z \right) = \lambda x \cos{\left( \alpha \pi y \right)} \sin{\left( \beta \pi z \right)}, \qquad \left( x, y, z \right) \in \Omega .
        % \end{equation}
        The application of the FEM Submodule of the Offline Module requires the derivation of the weak form \eqref{rom:eq2} where $\mu = (\lambda, \alpha, \beta)$, and the forms are defined as follows:
        \begin{equation}\label{tp_1:eq4}
            a \left( u, v \right) = - \int_\Omega \nabla u \cdot \nabla v \, \mathrm{d}x, \qquad
        % \end{equation}
        % \begin{equation}\label{tp_1:eq5}
            L \left( v \right) = - \int_\Omega \Bigl( \left( \alpha^2 + \beta^2 \right) \pi^2 \lambda x \cos{\left( \alpha \pi y \right)} \sin{\left( \beta \pi z \right)} \Bigr) v \, \mathrm{d}x .
        \end{equation}
        % \fp{mi sembra sbagliato il segno della forzante qui}

        The application of the FEM to the problem \eqref{tp_1:eq1} exploits a mesh of $N_h = \num{42343}$ nodes, and the polynomial basis functions are chosen in the first-order Lagrange Finite Element space $\mathbb{P}_1$.
        
\begin{figure}[!ht]
        \centering
        \includegraphics[width=0.6\textwidth]{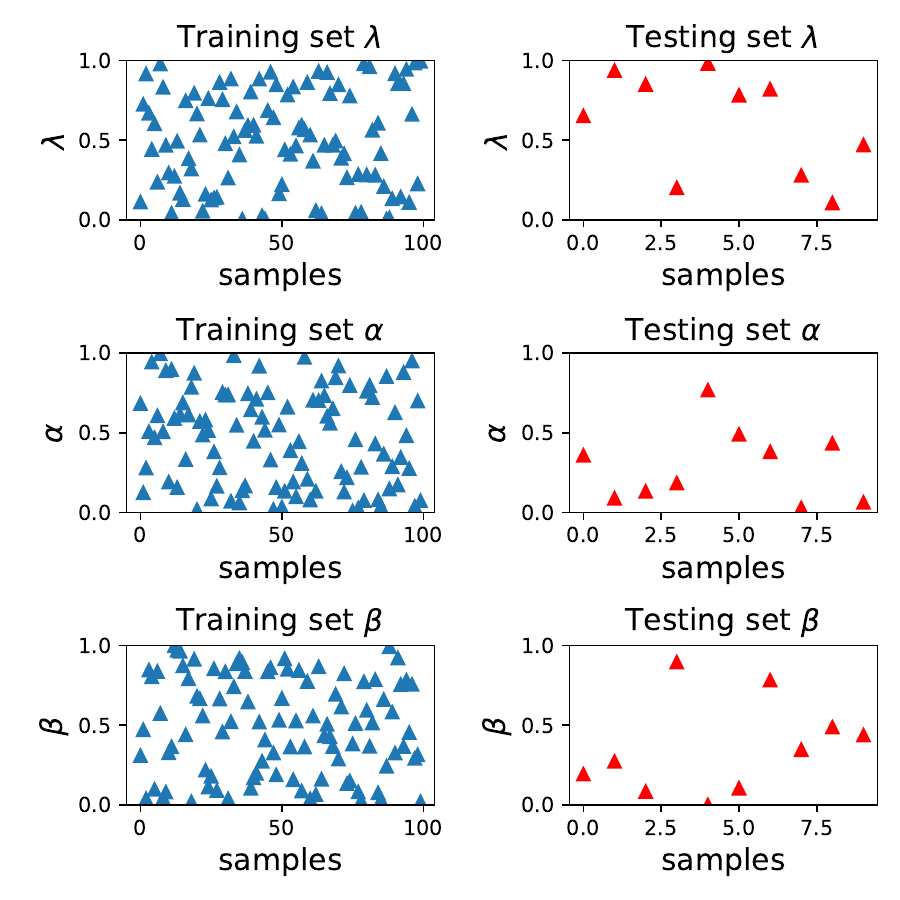}
        \caption{Normal random sampling for $\mu \in [0,1]^3$ and $M = 100$.}
        \label{tp_1:fig_2_1}
        \end{figure}

        The POD procedure then requires the definition of a set of snapshots corresponding to the normal random sampling $\left\{ \left( \lambda_i, \alpha_i, \beta_i \right) \in [0, 1]^3 \, : \, i=1, \dots, M \right\}$, for which we set $M=100$ (see Figure \ref{tp_1:fig_2_1}).         
        The FEM Submodule allows to numerically solve the problem for each sample $\left( \lambda_i, \alpha_i, \beta_i \right)$, defining the matrix $S \in \mathbb{R}^{N_h \times M}$, where $N_h$ represents the number of mesh points (see Figure \ref{pre:fig_1b}). From the POD we extract the reduced basis of order $k$, allowing us to project the system in a lower dimensional space, obtain a surrogate solution in a more efficient way, and investigate the ``reducibility'' of the problem for $k \leq 25$, meaning the dimensionality of the most important modes to be retained to obtain an accurate reconstruction of the reduced approximation. 
        
        % . In Figure \ref{tp_1:fig_2a}, we show the singular value decay expressing the 
        % for the values $k=25$ selected for applying POD for the test problem 1). 
        
        % Applying the Singular Value Decomposition (SVD) \cite{Aggarwal2020} to the matrix $S$ as follows
        % \begin{equation}
        %     S = U \Sigma V^t, \qquad U \in \mathbb{R}^{N_h \times N_h}, \, \Sigma \in \mathbb{R}^{N_h \times M}, \, V \in \mathbb{R}^{M \times M} ,
        % \end{equation}
        % where the matrix $\Sigma$ has only the diagonal entries as non-zero elements. In addition, defining $k \leq r \left(S\right)$, where $r \left( \cdot \right)$ represents the function returns the rank of the matrix, the reduced factorization allows to identify the matrices $U_k \in \mathbb{R}^{N_h \times k}$, $V_k \in \mathbb{R}^{M \times k}$, and $\Sigma_k = \mathrm{diag} \left( \sigma_1, \dots, \sigma_k \right) \in \mathbb{R}^{k \times k}$ (Figure \ref{tp_1:fig_2a} shows the singular values for the values $k=25$ selected for applying POD for the test problem 1). 
        
        Selecting $k = 25$, and running the ROM Saving Submodule, we save the reduced basis and assemble the relevant reduced operators, including their storage in the database through the connection between the Offline Module in the Inference Engine Layer and the database in the Knowledge-Base Layer. Moreover, the Submodule also can perform the error analysis to investigate and a-posteriori estimation of the accuracy of the model when a different amount of basis function is employed for the dimensionality reduction. %As shown in Figure \ref{tp_1:fig_2b}, the error has an exponential trend, indicating great performance even for low-dimensional basis.

        \begin{figure}[!ht]
    \centering
    % Prima immagine
    \subfigure[Singular Values Decay]{
        \includegraphics[width=0.4\textwidth]{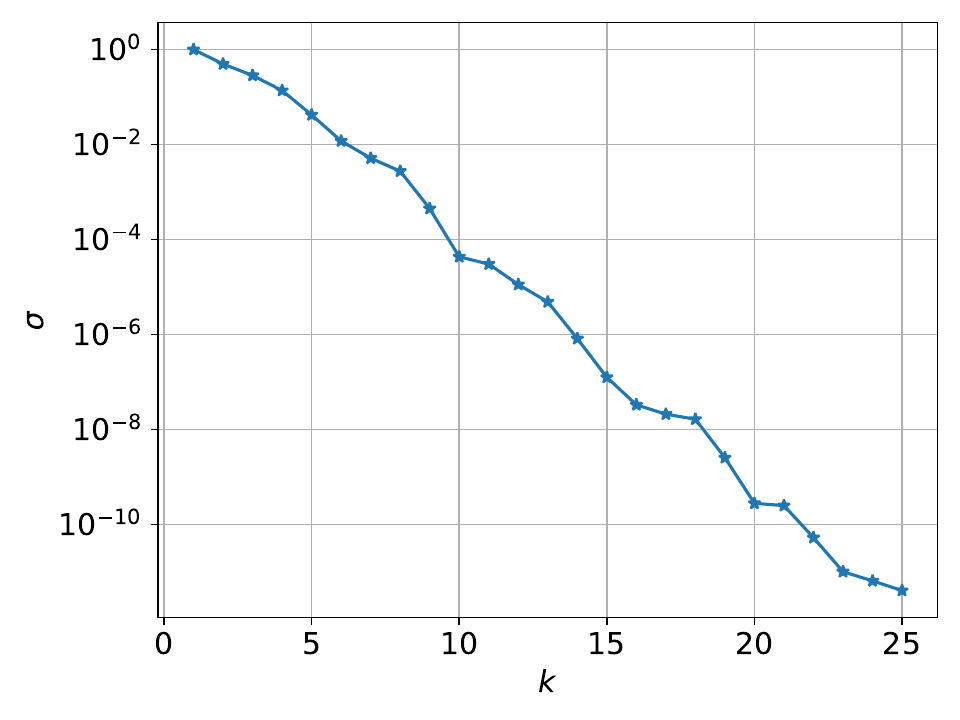}
        \label{tp_1:fig_2a}
    }
    % Seconda immagine
    \subfigure[Error Analysis]{
        \includegraphics[width=0.4\textwidth]{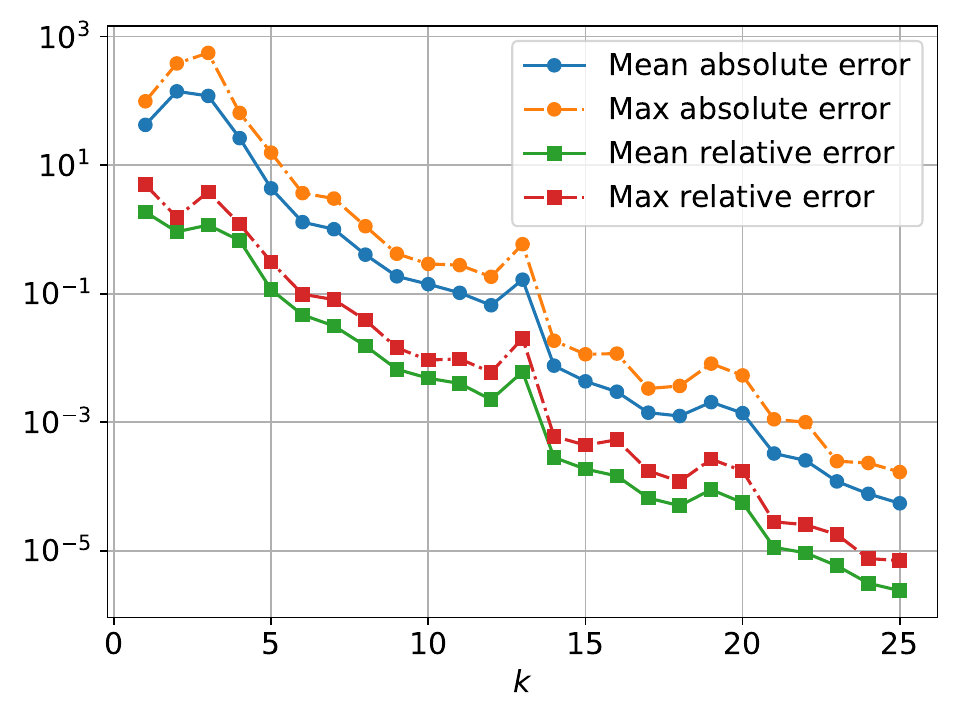}
        \label{tp_1:fig_2b}
    }
    \caption{Decay of the singular values and error analysis, respectively left and right, obtained from the ROM Submodule for the Poisson equation in Test Problem 1.}
    \label{tp_1:fig_2}
\end{figure}
% \fp{Qui dobbiamo aumentare il font di tutte le scritte, cambiare order e N con k, normalizzare gli autovalori rispetto al primo $\lambda_0$, levare eigenvalues sopra, cambiare label per errori e definirli nel testo (max e mean, abs e rel)}\fp{Specifichiamo dove testiamo per analisi dell'errore, mentre non mi ricordo come avevamo giustficato il fatto che l'errore in alcuni casi crescesse se aumentiamo le basi}

        We show in Figure \ref{tp_1:fig_2} the behavior of the singular values and the errors on the selected modes.
	    In particular, Figure \ref{tp_1:fig_2a} describes the decay of the normalized singular values for the snapshots' matrix $S$.
		In Figure \ref{tp_1:fig_2b}, we depict the performance of the model while varying the amount of basis in the reduced order expansion. The plot shows the mean and maximum absolute and relative $L^2$ errors w.r.t.\ the full-order solution for 10 randomly sampled snapshots.
		The validity of ROM assumption is validated by the exponential trend of the error, indicating excellent performance even for a low-dimensional basis.
		% However, the figure shows some noise for the modes 13, 16, and 17, where the errors increase instead of decrease, underlying some instability related to the mesh and the analysed problem.

        When sensor information are available, one could be interested in the identification of a specific tuple of parameter $\left( \lambda, \alpha, \beta \right)$ that are related to the observed data.
        % require the collection fo data related to the phenomenon. 
        Specifically, we exploit (surrogate) simulated data obtained by choosing the potentially unknown parameter sample $\left( \lambda, \alpha, \beta \right) = \left( 0.1, 0.2, 0.5 \right)$ to discover the properties of the physical model.

\begin{figure}[!ht]
    \centering
    \includegraphics[width=.55\linewidth]{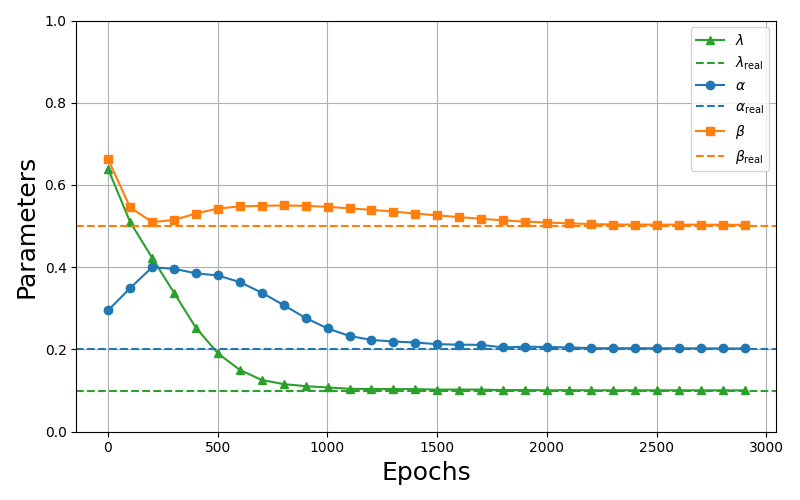} % Sostituisci con il percorso della tua immagine
    \caption{Convergence of $\lambda, \alpha, \beta$ during training to the real values $\lambda_\mathrm{real}, \alpha_\mathrm{real}, \beta_\mathrm{real}$.}
    \label{tp_1:fig3}
\end{figure}

        Indeed, the parameter identification task requires solving an inverse problem through the employment of Physics-Informed Neural Networks. The PINN solver selected is the Residual-based Attention PINN \cite{Anagnostopoulos2024}, based on the Residual Feed Forward neural network \cite{Wang20213055}.
        We report in Table \ref{tp_1:tab_1} the details of the architecture exploited. Concerning the sampling points, we considered 200 collocation points inside the domain $\Omega$, 50 boundary points to enforce the boundary conditions finally, and 500 data points representing the data collected through simulated sensors on the boundary of the rock.
        During the training, the Inverse Problem Submodule of the PINN Module learns an approximation of the physical parameters, obtaining a great relative accuracy as depicted in Figure \ref{tp_1:fig3}, showing the convergence during the training process to the unknown parameter value to be identified, and with quantitative metric reported in Table \ref{tp_1:tab2}.

        \begin{table}[!ht]
    \centering
    \caption{PINN hyperparameters for Test Problem 1.}
    \begin{tabular}{l|c}
        {\bf Architecture}       & {\bf Value} \\
        \toprule
        Collocation Points & 200 \\
        Boundary Points    & 50 \\
        Data Points        & 500 \\
        Epochs             & 3000 \\
        Batch size         & 50 \\
        Learning rate      & $1e-03$ \\
        Decay rate         & $1e-08$ \\
        Optimizer          & Adam \\
        Network structure  & $\left[3, 400, 400, 1 \right]$ \\ \bottomrule
    \end{tabular}
    \label{tp_1:tab_1}
\end{table}
% \fp{Quando citiamo la tabella chiariamo la differenza tra collocation, boundary e data points? Scriviamo anche quanti punti sono in tutto la FEM, e quale tipo di elementi polinomiali (non in tabella ma nel testo).}
        
% \fp{Cambiare epoche con epochs, e label mettendo il formato latex, aumentare in font ovunque , aggiungerei delle linee tratteggiate orizzontali in corrispondenza dei valori reali. Metterei questa tabella accanto alla tabella due con minipage.}

        \begin{table}[!ht]
	\centering
	\caption{Results obtained exploiting PINN for the inverse problem. The Table reports the approximated and the expected parameters, and the relative errors.}
	\begin{tabular}{c|c|c|c}
		\textbf{Parameter} & \textbf{Approximated Value} & \textbf{Expected Value} & \textbf{Relative Error} \\ \toprule
		$\lambda$          & 0.1006                      & 0.1000                  & 5.8391e-03              \\
		$\alpha$           & 0.2024                      & 0.2000                  & 1.2022e-02              \\
		$\beta$            & 0.5032                      & 0.5000                  & 6.3606e-03              \\ \bottomrule
	\end{tabular}
	\label{tp_1:tab2}
\end{table}

        Finally, the Online Module load the reduction method through the ROM Loading Submodule for the identified parameter value and solve the analyzed problem by exploiting the Online Solver Submodule.

        The \textsf{msh2xdmf} Module allows the construction of the $\textsf{xdmf}$ file supported by a \textsf{h5} file, enabling the visualization and inspection of the results.
        Indeed, the module provide the files to the Simulation Module of the Application Layer to visualize the simulation of the phenomenon. Moreover, the \textsf{msh2xdmf} modul provide the error comparing the reduced solution obtained through the Online Solver Submodule with the full order one obtained through the FEM Module.

\begin{figure}[!ht]
    \centering
    % Prima immagine
    \subfigure[Boundary - Front view]{
        \includegraphics[width=.45\textwidth]{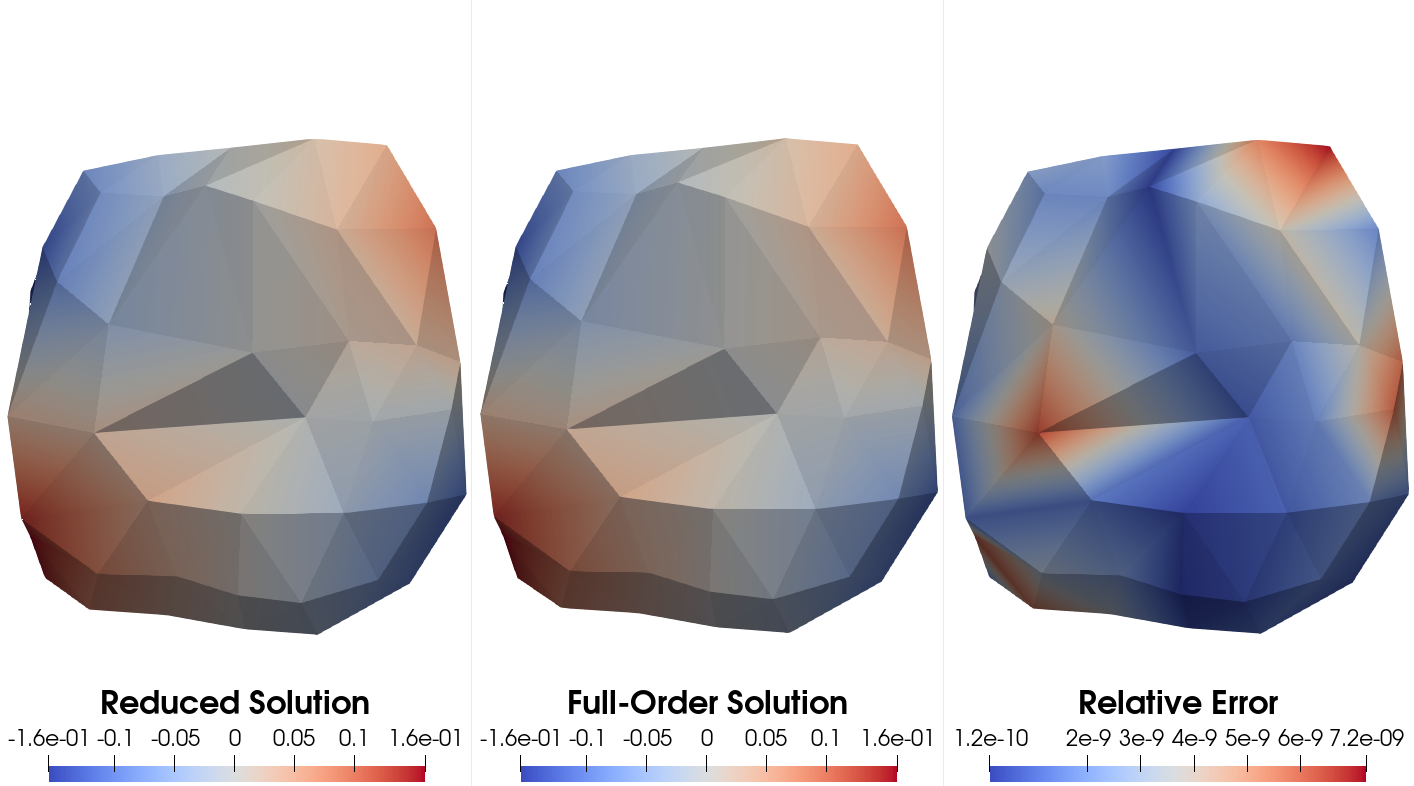}
    }
    \subfigure[Boundary - Isometric View]{
        \includegraphics[width=.45\textwidth]{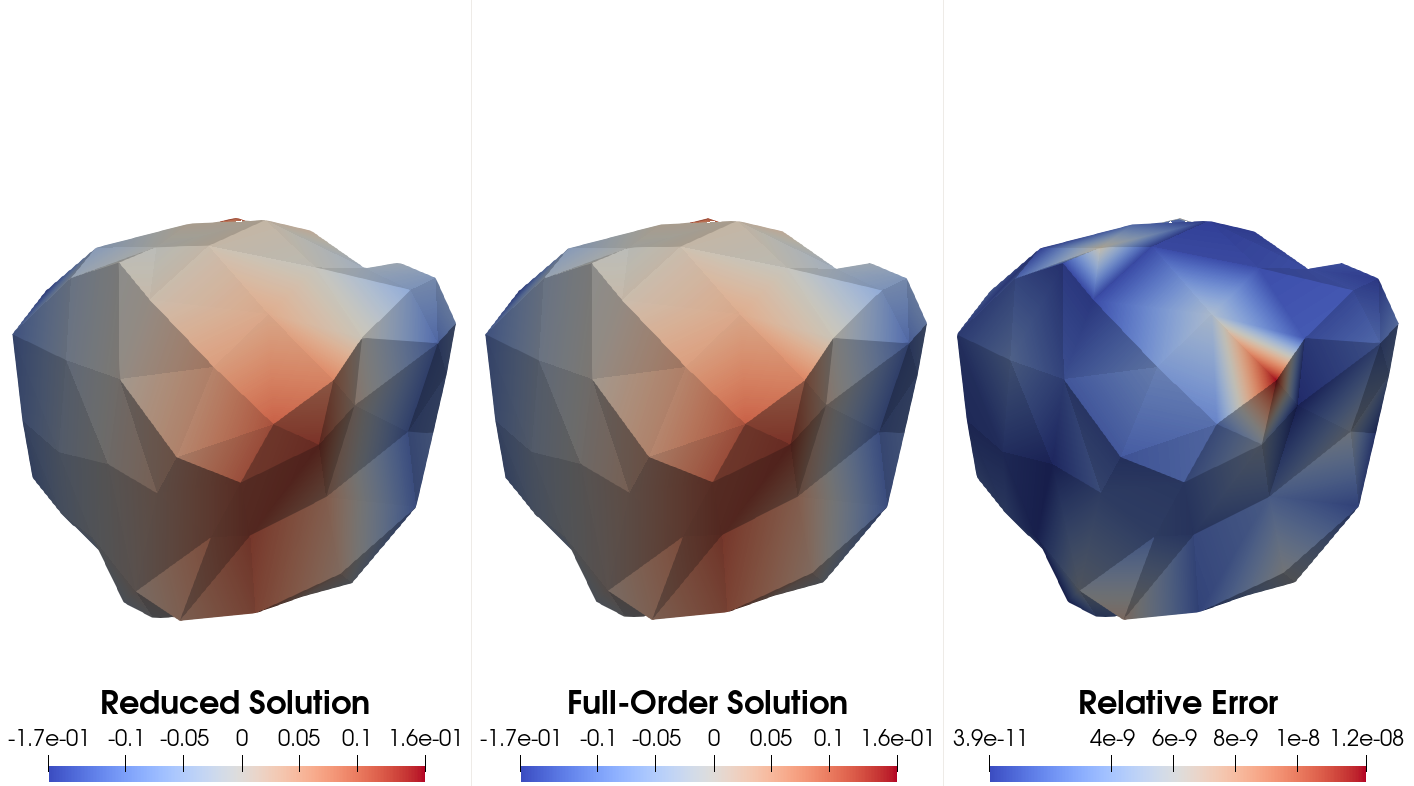}
    }
    
    % Seconda immagine
    \subfigure[Slice - Front view]{
        \includegraphics[width=.45\textwidth]{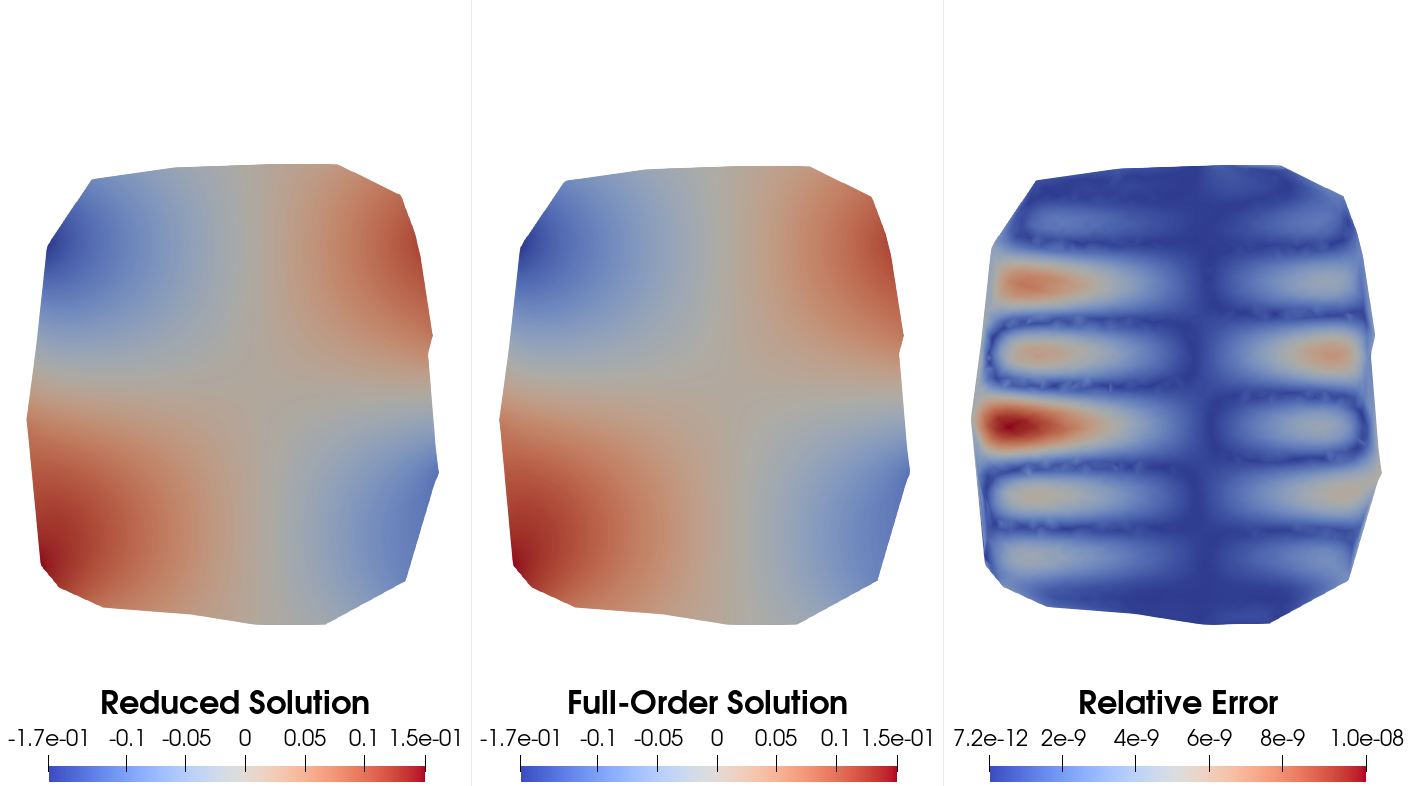}
    }
    % Seconda immagine
    \subfigure[Slice - Isometric View]{
        \includegraphics[width=.45\textwidth]{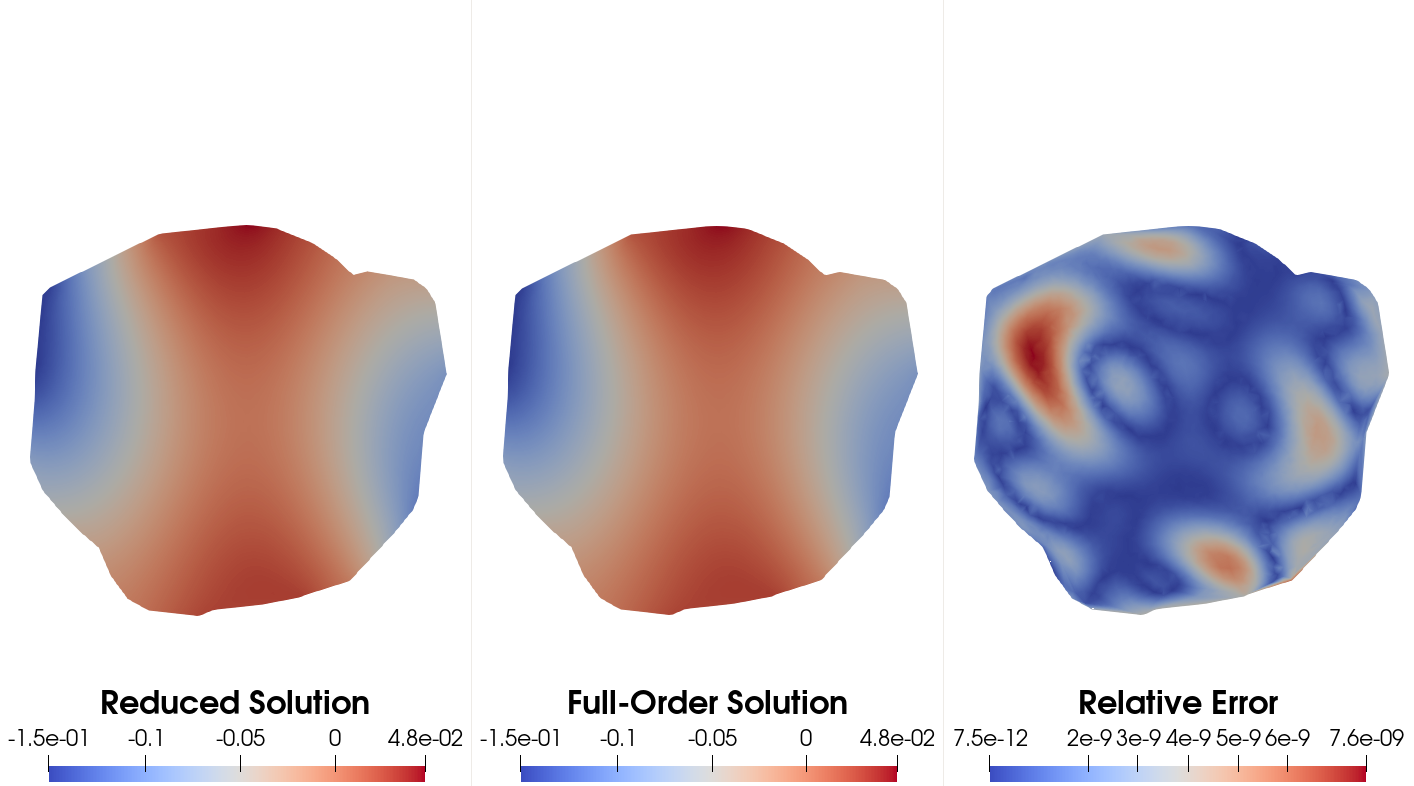}
    }
    
    \caption{Reduced approximation, full order solution, and relative error (left, middle, and right respectively) for Test Problem 1 on the boundary and on a slice with different views.}
    \label{tp_1:fig4}
\end{figure}

        We show in Figure \ref{tp_1:fig4} the reduced solution, the full order solution, and the relative error on boundary and on a slice of the domain.

% \fp{Qui dobbiamo estrarre solo il campo con background trasparente, tenere stessa camera/grandezza per i tre campi in tre immagini separate, aumentare il font della colorbar e rimuovere max min sotto. Lo slice dice poco se tenuto nello stessa scala colorbar, forse meglio riscalare rispetto al campo visualizzato. Poi aggiustiamo la caption di conseguenza}

        %In particular, the reduced solution is obtained considering the approximation of parameters obtained through the Inverse Problem Submodule, instead the full order solution is evaluated on the effective parameters $\left( \lambda, \alpha, \beta \right) = \left( 0.1, 0.2, 0.5 \right)$.
        In particular, the reduced solution is computed using the approximated parameters obtained from the Inverse Problem Submodule, rather than the exact ones $\left( \lambda, \alpha, \beta \right) = \left( 0.1, 0.2, 0.5 \right)$. This choice is aimed at evaluating the robustness of the entire framework, not only the ability of the physics-informed inverse model to infer the governing parameters reliably, and of the reduced-order solver to reconstruct the physical state through the online phase, but also the overall capability of the methodology to detect and represent real-world phenomena accurately.
        % \fp{questa non è la cosa classica quindi rimarchiamolo e diciamo che è fatto apposta per dire la robustezza del problema inverso che sfrutta la fisica per trovare il parametro e poi quello del solutore ridotto che sfrutta la fisica per trovare lo stato completo tramite il problema diretto}
         
        Finally, the total relative error introduced with the FE approximation and the subsequent reduction approach with respect to the analytical solution in \eqref{tp_1:eq1} is only $7.77e-3$, and it is mostly coming from the first discretization itself relative error with respect the full-order solution is $4.56e-7$. We remark that more accurate high-fidelity solutions can be obtained by refining the original mesh obtained from the 3D model, or choosing higher order spaces for the polynomials. This comes at the cost of a more costly offline phase, but the online phase remains unaffected, still providing reliable and real-time approximations.

    \subsubsection{Test Problem 2: Parabolic problem on a column}\label{tp_2:}
    
        As a second benchmark, we study the parametric  solution of a vector parabolic PDE defined on the column domain $\Omega \subseteq \mathbb{R}^3$ depicted in Figure \ref{pre:fig_2}.
        % \fp{qui applicare le stesse correzioni anche del problema 1, riguardo testo e immagini}
        %  with boundary $\partial\Omega$, for analysing the following problem consisting of a system of 
        The system of equations governing the coupled heat equation with time-dependent sources is given by:
        \begin{equation}\label{tp_2:eq1}
            \mathbf{u}_t \left( x, y, z, t \right) = \Delta \mathbf{u} \left( x, y, z, t \right) + F \left( t \right), \qquad  \text{in}\ \Omega \times \left[ 0,1 \right]
        \end{equation}
        % \fp{se questo è vettoriale magari ha senso mettere tutte le quantità in grassetto, spieghiamo cosa rappresenta u, è perché è diversa dal campo scalare prima. Sosituirei x,y,z con x grassetto ovunque}
        where $\mathbf{u} = [u^{\left(1\right)},u^{\left(2\right)}]^T \in \mathbb{R}^2$
        represents both components of the linear parabolic system \eqref{tp_2:eq1}, and
        \begin{equation}\label{tp_2:eq1a}
            F \left(t \right) = \left(
                \begin{array}{c}
                    \lambda e^{\lambda t} \\
                    \lambda e^{\lambda t} - 2 \left( \alpha + \beta + 1\right)
                \end{array}
            \right) , \qquad \text{in}\ \Omega \times \left[0, 1 \right],
        \end{equation}
        where $\lambda$ is the temporal growth rate of the source, while $\alpha$ and $\beta$ are the parameters related to the spatial variation of the solution.
        By imposing the following boundary conditions
        \begin{equation}\label{tp_2:eq1b}
            \mathbf{u}_b \left( x, y, z, t \right) = \left(
                \begin{array}{c}
                    e^{\lambda t} + \alpha x + \beta y + z \\
                    e^{\lambda t} + \alpha x^2 + \beta y^2 + z^2
                \end{array}
            \right) , \qquad \text{on}\ \partial\Omega \times \left[0, 1 \right]
        \end{equation}
        and initial conditions
        \begin{equation}\label{tp_2:eq1c}
            \mathbf{u} \left( x, y, z, 0 \right) = \mathbf{u}_0 \left( x, y, z \right) = \left(
                \begin{array}{c}
                    1 + \alpha x + \beta y + z \\
                    1 + \alpha x^2 + \beta y^2 + z^2
                \end{array}
            \right), \qquad \text{in}\ \Omega
        \end{equation}
        the analytical solution of the problem is given by
        \begin{equation}\label{tp_2:eq2}
            \mathbf{u} \left( x, y, z, t \right) = \left(
                \begin{array}{c}
                    u^{\left(1\right)}\\
                    u^{\left(2\right)}
                \end{array}
            \right) = \left(
                \begin{array}{c}
                    e^{\lambda t} + \alpha x + \beta y + z \\
                    e^{\lambda t} + \alpha x^2 + \beta y^2 + z^2
                \end{array}
            \right) \in \mathbb{R}^2, \qquad \text{in}\ \Omega \times \left[0, 1 \right] .
        \end{equation}

        Test Problem 2 further extend the previous setting on a more complex geometry and in the case of a system of time-dependent PDEs. As before, we aim to apply the POD procedure, identify the parameters that fit the simulated data with PINN, and compare the reduced approximation with the full-order solution.
        
        First we perform the semi-discretization projecting the problem via FE method in space and considering time dependent coefficients for the standard Galerkin approach, and obtain the weak formulation of \eqref{tp_2:eq1} as

        \begin{equation}\label{tp_2:eq6}
            m \left( \mathbf{u(t)}, v; t \right) + a \left( \mathbf{u(t)}, v; t \right) = L \left( v; t \right), \qquad \forall v \in V ,
        \end{equation}
        where
        \begin{equation}\label{tp_2:eq7}
            m \left( \mathbf{u}(t), v; t \right) = \int_\Omega \dot{\mathbf{u}}(t) v\, \mathrm{d}x , \qquad a \left( \mathbf{u}(t), v; t \right) = \int_\Omega \nabla \mathbf{u}(t) \cdot \nabla v\, \mathrm{d}x , \qquad  L \left(v; t \right) = \int_\Omega F \left( t \right) v\, \mathrm{d}x .
        \end{equation}
        % \begin{equation}\label{tp_2:eq8}
        %     L_{n+1} = \int_\Omega \left( u_n + h F \left( x, y, z, t_{n+1}, u_{n+1} \left( x, y, z \right) \right) \right) v \mathrm{d}x, \qquad n=0, \dots, N-1 
        % \end{equation}
        
        To deal with the fact that the discrete problem is unsteady we exploited a Finite Difference scheme, namely the implicit Euler method, defining the temporal grid as
        \begin{equation}\label{tp_2:eq3}
            0 = t_0 < t_1 < \cdots < t_{N-1} < t_N = 1, \qquad t_i = h i, \quad i=0,\dots, N, \qquad h = \frac{1}{N} .
        \end{equation}
        In this way, by defining $\mathbf{u}_{n}(x,y,z) = \mathbf{u}(x,y,z,t_{n}) \in \mathbb{R}^2$ and approximating $\dot{\mathbf{u}}(t) = \frac{\mathbf{u}_{n+1} - \mathbf{u}_n}{h}$, we can express the iteration in time via the Euler method as

        % \begin{equation}\label{tp_2:eq4}
        %     \mathbf{u}_{n+1} \left( x, y, z \right) = \mathbf{u}_n \left( x, y, z \right) + h f_{n+1} \left( x, y, z, t_{n+1}, \mathbf{u}_{n+1} \right) , \qquad n=0, \dots, N-1 ,
        % \end{equation}

        % where
        % \begin{equation}\label{tp_2:eq4a}
        %     f_{n+1} \left( x, y, z, t_{n+1}, \mathbf{u}_{n+1} \right) = \Delta \mathbf{u}_{n+1} \left( x, y, z \right) + F \left(t_{n+1} \right) .
        % \end{equation}

        % Finally, the system to be solved at each time step reads
        \begin{equation}\label{tp_2:eq5}
            M \frac{\mathbf{u}_{n+1} - \mathbf{u}_{n}}{h} + A \mathbf{u}_{n+1} = \mathbf{f}_{n+1}, \qquad n=0, \dots, N-1 ,
        \end{equation}    
        where we denoted the mass matrix $M$, the stiffness matrix $A$, and the load vector $\mathbf{f}_{n+1}$ at time $t_{n+1}$, from the assembled bilinear and linear forms in Equation \eqref{tp_2:eq7}.

        In the time-dependent context, applying the POD procedure requires managing the temporal dependence of the problem. Having discretized the temporal variable according to \eqref{tp_2:eq3}, time can be regarded as an additional (special) parameter for which to perform separate reduction through a nested application of the POD.

\begin{figure}[!ht]
    \centering
    % Prima immagine
    \subfigure[Singular Values Decay]{
        \includegraphics[width=0.4\textwidth]{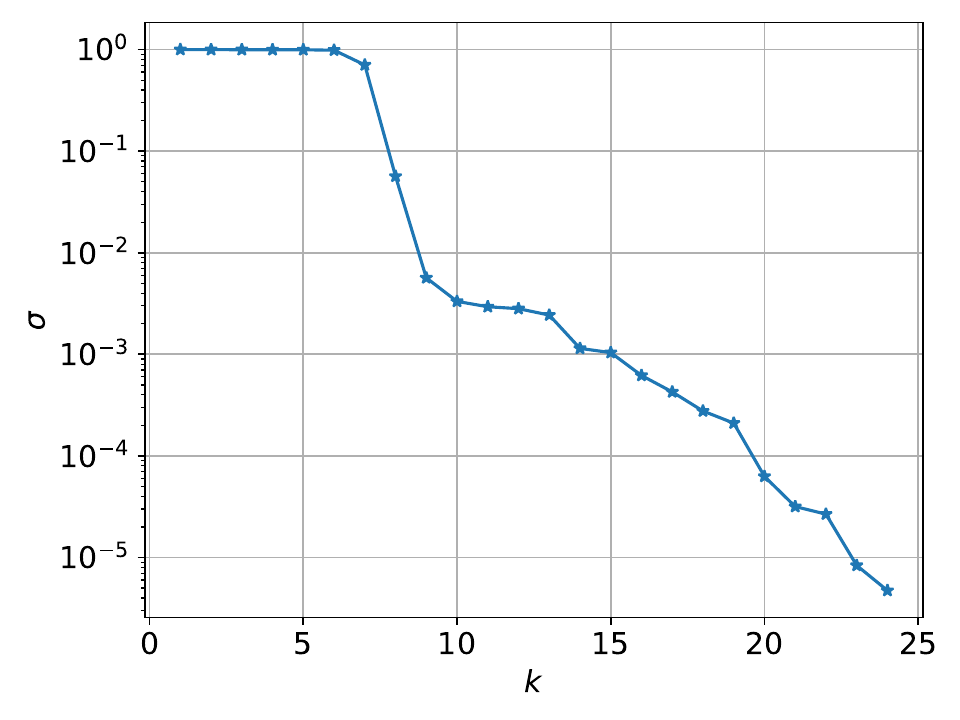}
        \label{tp_2:fig_2a}
    }
    % Seconda immagine
    \subfigure[Error Analysis]{
        \includegraphics[width=0.4\textwidth]{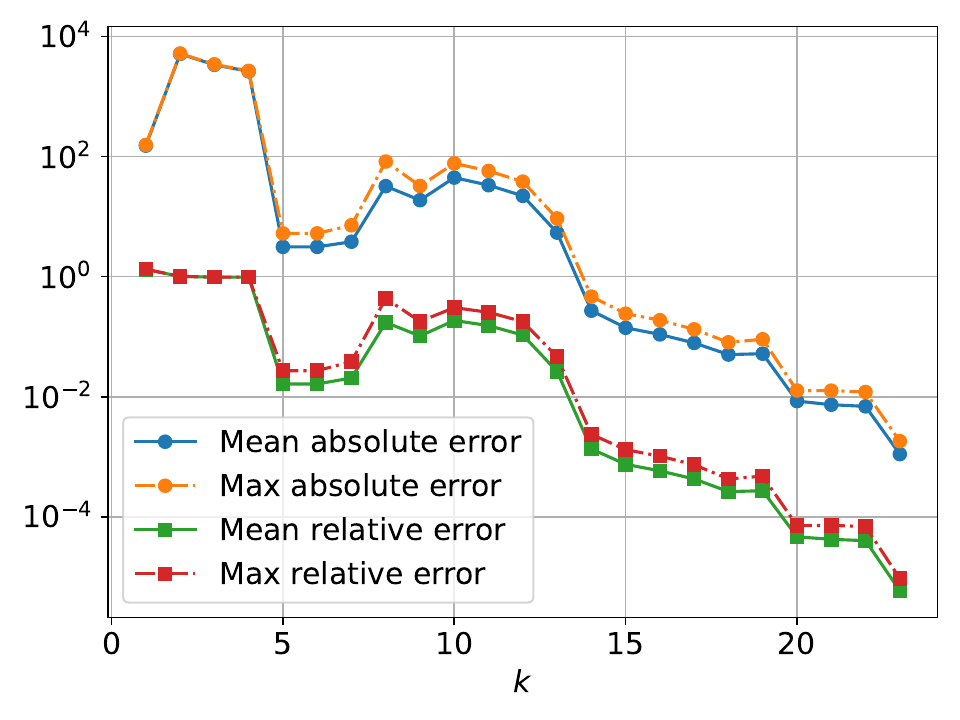}
        \label{tp_2:fig_2b}
    }
    \caption{Decay of the singular values and error analysis, respectively left and right, obtained from the ROM Submodule for the Parabolic system in Test Problem 2.}
    \label{tp_2:fig_2}
\end{figure}

        As before, we compute the snapshots using $\mathbb{P}_1$ FE space with $N_h = \num{12375}$ nodes, $M=20$ parameter snapshots, $T=1$, and $N = 21$, and we store the reduced quantities exploiting the ROM Saving Submodule with $k=24$. Figure \ref{tp_2:fig_2} shows the singular values obtained by imposing a tolerance of $1e-6$ (Figure \ref{tp_2:fig_2a}) and the error analysis when testing the reduced model on a set of 5 randomly sampled testing snapshots (Figure \ref{tp_2:fig_2b}).

% \fp{anche qui non ricordo bene il perché dei risultati e del decay degli autovalori, in caso va giustificato bene nel testo.}\fp{remark per me di modificare le caption alla fine come in TP1}

        Figure \ref{tp_2:fig_2} depicts the decay of the singular values and the reduced error w.r.t.\ the number of modes. Specifically, we observe the initial difficulty of the model in compressing the information, given by the small amount of parametric samples in a three-dimensional space, and the additional complexity due to time. 
        Despite this, as we can see from Figure \ref{tp_2:fig_2b}, describing the behavior of the mean and maximum absolute/relative errors, the reduction strategy keeps improving with the exponential rate eventually reaching a great accuracy for the surrogate model.

        Following the same workflow as for Test Problem 1, we proceed with the identification of the parameters $\left( \lambda, \alpha, \beta \right)$ related to simulated data obtained by choosing the potentially unknown parameter sample $\left( \lambda, \alpha, \beta \right) = \left( 0.1, 0.2, 0.5 \right)$.
        Table \ref{tp_2:tab_1} describes the hyperparameters of the networks, where the necessity of facing a system of time-dependent PDEs requires the employment of a greater number of collocation and boundary points. In addition, the PINN Module also requires points that reduce the component of the loss function related to the initial conditions, and consequently the amount of simulated data and the number of epochs. Properly choosing the ``right'' hyperparameter is always very difficult, for this reason we decided to test our strategy by fixing the network's structure, of course except for the input and output layers, and observe the robustness of the procedure.
        
        \begin{table}[!ht]
    \centering
    \caption{PINN hyperparameters for Test Problem 2.}
    \begin{tabular}{l|c}
        {\bf Detail}       & {\bf Value} \\
        \hline
        Collocation points & 1000 \\
        Boundary points    & 400 \\
        Initial points     & 400 \\
        Data points        & 1000 \\
        Epochs             & 10000 \\
        Batch size         & - \\
        Learning rate      & $5e-04$ \\
        Decay rate         & $1e-08$ \\
        Optimizer          & Adam \\
        Network structure  & $\left[4, 400, 400, 2 \right]$
    \end{tabular}
    \label{tp_2:tab_1}
\end{table}

\begin{figure}[!ht]
    \centering
    \includegraphics[width=.55\linewidth]{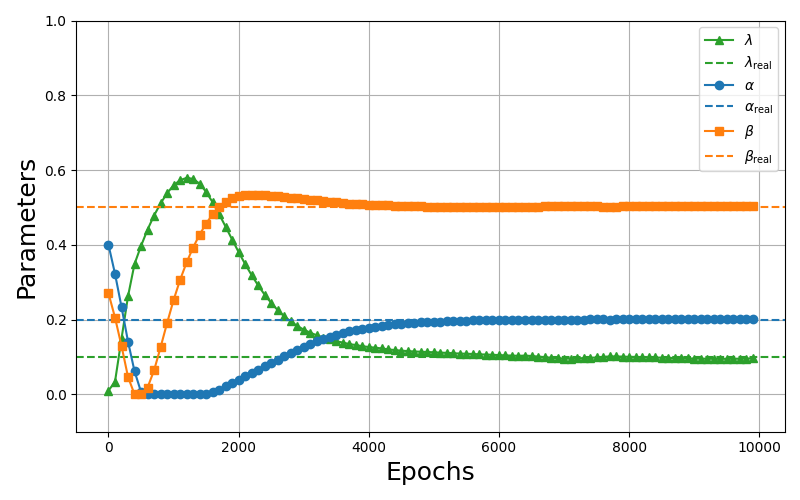} % Sostituisci con il percorso della tua immagine
    \caption{Convergence of $\lambda, \alpha, \beta$ during training to the real values $\lambda_\mathrm{real}, \alpha_\mathrm{real}, \beta_\mathrm{real}$.}
    \label{tp_2:fig3}
\end{figure}

        We report in Table \ref{tp_2:tab2} the results of exploiting the Inverse Problem Submodule for the identification of the parameters. Specifically, the submodule achieves a relative error of order $10^{-2}$ for parameter $\lambda$ and $10^{-3}$ for parameters $\alpha$ and $\beta$. Additionally, Figure \ref{tp_2:fig3} describes the convergence of PINN in identifying the parameters through the epochs, correctly identifying the real values of the simulated data already at $5000$ epochs.

        %Figure \ref{tp_2:fig3} shows how the PINN can identify the parameters starting from the $5000$ epoch and the approximation estimation shown in Table \ref{tp_2:tab2} describe the relative error of order $10^{-2}$ for parameter $\lambda$ and $10^{-3}$ for parameters $\alpha$ and $\beta$.

        \begin{table}[!ht]
	\centering
	\caption{Results obtained exploiting PINN for the inverse problem. The Table reports the approximated and the expected parameters, and the relative errors.}
	\begin{tabular}{c|c|c|c}
		\textbf{Parameter} & \textbf{Approximated Value} & \textbf{Expected Value} & \textbf{Relative Error} \\ \toprule
		$\lambda$          & 0.0960                      & 0.1000                  & 4.0492e-02              \\
		$\alpha$           & 0.2014                      & 0.2000                  & 7.1611e-03              \\
		$\beta$            & 0.5026                      & 0.5000                  & 5.1683e-03              \\ \bottomrule
	\end{tabular}
	\label{tp_2:tab2}
\end{table}

% \fp{in questa figura, e anche le altre per evoluzione dei parametri, stiamo fissando $y_min$ e $y_max$ ma forse possiamo tenerli liberi in modo che il grafico sia più leggibile senza essere per metà vuoto}
    
        \begin{figure}[!htp]
    \centering
    % Prima immagine
    \subfigure[Boundary - $T=0.5$ - Front view]{
        \includegraphics[width=.42\textwidth]{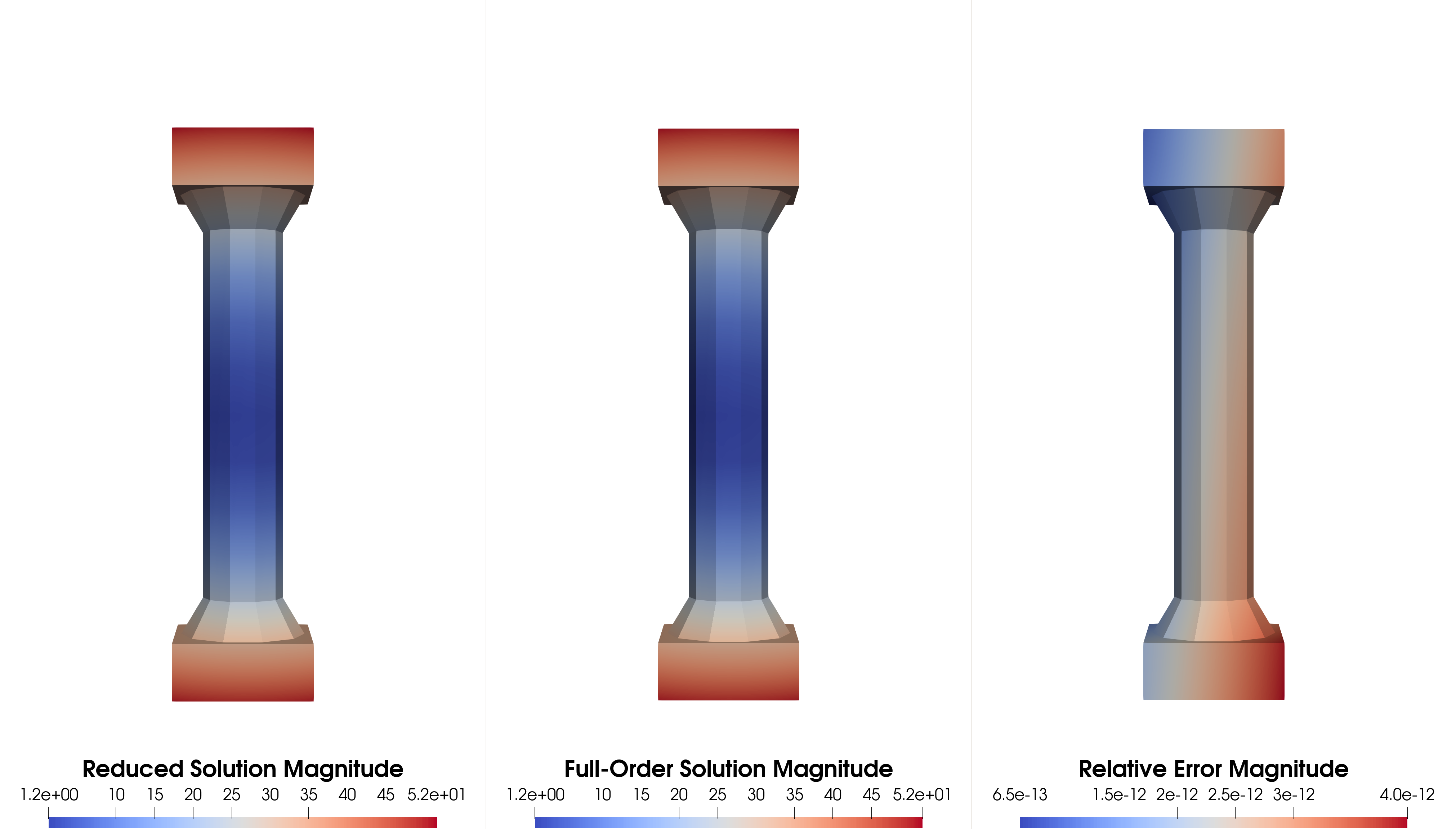}
    }
    \subfigure[Boundary - $T=0.5$ - Isometric View]{
        \includegraphics[width=.42\textwidth]{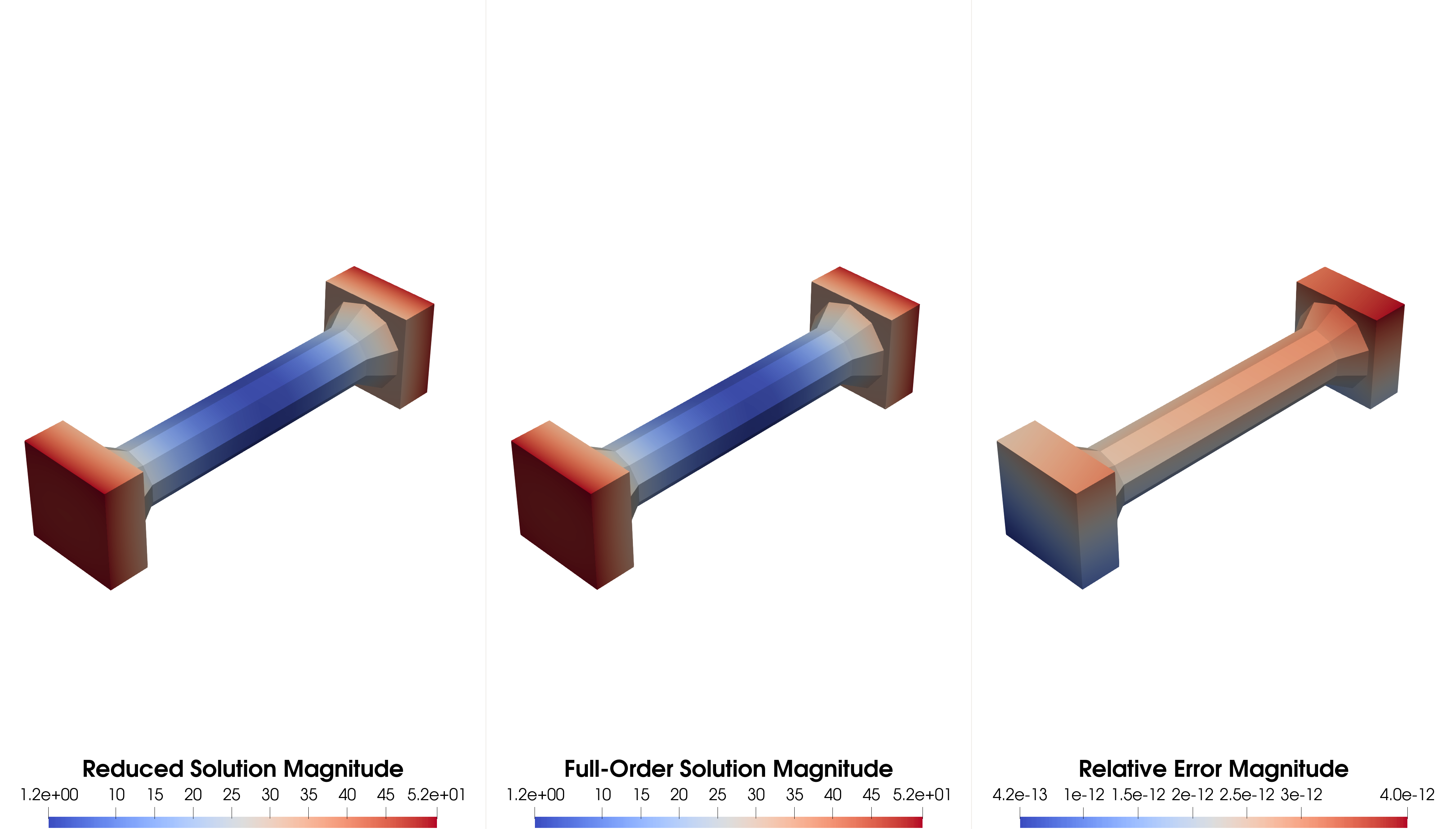}
    }
    
    % Seconda immagine
    \subfigure[Slice - $T=0.5$ - Front view]{
        \includegraphics[width=.42\textwidth]{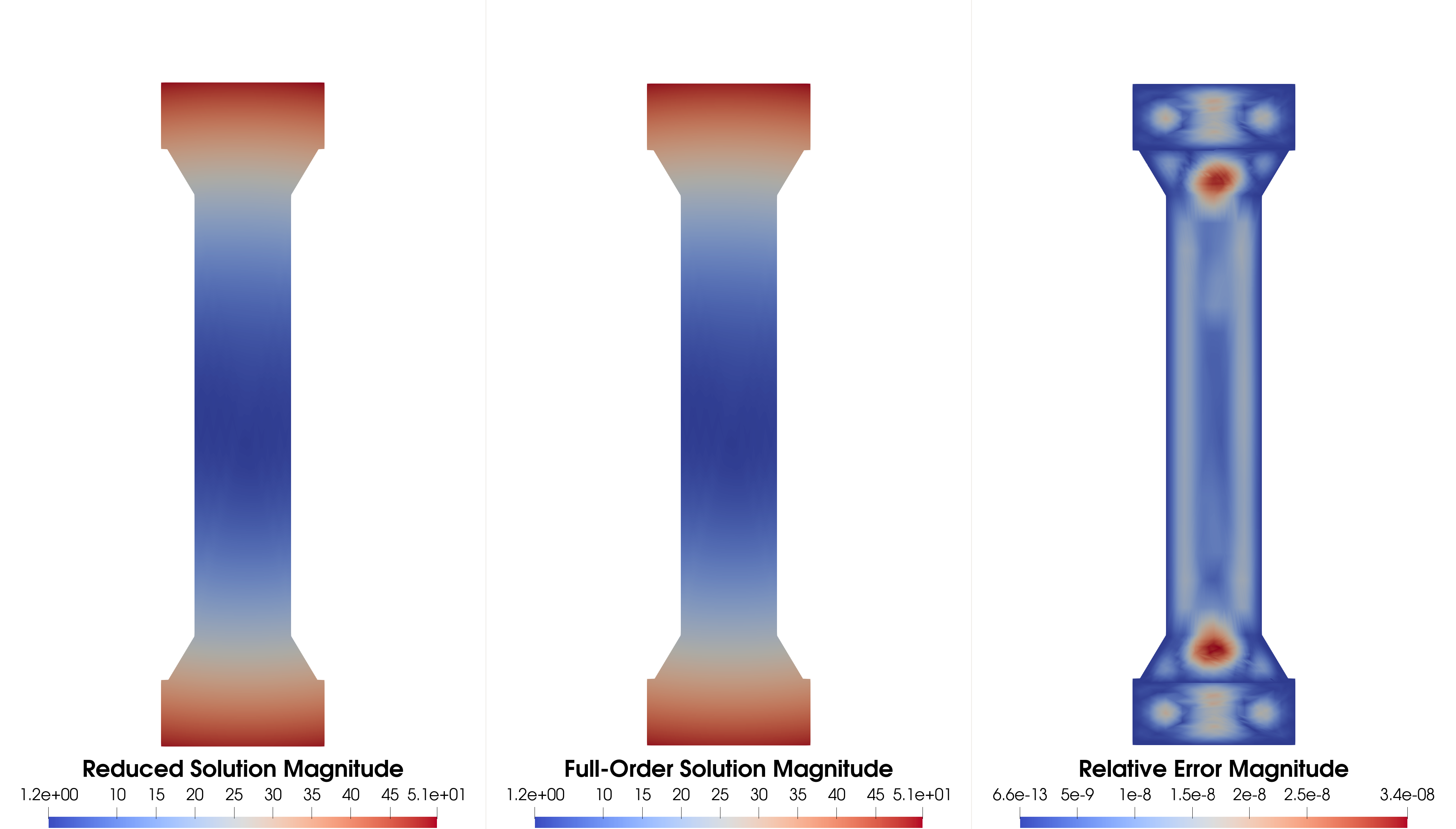}
    }
    % Seconda immagine
    \subfigure[Slice - $T=0.5$ - Isometric View]{
        \includegraphics[width=.42\textwidth]{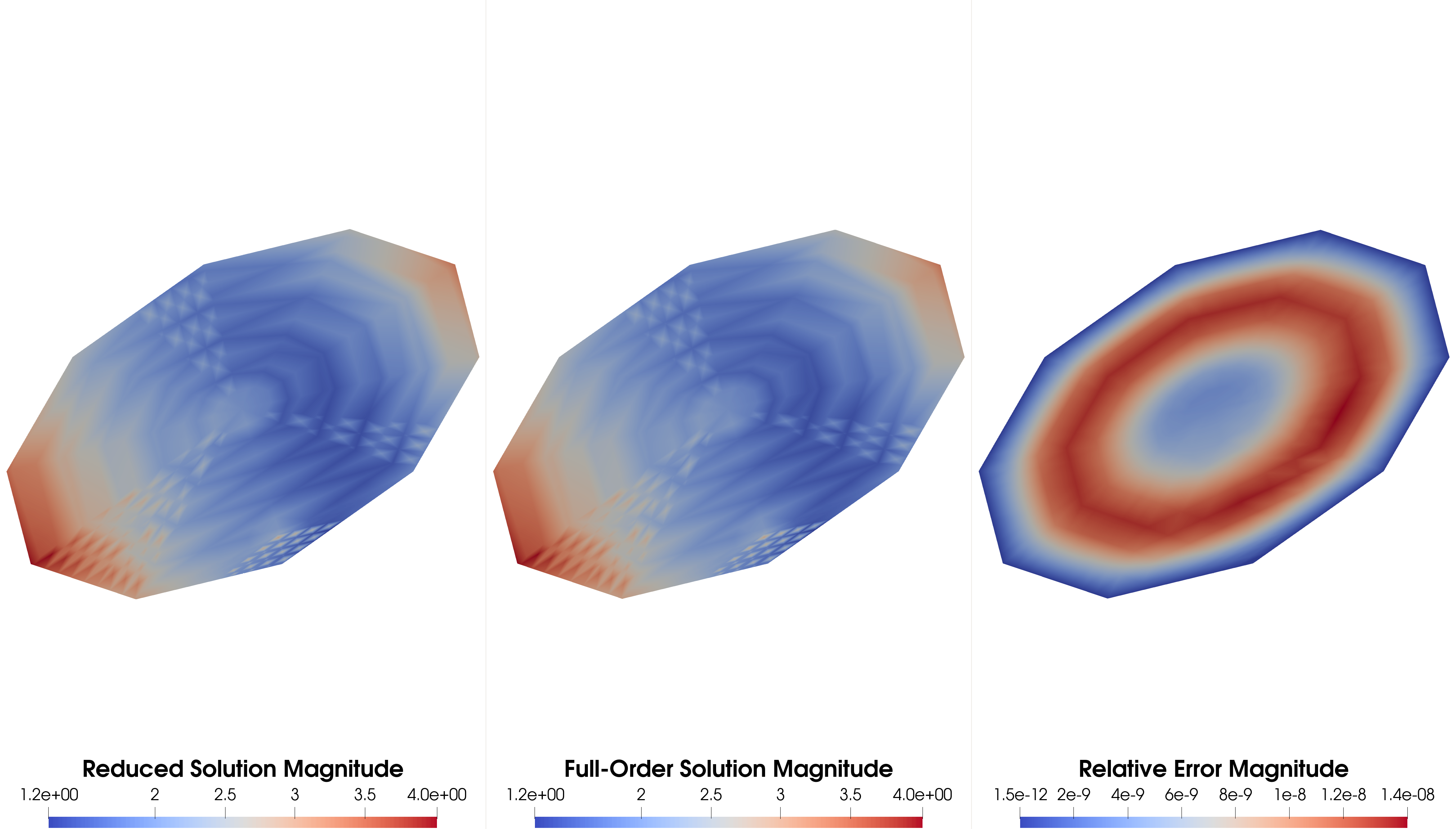}
    }

    % Terza immagine
    \subfigure[Boundary - $T=1.0$ - Front view]{
        \includegraphics[width=.42\textwidth]{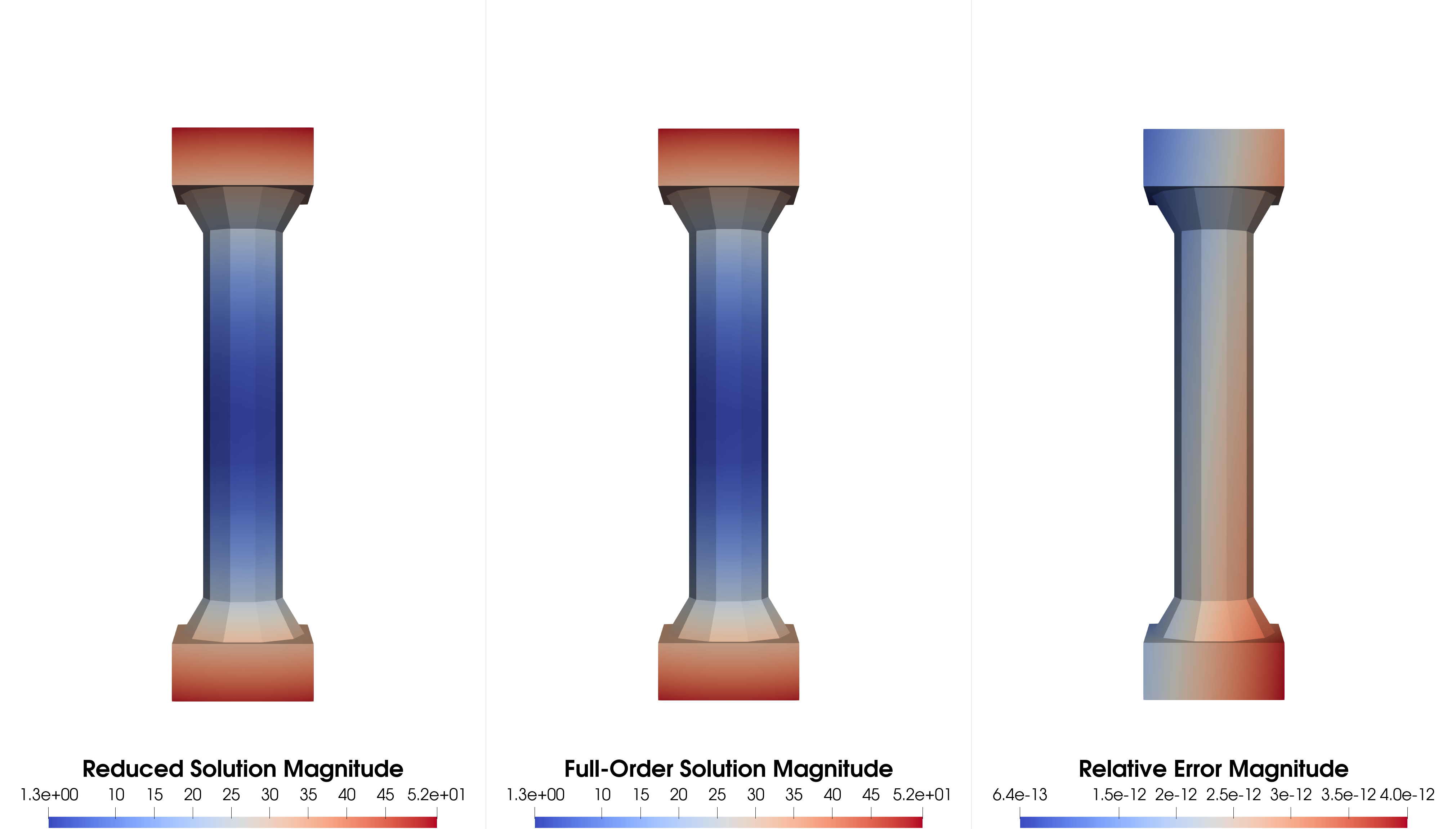}
    }
    % Terza immagine
    \subfigure[Boundary - $T=1.0$ - Isometric View]{
        \includegraphics[width=.42\textwidth]{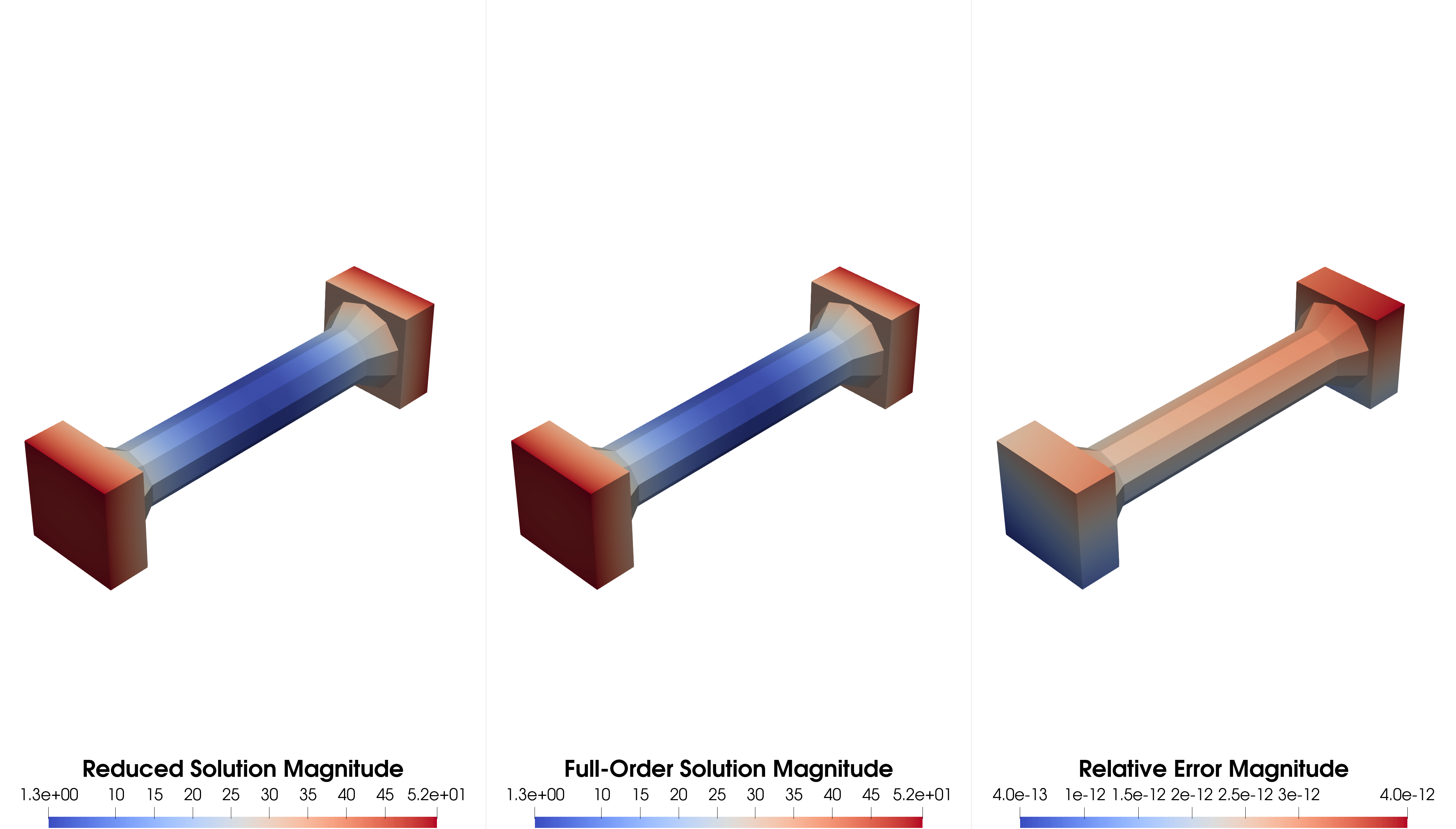}
    }

    % Quarta immagine
    \subfigure[Slice - $T=1.0$ - Front view]{
        \includegraphics[width=.42\textwidth]{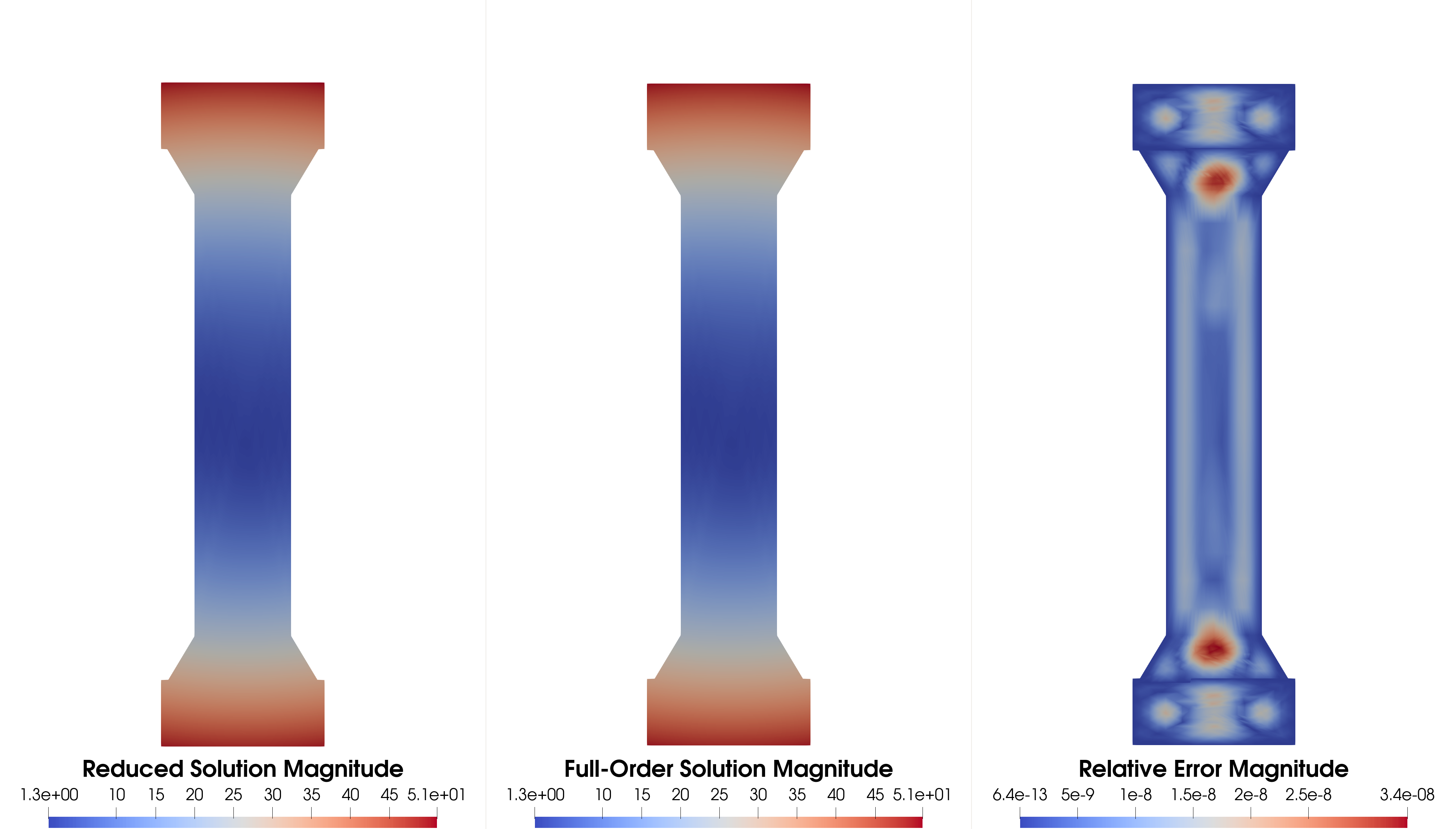}
    }
    % Quarta immagine
    \subfigure[Slice - $T=1.0$ - Isometric View]{
        \includegraphics[width=.42\textwidth]{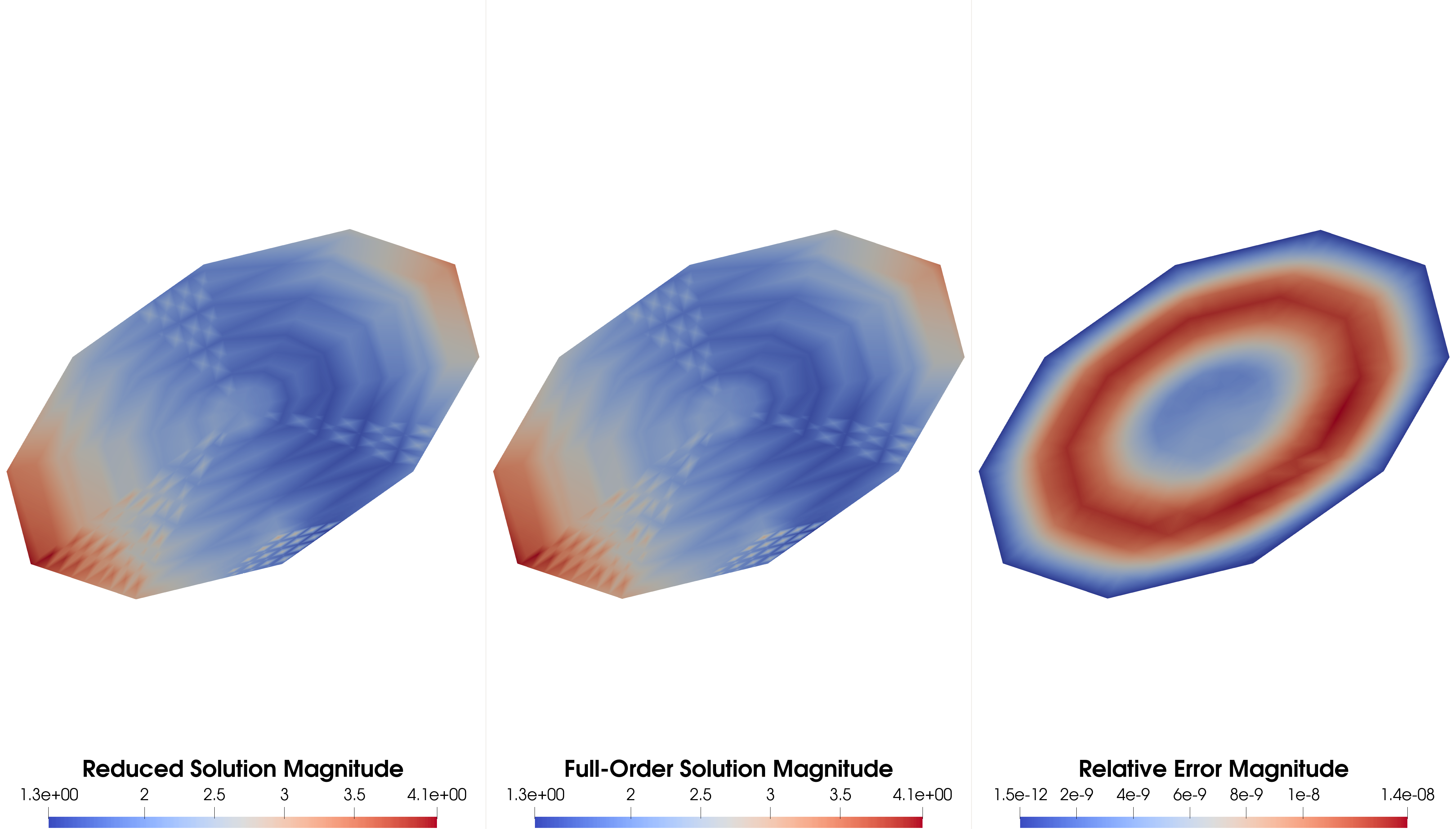}
    }
    
    \caption{Reduced approximation, full order solution, and relative error (left, middle, and right respectively) for Test Problem 2 on the boundary and on a slice with different views at time instances $T=0.5$ and $T=1$.}
    \label{tp_2:fig4}
\end{figure}

        Finally, Figure \ref{tp_2:fig4} shows results related to the comparison between the reduced approximation and the full-order solution in magnitude computed on the exact parameters. The error shows that the reduced simulation achieves an impressive accuracy  with maximum point-wise error of order $1e-08$. The reduced approximation has $L^2$ relative errors w.r.t.\ the full-order solution of $4.39e-06$ and $2.41e-06$ on the first and second components, respectively, and equal to $2.16e-02$ and $9.90e-03$ w.r.t.\ the exact solution.

    \subsection{Direct Problem Submodule}\label{d:}

\begin{figure}[!ht]
        \centering
        \includegraphics[width=\linewidth]{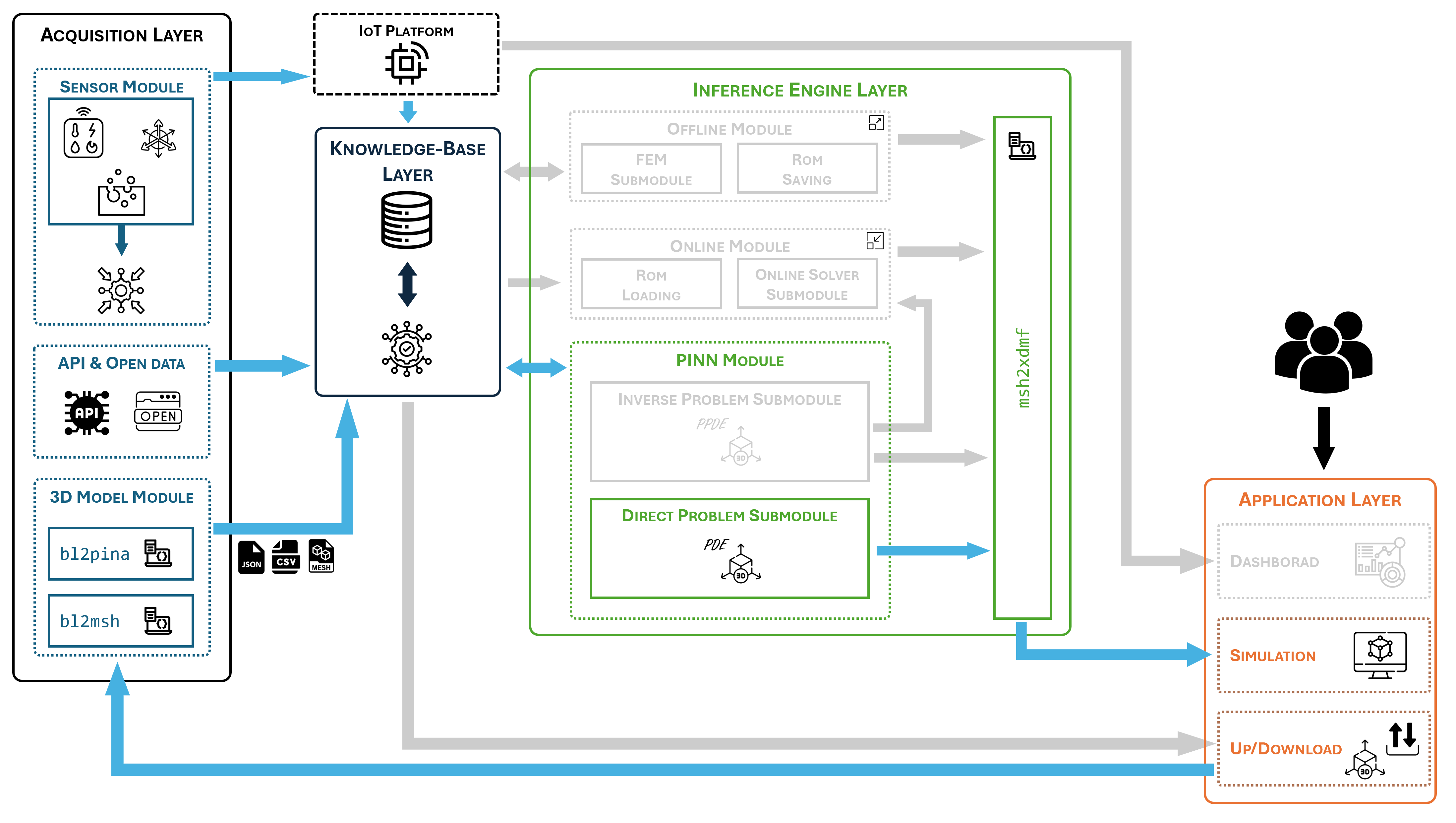}
        \caption{Architecture's active modules to obtain direct problem solutions via PINNs with known parameters.}
        \label{d:fig1}
    \end{figure}

    In the previous Section \ref{fr:}, we focused on the analysis of the framework's ability to handle parametrized PDEs by identifying the parameters via the Inverse Problem Submodule of the PINN Module, and then use the online phase of the ROM to obtain efficient simulations. However, not all problems in the cultural heritage field require the analysis of parametric PDEs, since sometimes the parameters are known, and only the evaluation of the PDE, or some output of interest, are required. As an alternative approach, within the proposed comprehensive framework, here we investigate the physical phenomena by applying PINNs for direct problems. We refer to Figure \ref{d:fig1} for the active modules in this setting.

    % In the case of experimentation for the resolution of direct problems in which parameter of PDEs are known, the architecture reacts differently to provide the services related to the simulation to the user.
    % .\fp{anche qui per figure e testo seguire i consigli in TP1}\fp{le sezioni mi stanno confondendo perché qui è Direct Problem Submodule ma usiamo PINN per inverse problem non è riportato nello schema di Figura 12}

    The workflow for acquiring 3D models is similar to the previous case: the Blender model is uploaded by the user via the upload/download module in the Application Layer, activating the 3D Model Module in the Acquisition Layer.
    The data related to the asset are stored in the Knowledge-Base Layer, also including the sensor information filtered through the IoT platform, or via API services and Open Data.
    
    The task of the Inference Engine is to tackle monitoring problems by combining physics-based and a data-driven approach. Thus, we employ PINNs and the Direct Problem Submodule as the Module of Interest to integrate the physical knowledge of the phenomena (known parameters and governing equations) with the data acquired either via the API or through the Sensor Module.

    % Consequently, on the one hand, we have a differential problem in which, unlike the case discussed in Section \ref{fr:}, the parameters are known.
    % Through the use of PINNs, it is possible to integrate the 
    % It follows that the Module of Interest in this experimental step is still the PINN Module, particularly the , where the processing takes place, allowing the integration of data with PDE-based problems in order to provide simulations to users.
    This experimental step uses benchmark problems in which the data simulate the boundary conditions of the analyzed PDE-based problems, while for the visualization part, the \textsf{msh2xdmf} Module is used again to post-process the simulation and store it in an $\textsf{xdmf}$ file.
    
    \subsubsection{Test Problem 3: Temperature monitoring via PINNs on a rock}\label{tp_4:}

\begin{figure}[!ht]
        \centering
        \includegraphics[width=0.9\linewidth]{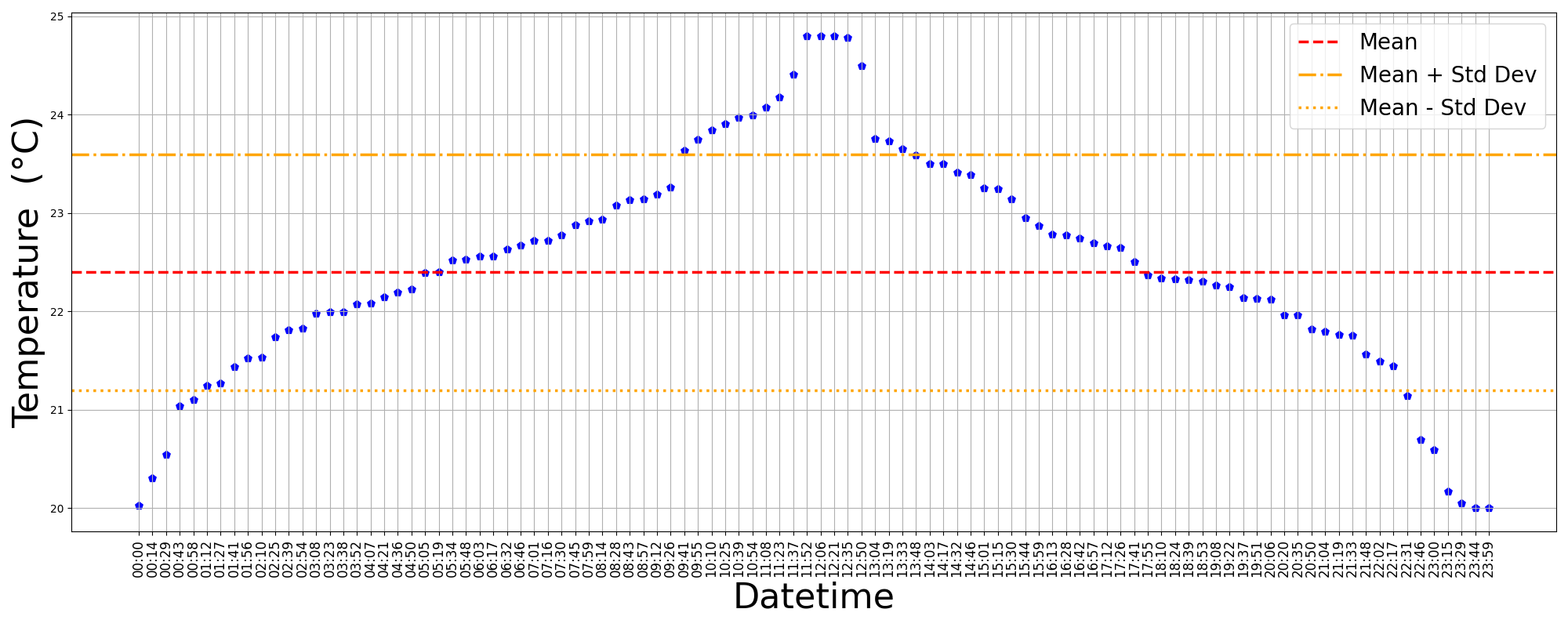}
        \caption{Simulated boundary temperature data for the rock domain during 24 hours for Test Problem 3.}
        \label{tp_4:fig1}
    \end{figure}

    We now consider a more realistic scenario related to the monitoring of temperature of an outdoor cultural asset. This test problem combines the governing physics of heat equation with data-driven information from the simulated setting showed in Figure \ref{tp_4:fig1}, which represents the boundary conditions on the rock domain $\Omega$ shown in Figure \ref{pre:fig_1}.
    Specifically, we consider the following heat equation
    \begin{equation}\label{tp_4:eq1}
        u_t \left( x, y, z, t \right) - \Delta u \left( x, y, z, t \right) = 0, \qquad \text{in}\ \Omega \times \left[ 0, 1 \right] ,
    \end{equation}

    where $u$ represents the scalar temperature distribution over the time interval $\left[ 0, 1 \right]$.

    % where the domain $\Omega$ consists of the rock of Figure \ref{pre:fig_1} introduced in Sub Section \ref{pre:}, \textcolor{red}{$u: \Omega \cup \partial \Omega \mapsto \mathbb{R}$ represents the solution,} and $k=1$. In addition, the first simulated data represents the initial condition of the problem seen as constant temperature.
    % \fp{qui sotto non ripeterei da capo tutta la parte discretizzazione tempo dip, diciamo come è stato fatto sopra.}

    % \fp{questa va rifatta più grande e con font sennò non si legge (acnhe leggenda)}

    Elaborating the benchmark problem requires the semi-discretization in time, as seen previously for the Test Problem 2, and the application of the implicit Euler method, from which we obtain a weak formulation similar to the one reported in Equation \eqref{tp_2:eq5}.

    Concerning the direct resolution via PINNs, we construct the network as described in Table \ref{tp_4:tab_1}. In particular, the main architecture is consistent with the ones before, meaning hidden layers with 200 neurons where the output represents the temperature $u \left( x, y, z, t \right)$ in the input point $\left( x, y, z \right)$ at time $t$.
    %, the input layer requires 4 inputs (three spatial variables $a, y, z$ and one temporal variable $t$),
    To deal with the more complex setting, we define the training phase exploiting $9900$ simulated data on the boundary and $400$ collocation points.
    
    \begin{table}[!ht]
    \centering
    \caption{PINN hyperparameters for Test Problem 3.}
    \begin{tabular}{l|c}
        {\bf Detail}       & {\bf Value} \\
        \hline
        Collocation points & 400 \\
        Data points        & 9900 \\
        Epochs             & 30000 \\
        Batch size         & - \\
        Learning rate      & $1e-03$ \\
        Decay rate         & $1e-08$ \\
        Optimizer          & Adam \\
        Network structure  & $\left[3, 200, 200, 1 \right]$
    \end{tabular}
    \label{tp_4:tab_1}
\end{table}
    
    The boundary data consists of $100$ fixed spatial points that simulate the sensors installed on the cultural asset analyzed. 
    For each spatial point, we have the temperature at each time of sampling $u \left( x_j, y_j, z_j, t_i \right) = T_i$, $j=1, \dots, 100$, $i=1, \dots, N$, where $\left\{ \left( x_j, y_j, z_j \right) \ : \ j=1, \dots, 100 \right\} \subseteq \partial \Omega$ represent the sensors' location on the boundary $\partial \Omega$, and $N=99$ being the number of time samples (excluding the first one $T_0$ employed as initial condition).
    
    In this case, we employ the classical PINN structure with loss $\mathcal{L}$ balancing the contribution of the physics-based term and the data-driven one \eqref{pinn:eq2} as
    % , as defined in \cite{Raissi2019686}, without the employment of the Residual-Basis Attention approach \cite{Anagnostopoulos2024}.
    % Therefore, the definition of the loss $\mathcal{L}$ composed by the one related to the residual  and the one related to the data 
    \begin{equation}\label{tp_4:eq8}
        \mathcal{L} = w_\mathrm{residual} \mathcal{L}_\mathrm{residual} + w_\mathrm{data} \mathcal{L}_\mathrm{data},
    \end{equation}
    where the hyperparameters $w_\mathrm{residual}$ and $w_\mathrm{data}$ weight the information for the optimization task, chosen as $w_\mathrm{residual} = 0.1$ and $w_\mathrm{data} = 0.9$.
    
    We remark that in order to be more physically consistent with the dynamical setting, we chose to apply here hard constraints \cite{Lagaris1998987} to automatically satisfy the initial condition $u \left(x, y, z, 0 \right) = T_0,$ for $\left( x, y, z \right) \in \Omega$ of the problem \eqref{tp_4:eq1}, with $T_0 \approx 20.03$. The integration of hard constraints avoids the training of the initial condition by modifying the network output to match $T_0$ at time $t=0$. Specifically, the new output $\hat{u} \left( x, y, z, t \right)$ of the network is adapted as follows
    \begin{equation}\label{tp_4:eq10}
        \hat{u} \left( x, y, z, t \right) = T_0 + t \ u \left(x, y, z, t \right) .
    \end{equation}

\begin{figure}[!htp]
    \centering
    % Prima immagine
    \subfigure[Boundary - $T=0.5$ - Front view]{
        \includegraphics[width=.42\textwidth]{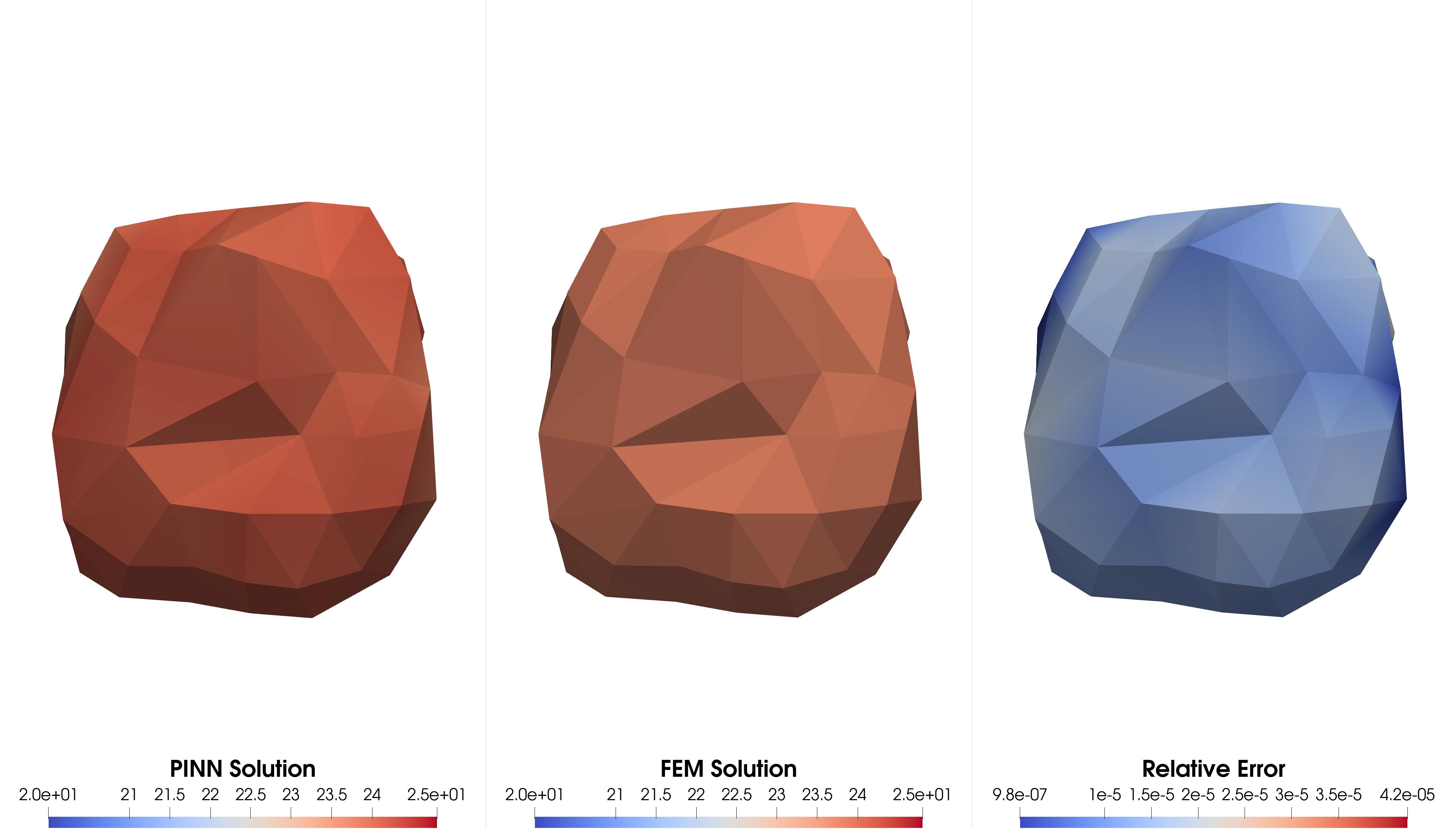}
    }
    \subfigure[Boundary - $T=0.5$ - Isometric View]{
        \includegraphics[width=.42\textwidth]{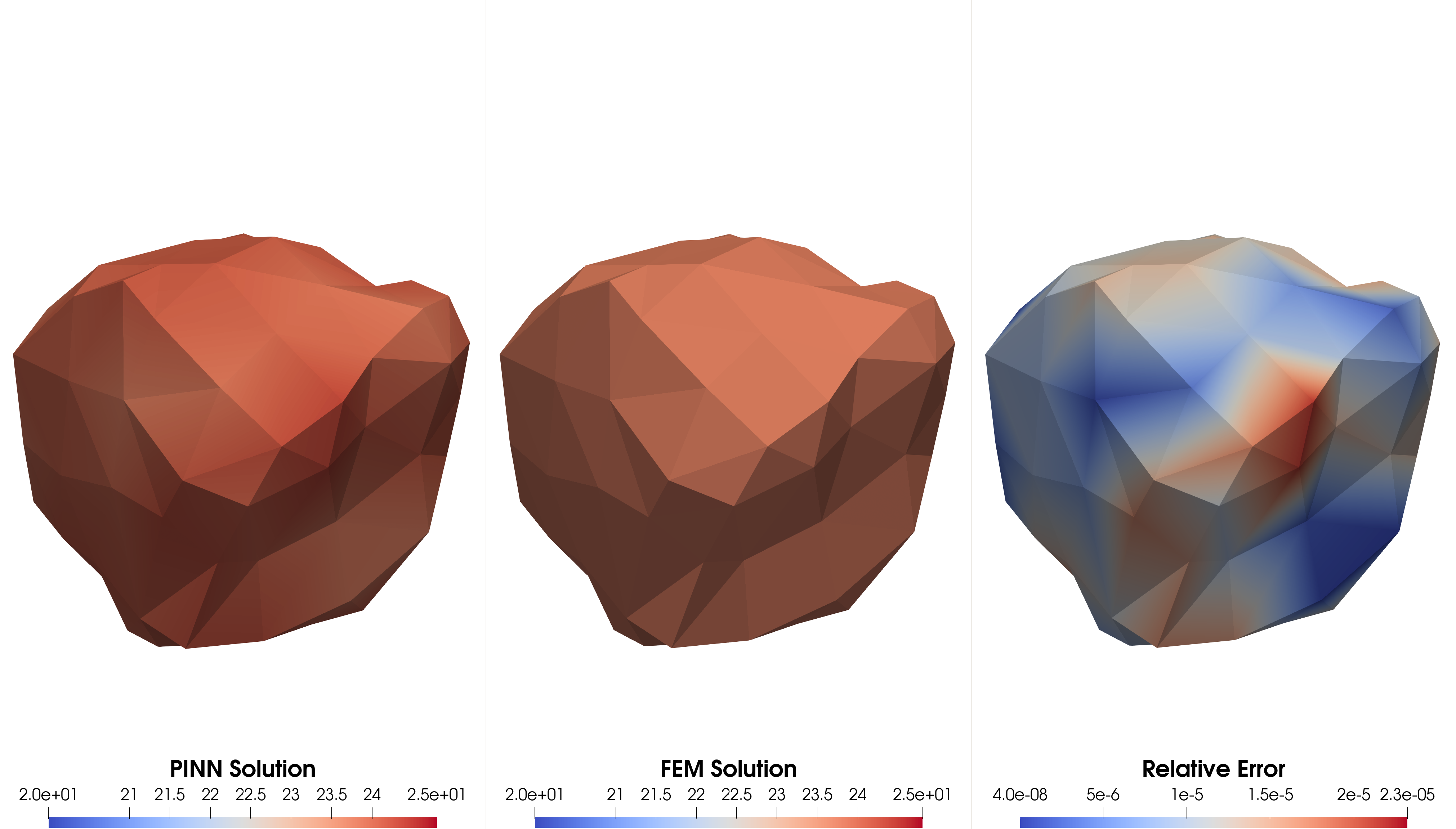}
    }
    
    % Seconda immagine
    \subfigure[Slice - $T=0.5$ - Front view]{
        \includegraphics[width=.42\textwidth]{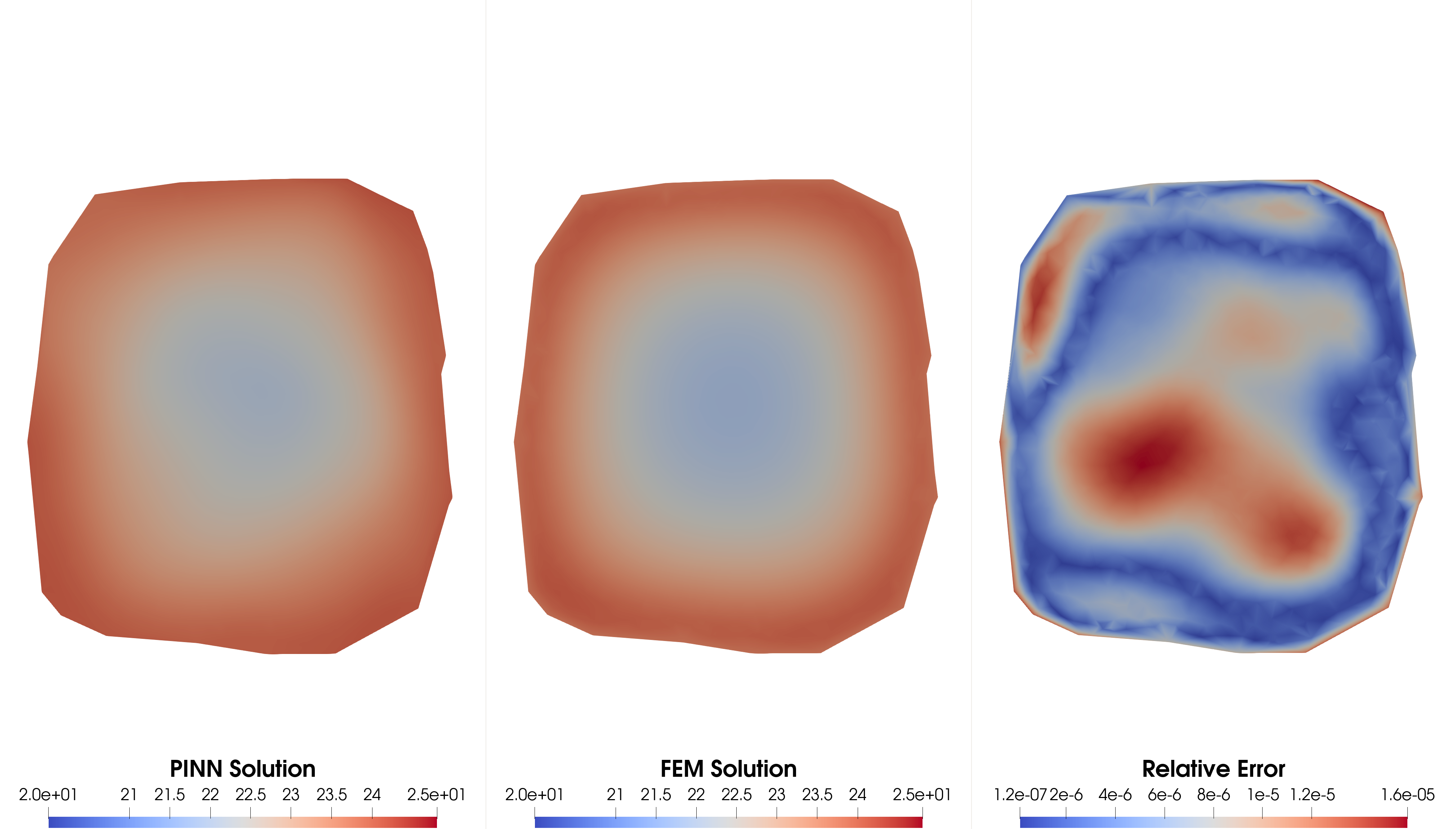}
    }
    % Seconda immagine
    \subfigure[Slice - $T=0.5$ - Isometric View]{
        \includegraphics[width=.42\textwidth]{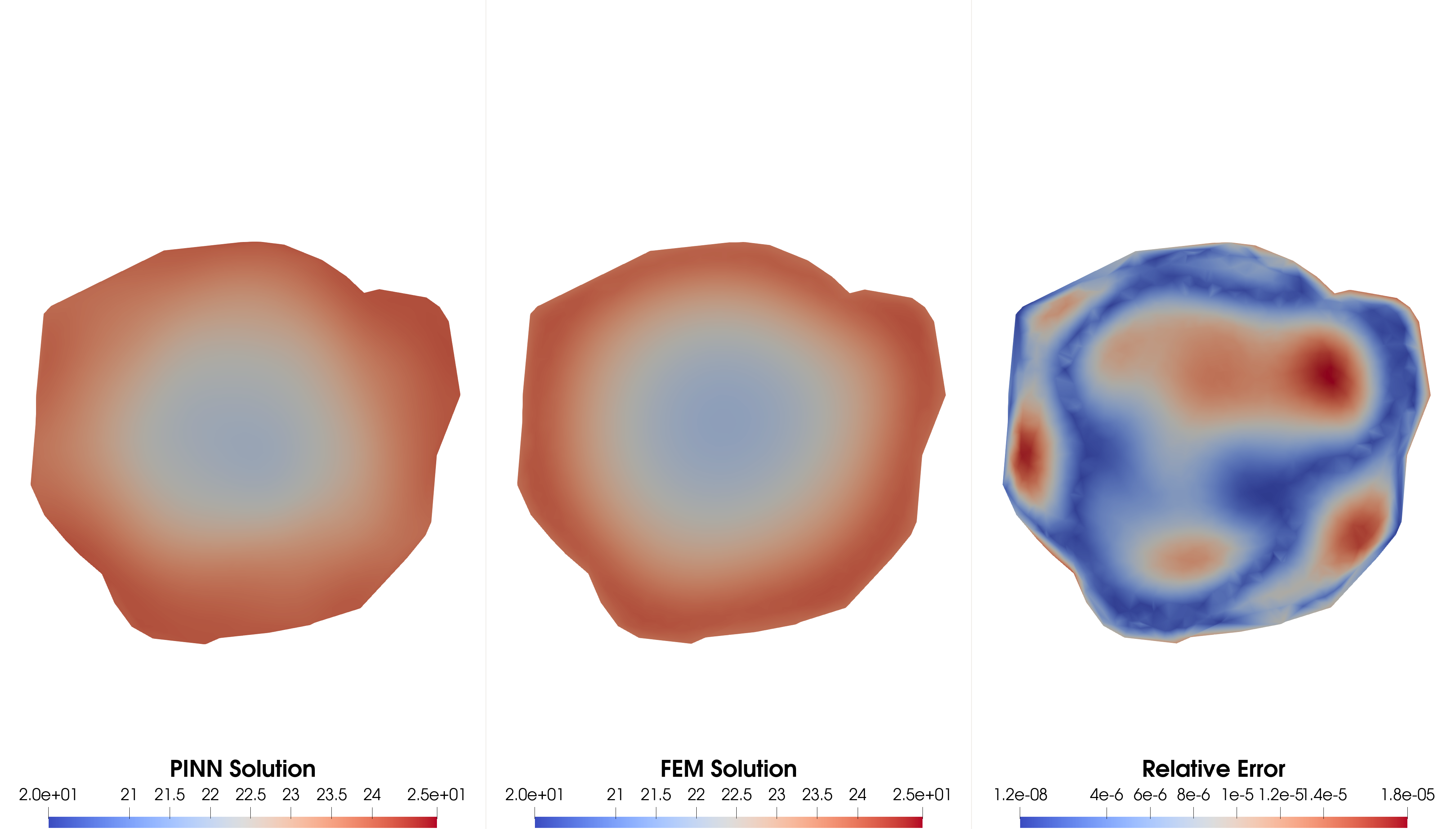}
    }

    % Terza immagine
    \subfigure[Boundary - $T=1.0$ - Front view]{
        \includegraphics[width=.42\textwidth]{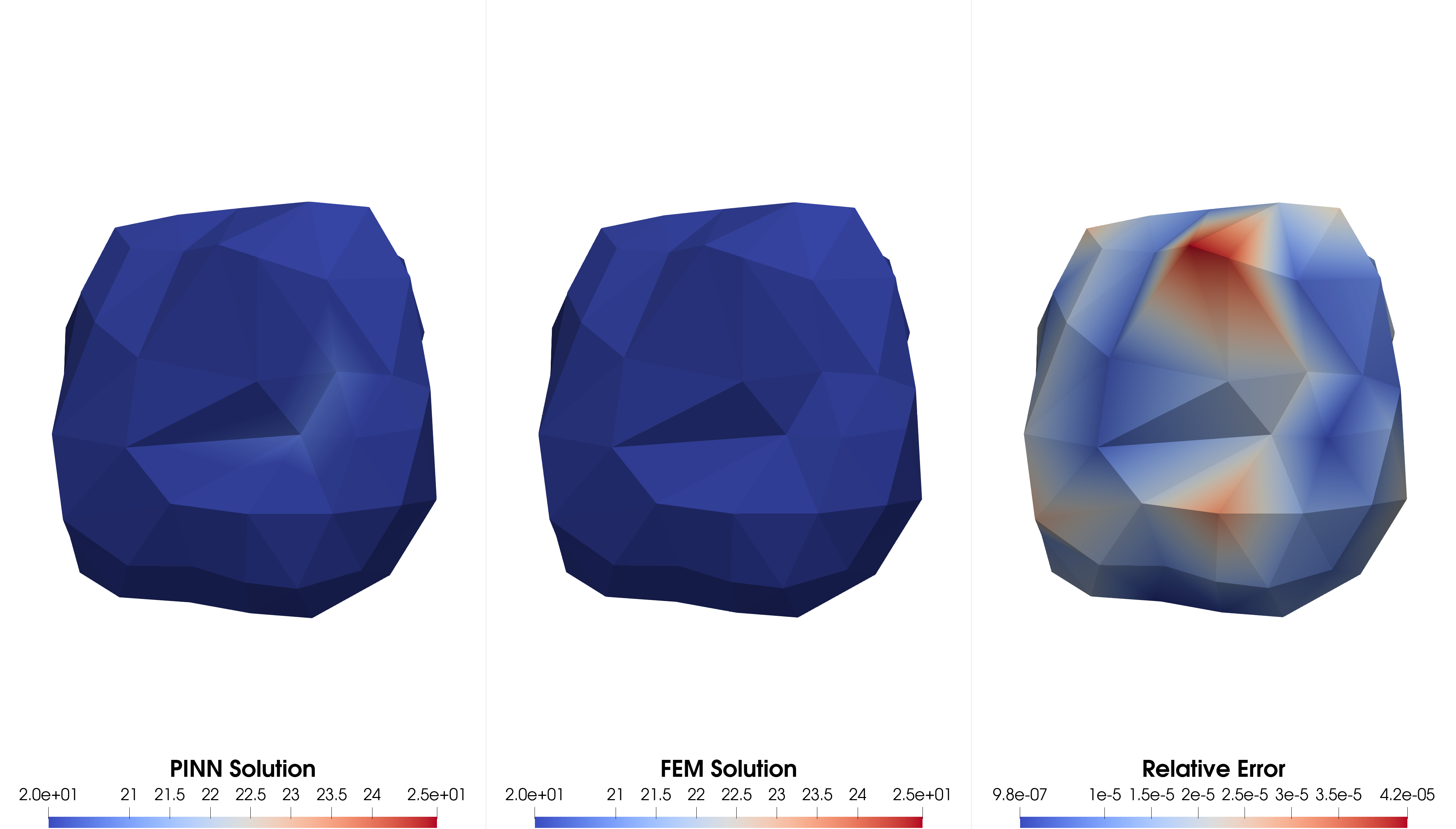}
    }
    % Terza immagine
    \subfigure[Boundary - $T=1.0$ - Isometric View]{
        \includegraphics[width=.42\textwidth]{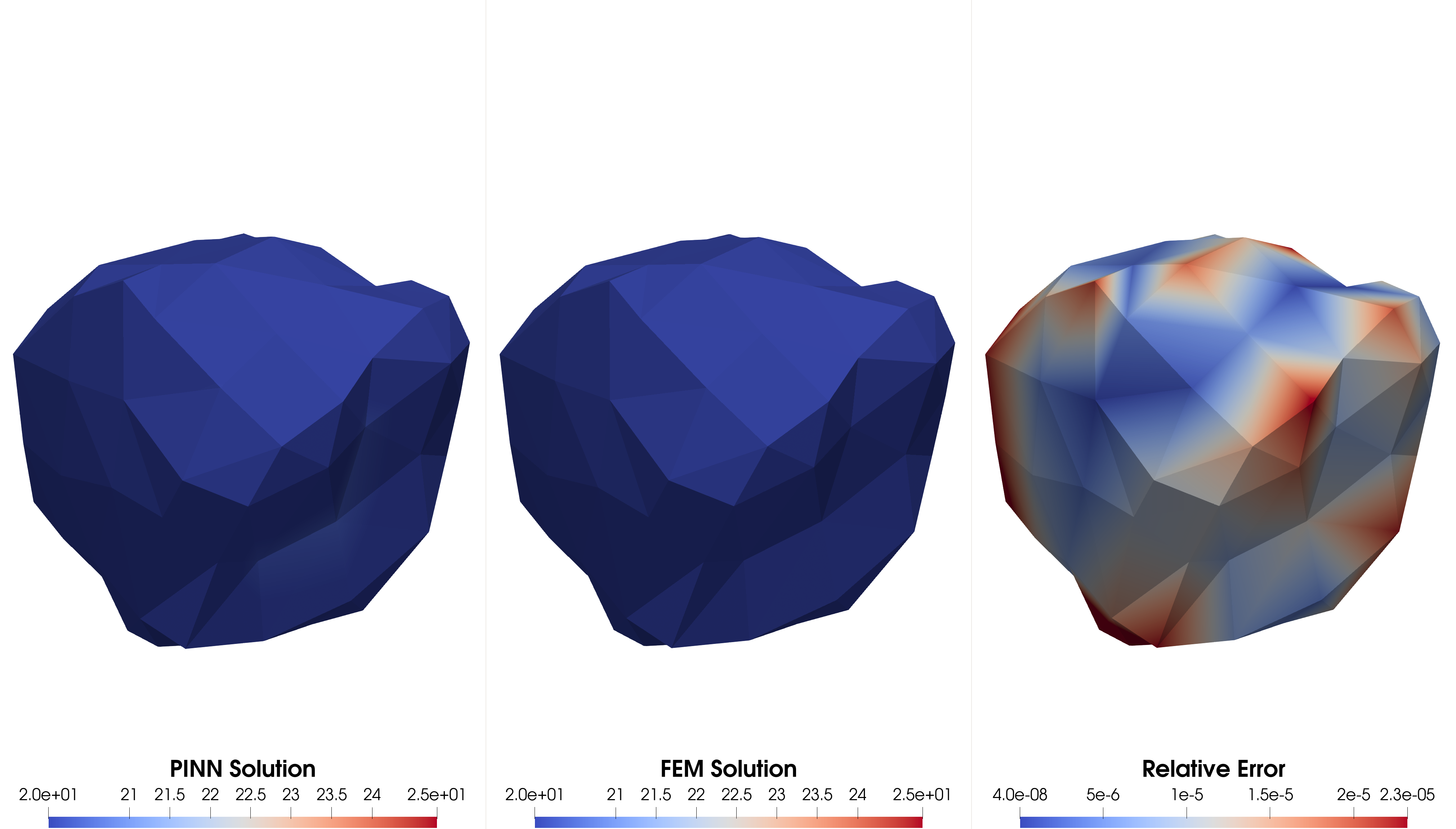}
    }

    % Quarta immagine
    \subfigure[Slice - $T=1.0$ - Front view]{
        \includegraphics[width=.42\textwidth]{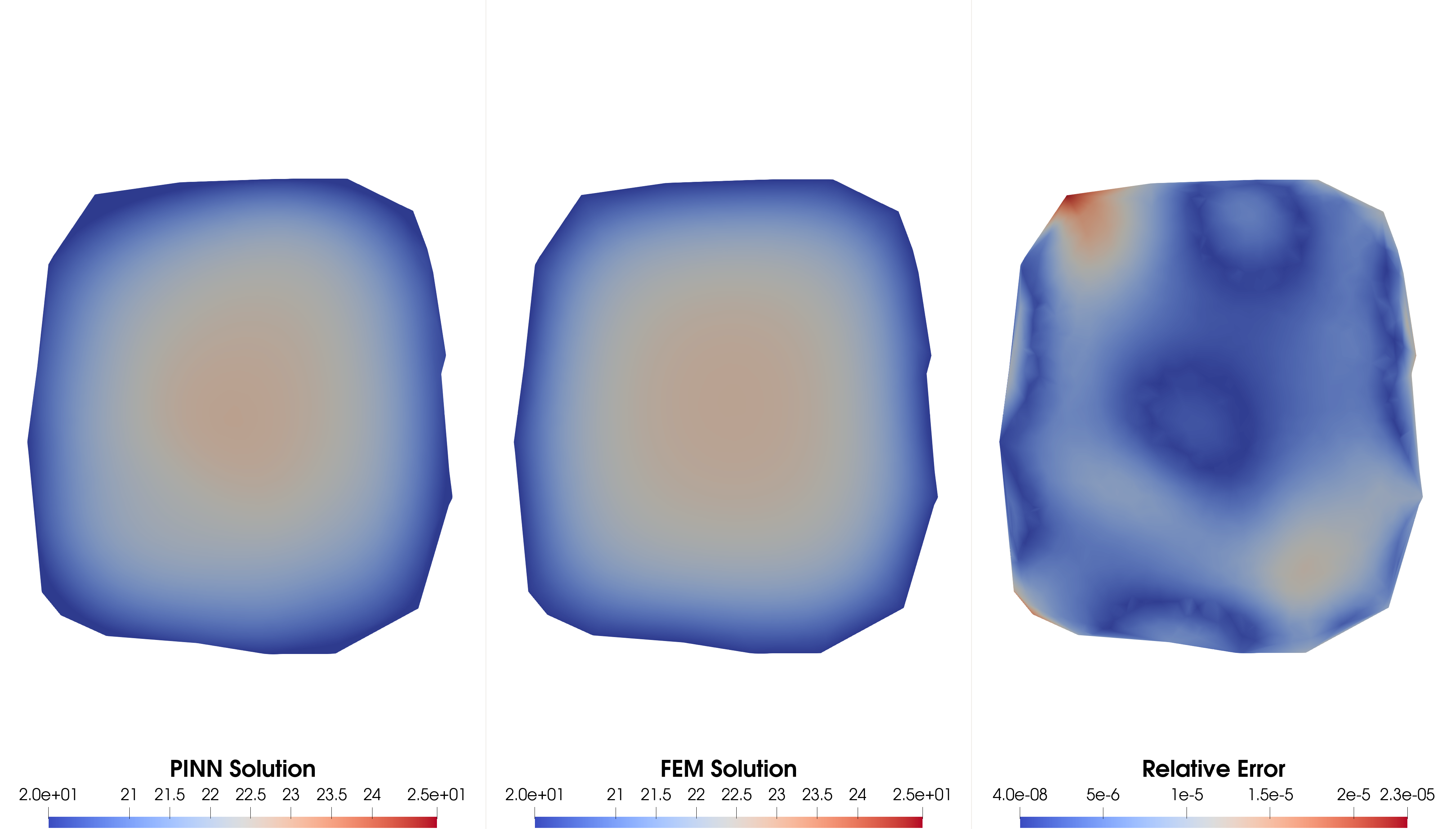}
    }
    % Quarta immagine
    \subfigure[Slice - $T=1.0$ - Isometric View]{
        \includegraphics[width=.42\textwidth]{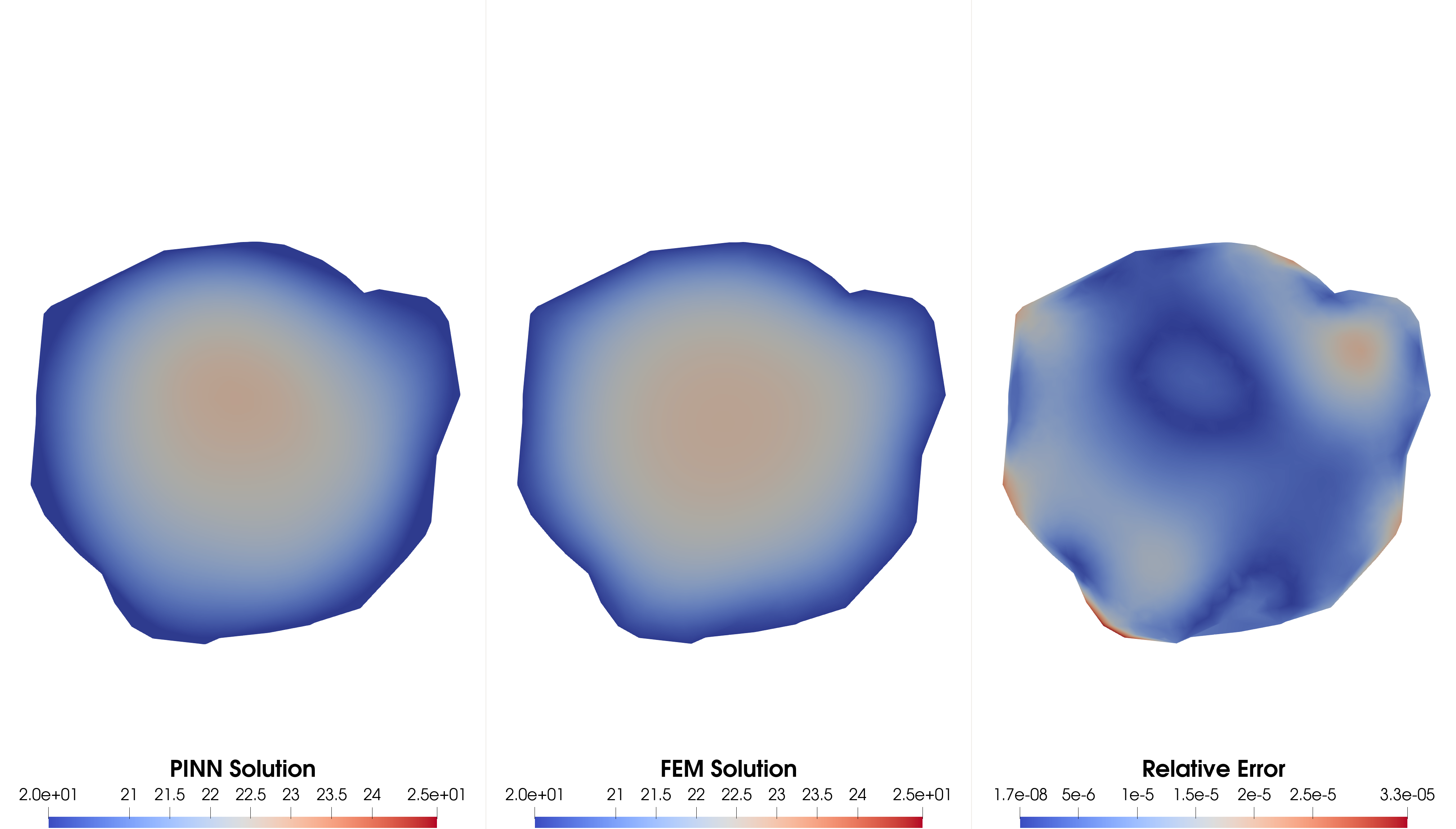}
    }
    \caption{Reduced approximation, full order solution, and relative error (left, middle, and right respectively) for Test Problem 3 on the boundary and on a slice with different views at time instances $T=0.5$ and $T=1$.}
    \label{tp_4:fig4}
\end{figure}

    In Figure \ref{tp_4:fig4} we show the solution approximated via PINN, the benchmark solution elaborated through FEM, and the relative error at time instances $T=0.5, 1$ on the boundary and inside the domain.

    The results show that the solution approximated with the PINN is reliable compared with the benchmark one, validating the correctness of the proposed integrated framework for temperature monitoring when external simulated scenario have to be imposed on the system, resulting in an $L^2$ relative error w.r.t.\ the benchmark solution equal to $8.70e-03$. In addition, the advantage of integrating data with a physics-based approach represent a significant step for exploiting the Internet of Things paradigm and the physical knowledge in a unified workflow.

    \subsubsection{Test Problem 4: Diffusion reaction system on a column}\label{tp_5:}

    Here, we aim at investigating the capabilities of the framework when approximating the solution of the following system of diffusion-reaction PDEs on the column domain $\Omega \subseteq \mathbb{R}^3$ shown in Figure \ref{pre:fig_2b}
    \begin{equation}\label{tp_5:eq1}
        \Delta \mathbf{u} \left( x, y, z \right) = F \left( x, y, z, \mathbf{u} \left( x, y, z \right) \right), \qquad \text{in}\ \Omega
    \end{equation}
    where the solution $\mathbf{u} = [u^{\left( 1 \right)}, u^{\left( 2 \right)}]^T \in \mathbb{R}^2$ could represent the concentration of some species, such as corrosion or contaminants, of fundamental importance for the preservation of the cultural asset, and the forcing term and boundary conditions (coinciding with the analytical solution) are respectively given by 
    \begin{equation}\label{tp_5:eq2}
        F \left( x, y, z, \mathbf{u} \right) = \left(
            \begin{array}{c}
                2 u^{\left( 1 \right)} \\
                2
            \end{array}
        \right), \ \text{in}\ \Omega \qquad  \text{and} \qquad
        u \left( x, y, z \right) = \left(
            \begin{array}{c}
                e^{x+y}\\
                x^2 - z
            \end{array}
        \right), \ \text{on}\ \partial\Omega .
    \end{equation}

\begin{figure}[!ht]
        \centering
        \includegraphics[width=.9\linewidth]{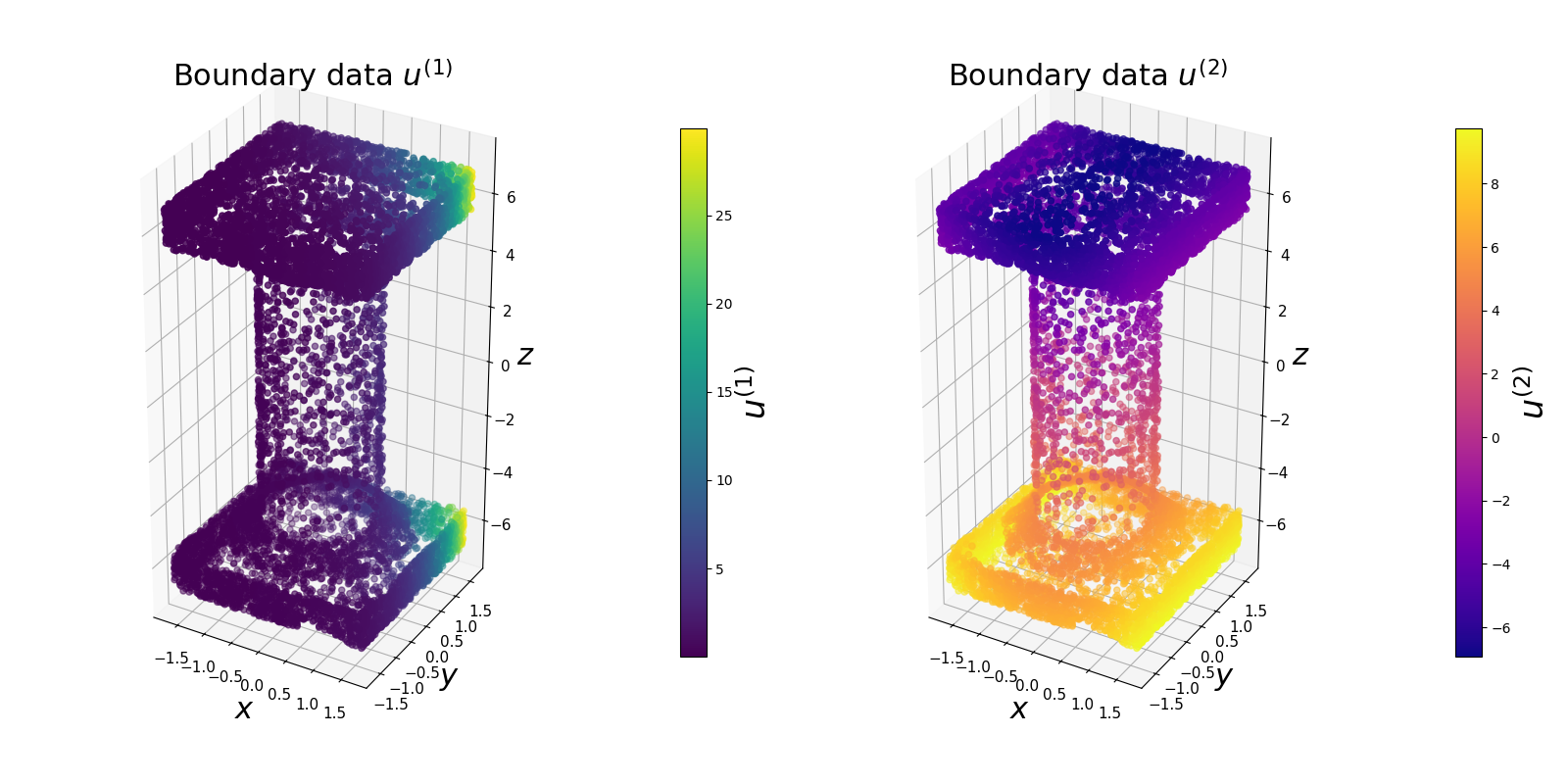}
        \caption{Values of boundary data with respect the two components $u^{\left( 1 \right)}$ and $u^{\left( 2 \right)}$.}
        \label{tp_5:fig1}
    \end{figure}

    With this benchmark we focus on the comparison between two different analysis towards the real-world application setting: (i) a physics-only approach defining the residual and boundary losses based on \eqref{tp_5:eq1} and \eqref{tp_5:eq2}, and (ii) data-integration using boundary condition as simulated acquired data, as shown in Figure \ref{tp_5:fig1}.

    Table \ref{tp_5:tab_1} resumes the hyperparameters of the network employed that exploits the Residual-Basis Attention approach \cite{Anagnostopoulos2024} integrated with the Stochastic Weight Averaging strategy to improve the optimizer employed \cite{Izmailov2018876}.

    \begin{table}[!ht]
    \centering
    \caption{PINN hyperparameters for Test Problem 4 with and without data integration.}
    \begin{tabular}{ccc}
        {\bf Physics} & & {\bf Data integration} \\
        \begin{tabular}{l|c}
    {\bf Detail}       & {\bf Value} \\
    \hline
    Collocation Points & 1000 \\
    Boundary Points    & 500 \\
    Data Points                  & - \\
    Epochs             & 5000 \\
    Batch size         & - \\
    Learning rate      & $1e-03$ \\
    Decay rate         & $1e-08$ \\
    Optimizer          & Adam \\
    Network structure  & $\left[3, 200, 200, 1 \right]$
\end{tabular} & & \begin{tabular}{l|c}
    {\bf Detail}       & {\bf Value} \\
    \hline
    Collocation Points & 1000 \\
    Boundary Points    & 500 \\
    Data Points        & 500 \\
    Epochs             & 5000 \\
    Batch size         & - \\
    Learning rate      & $5e-04$ \\
    Decay rate         & $1e-08$ \\
    Optimizer          & Adam \\
    Network structure  & $\left[3, 200, 200, 1 \right]$
\end{tabular}
    \end{tabular}
    \label{tp_5:tab_1}
\end{table}

\begin{figure}[!htp]
    \centering
    % Prima immagine
    \subfigure[Boundary - Physics - Front view]{
        \includegraphics[width=.42\textwidth]{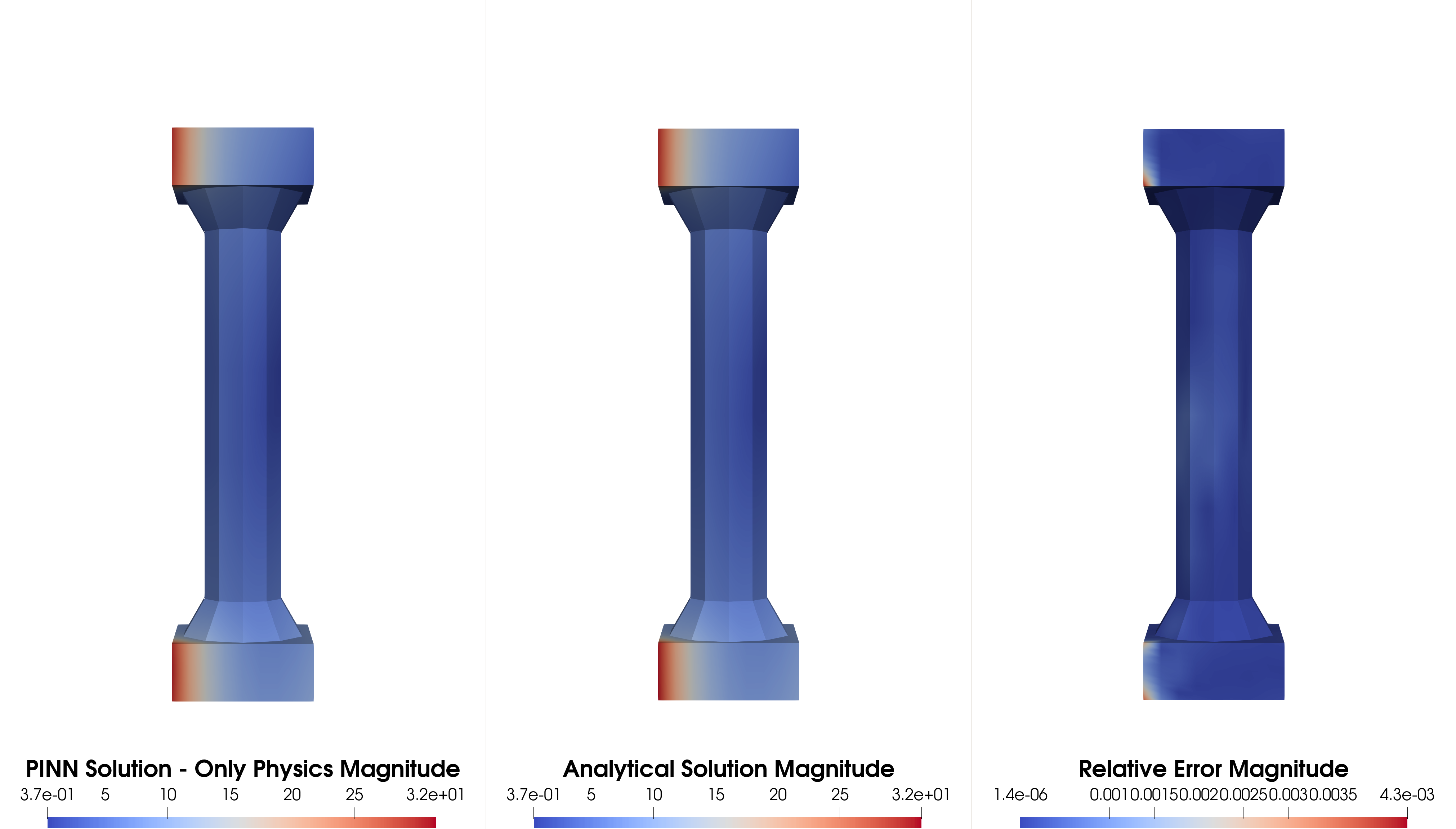}
    }
    \subfigure[Boundary - Data - Front View]{
        \includegraphics[width=.42\textwidth]{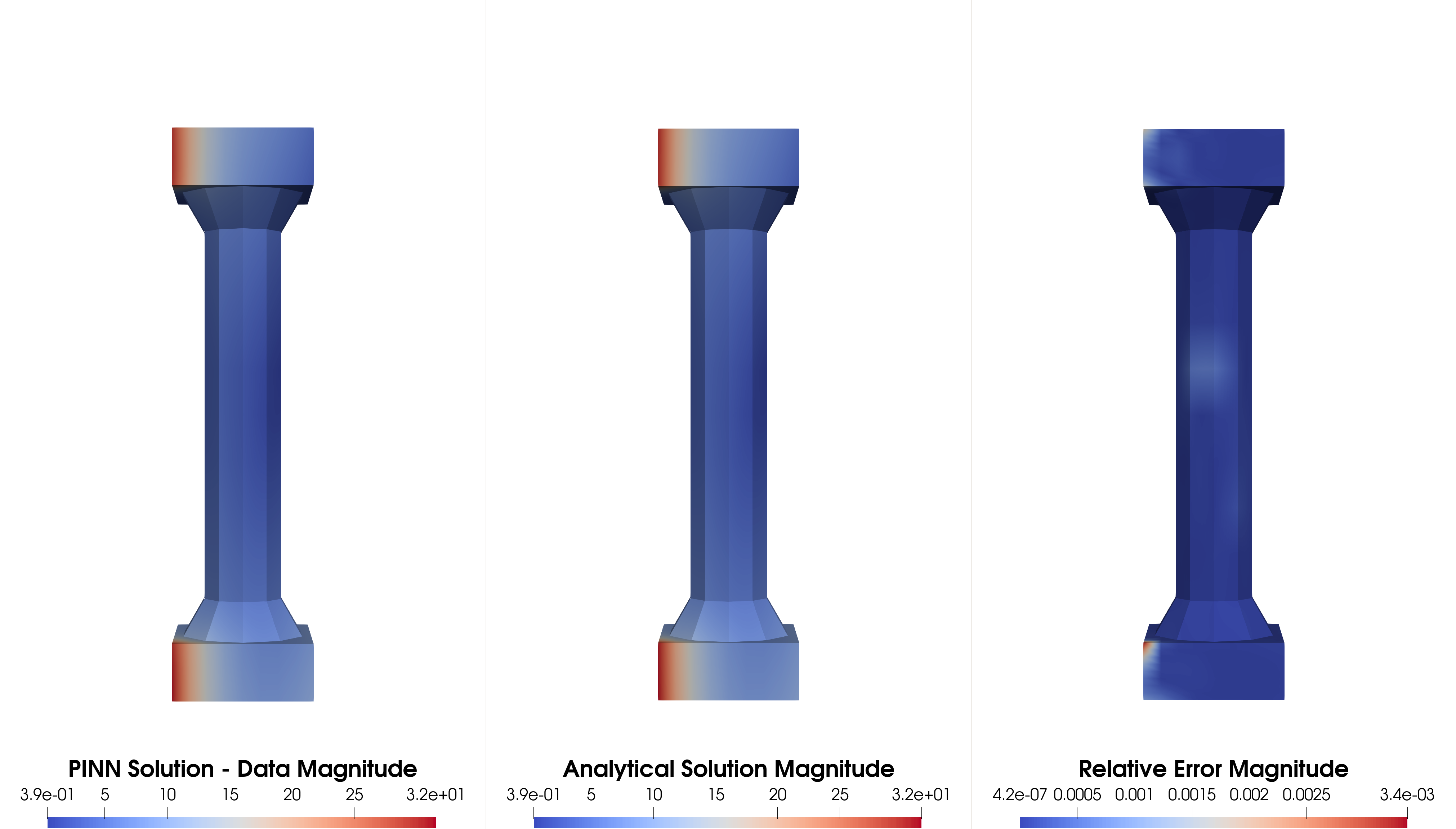}
    }
    
    % Seconda immagine
    \subfigure[Boundary - Physics - Isometric view]{
        \includegraphics[width=.42\textwidth]{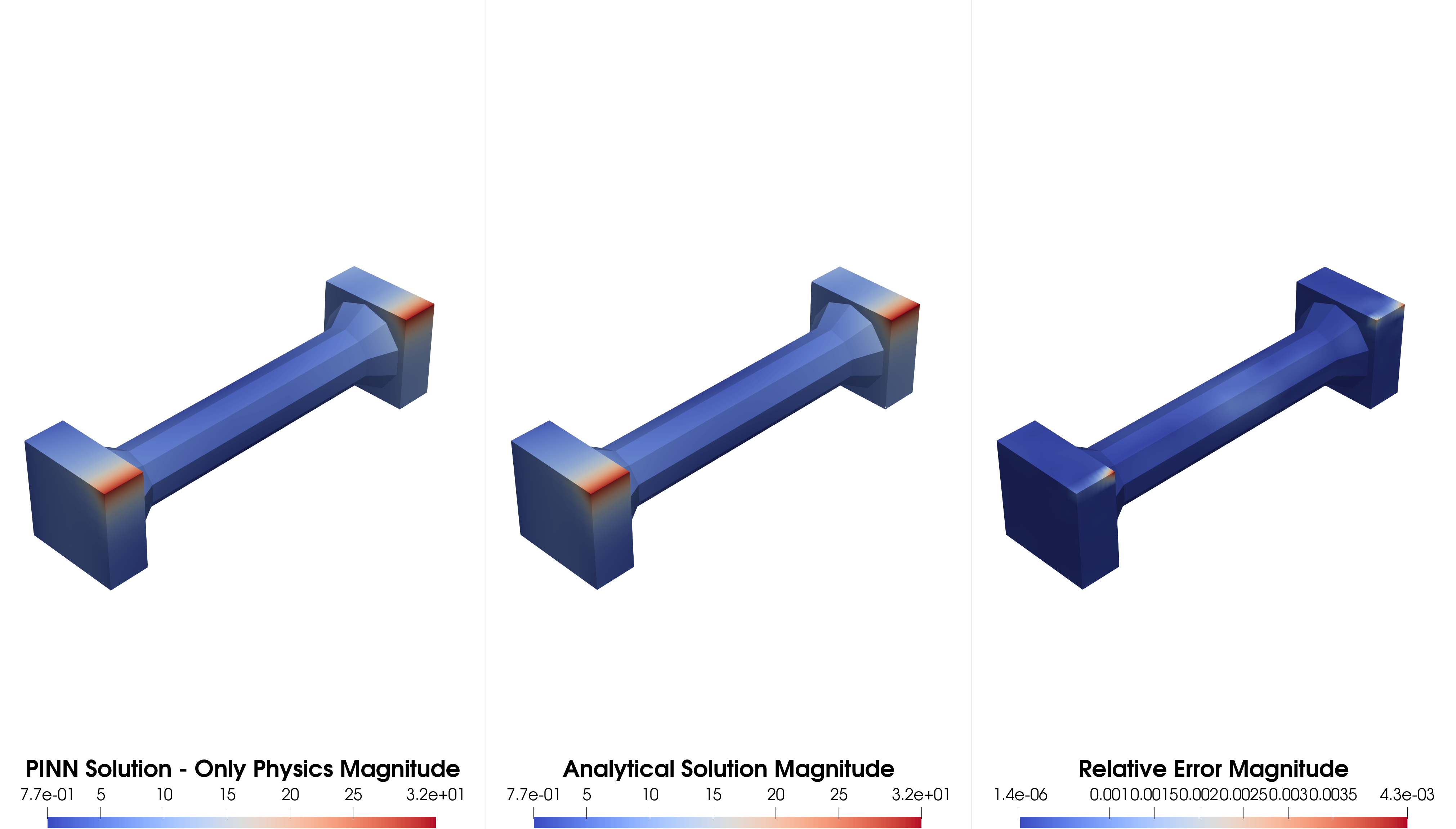}
    }
    % Seconda immagine
    \subfigure[Boundary - Data - Isometric View]{
        \includegraphics[width=.42\textwidth]{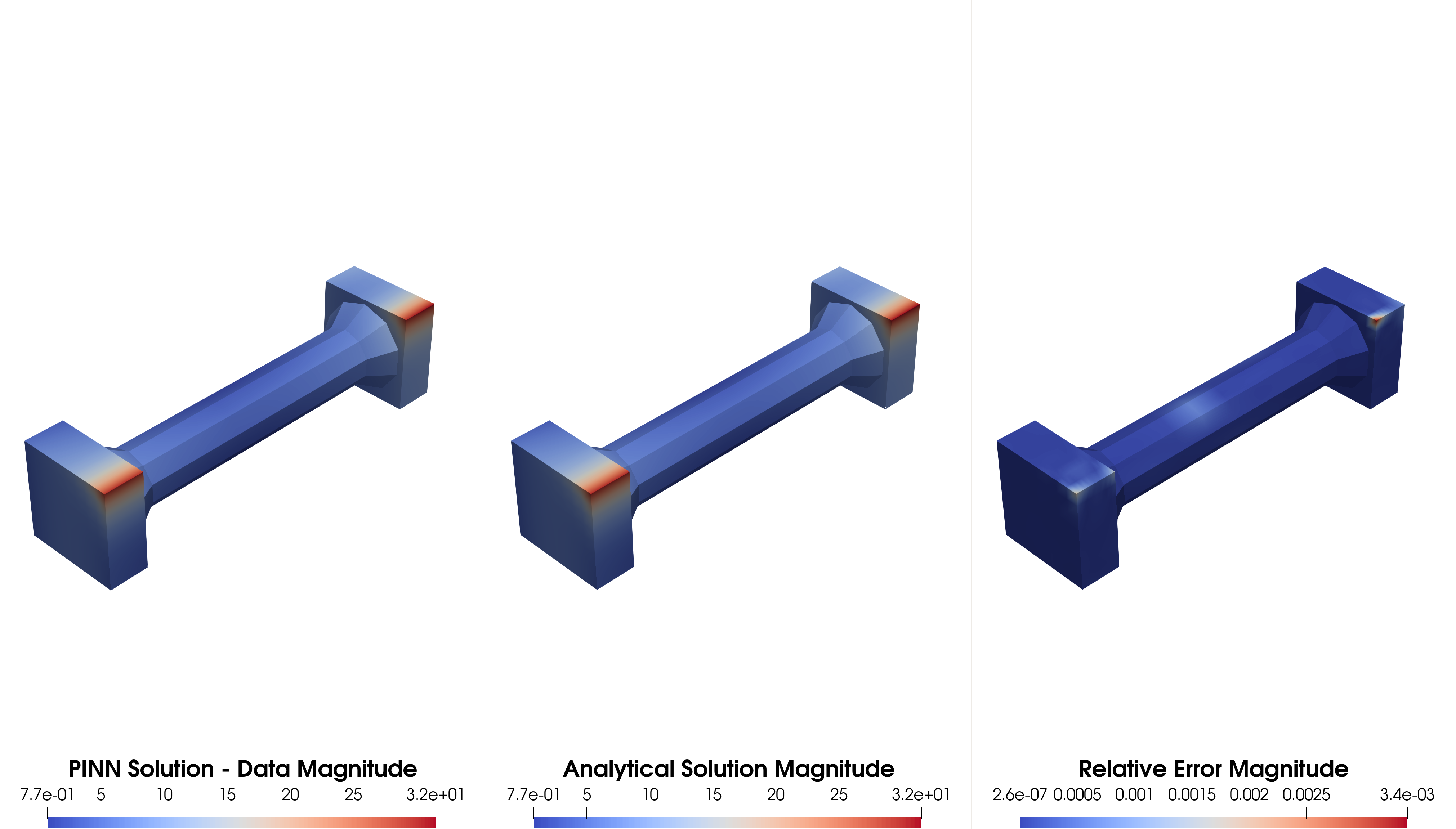}
    }

    % Terza immagine
    \subfigure[Slice - Physics - Front view]{
        \includegraphics[width=.42\textwidth]{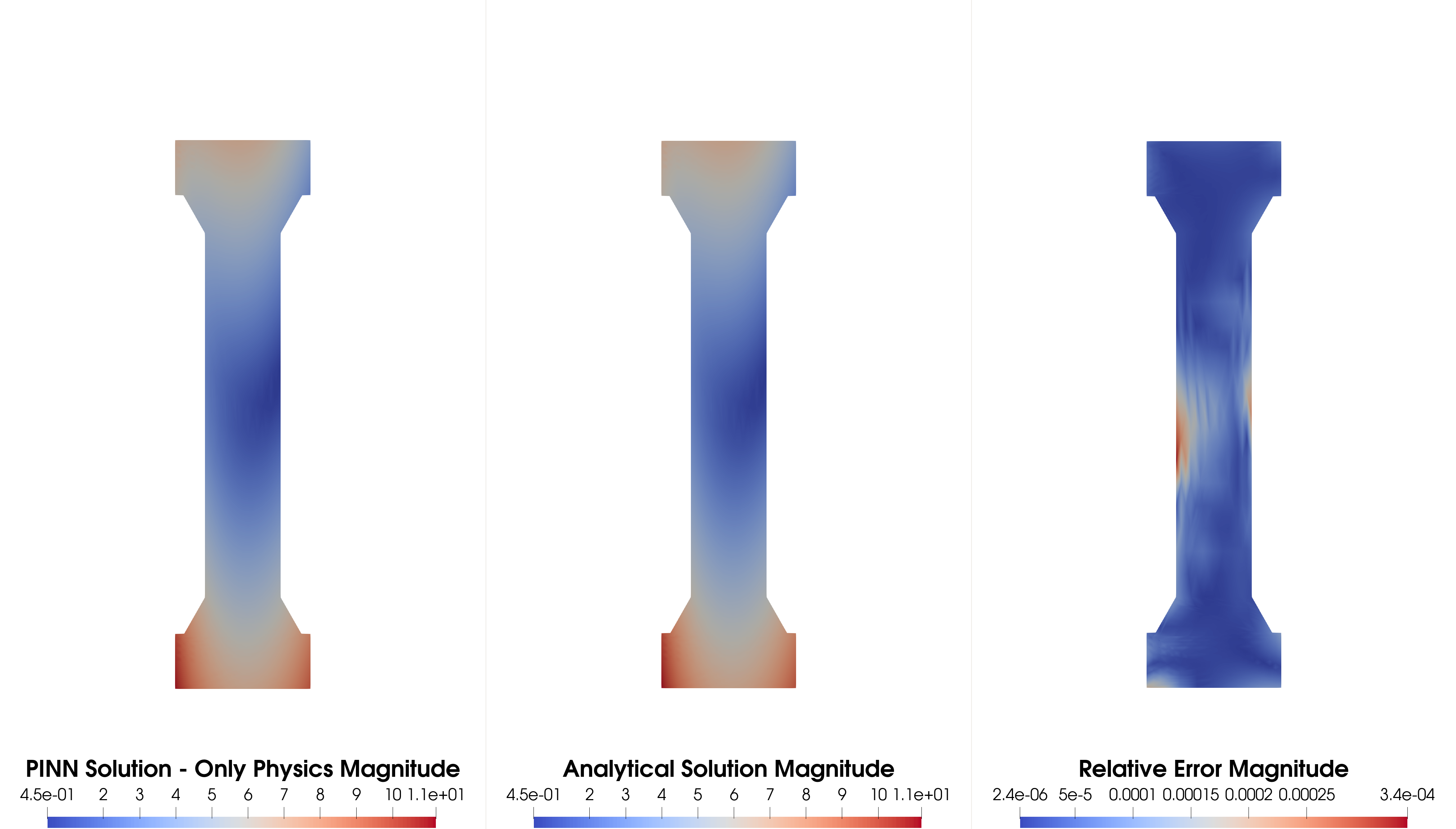}
    }
    % Terza immagine
    \subfigure[Slice - Data - Front View]{
        \includegraphics[width=.42\textwidth]{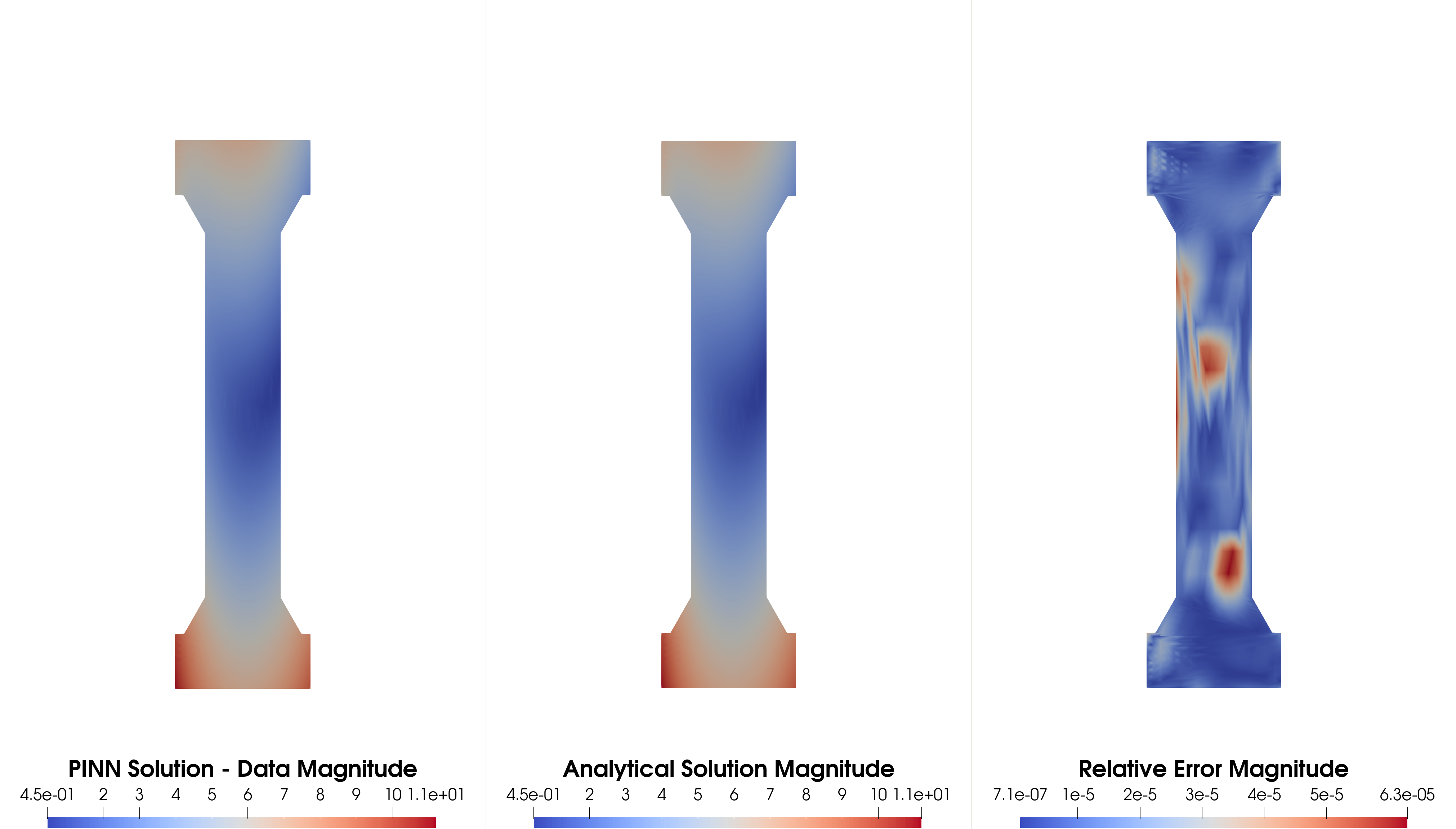}
    }

    % Quarta immagine
    \subfigure[Slice - Physics - Isometric view]{
        \includegraphics[width=.42\textwidth]{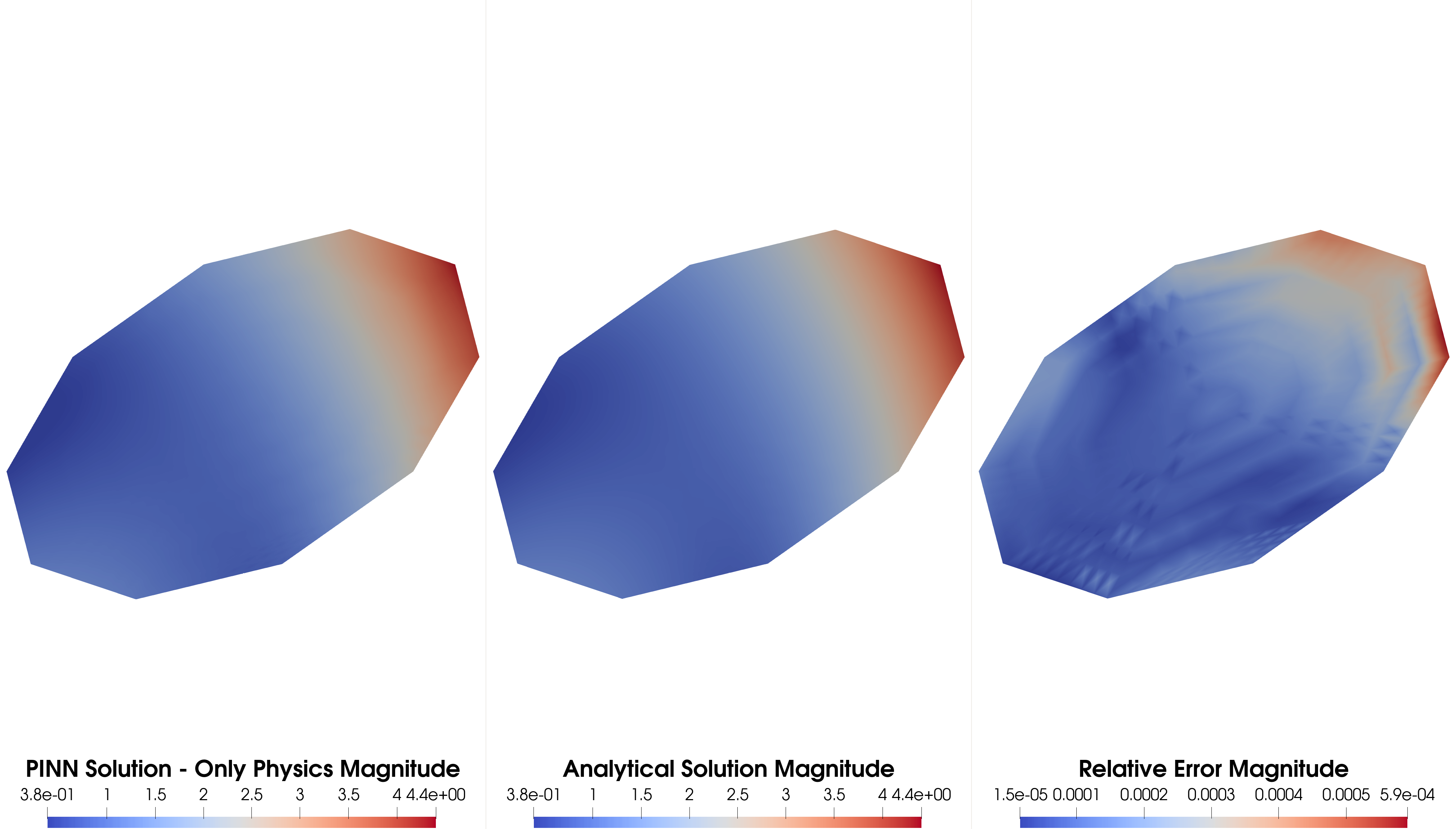}
    }
    % Quarta immagine
    \subfigure[Slice - Data - Isometric View]{
        \includegraphics[width=.42\textwidth]{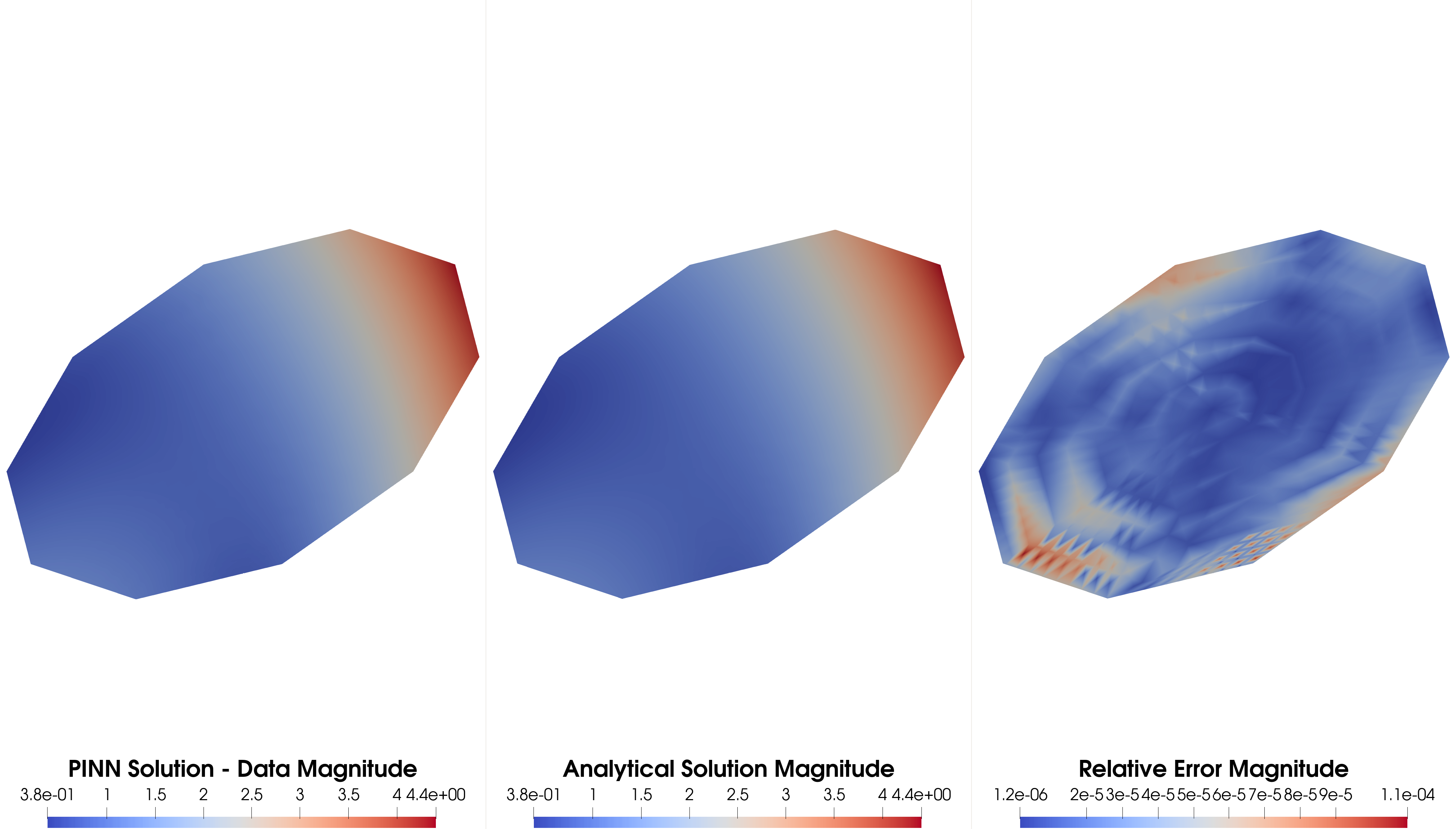}
    }
    
    \caption{ Comparison between the PINN and the full order FEM solutions, left and middle respectively, and the corresponding relative errors (right), for Test Problem 4 on the boundary and on a slice with different views.}
    \label{tp_5:fig4}
\end{figure}

    The results obtained are compared by means of the relative error metric with the analytical solution.
    In particular, Figure \ref{tp_5:fig4} shows the results obtained for the two analysis via the magnitude of the solution field on the boundary and inside the domain. For a more quantitative insight on the approximation accuracy obtained, Table \ref{tp_5:tab2} contains the errors on the specific components $u^{\left( 1 \right)}$ and $u^{\left( 2 \right)}$ of the solution and on its magnitude.

% \fp{Stessa cosa di sopra per questi risultati (qui bene il background trasparente), staccherei di più le due colonne a sinistra in alto ed eviterei di mettere tutti i valori che non si leggono nella colorbar, c e d vanno bene, ma a e b troppo. Possiamo rifarle tutte coerenti.}

    \begin{table}[!ht]
    \centering
    \caption{Relative errors for physics-based and data integration approaches on the components $u^{ \left( 1 \right) }$, $u^{ \left( 2 \right) }$, and the magnitude for the Test Problem 4.}
    \begin{tabular}{l|c|c|}
    \cline{2-3}
         & \multicolumn{2}{c|}{\bf Relative Error} \\ 
         \hline
        \multicolumn{1}{|l|}{{\bf Component}} & {\bf Physics} & {\bf Data integration} \\ \hline
        \multicolumn{1}{|l|}{$u^{\left( 1 \right)}$} & 1.18e-02 & 7.42e-03 \\
        \multicolumn{1}{|l|}{$u^{\left( 2 \right)}$} & 4.88e-03 & 1.01e-03 \\
        \multicolumn{1}{|l|}{{\sc Magnitude}}       & 7.60e-03 & 4.10e-03 \\ \hline
    \end{tabular}
    \label{tp_5:tab2}
\end{table}

    In general, we observed comparable results among the only-physics and data-integration approaches, underlying how the integrated framework provides robust and reliable results even for complex geometries coming from real-world scenarios. Moreover, the improvement of the results when employing data acquired from sensors validate the application of the PINN strategy, confirming the potential of this approach for real-time health structure monitoring. Specifically, the possibility of easily integrate data and physics in a SciML paradigm under the same framework enables a full set of novel technologies to provide reliable simulation aimed to the predictive maintenance of cultural heritage.

\section{Discussion and Conclusions}

    This work presented a comprehensive and modular framework for the conservation and predictive maintenance of cultural heritage assets, integrating Internet of Things (IoT) technologies, Artificial Intelligence, and physical knowledge of the phenomena of interest.
    By critically analyzing the state of the art, the study identified a lack of unified approaches capable of jointly exploiting data-driven techniques, physics-based modeling, and automated 3D model processing within a Digital Twin (DT) perspective.
    To address this gap, a four-layer architecture was introduced, enabling the acquisition of heterogeneous data and digital replicas, structured knowledge storage and pre-processing, advanced simulation and inference, and the visualization of results for expert users involved in cultural heritage conservation.

    The experimental results and architectural design demonstrate the feasibility and effectiveness of the proposed framework.
    A key contribution is the development of a modular, automated 3D Model Module capable of processing complex digital replicas acquired via laser scanning or photogrammetry.
    Using Blender APIs, geometric and semantic information is automatically extracted and converted into structured key-value representations, facilitating interoperability with downstream simulation tools.
    This process enables automatic domain sampling and mesh generation, effectively bridging the gap between raw geometric data and simulation-ready models, and has proven scalable across assets with varying topological complexity.

    The framework integrates Scientific Machine Learning techniques, combining Physics-Informed Neural Networks (PINNs) with Reduced Order Models (ROMs), thereby leveraging the interpretability of physics-based approaches and the adaptability of data-driven methods.
    The experimental phase validated this dual strategy on both direct and inverse problems, using parameterized partial differential equation benchmarks representative of degradation phenomena in cultural heritage.
    The results show strong generalizability, robustness, and accuracy, including reliable approximations of solution fields and parameter identification with relative errors below $10^{-2}$.
    Moreover, ROM techniques based on Proper Orthogonal Decomposition (POD) significantly improve computational efficiency, enabling real-time applications without sacrificing predictive performance.

    In the case of direct problems, the framework successfully simulated dynamic scenarios, such as temperature monitoring, by integrating data-driven boundary conditions with physical constraints.
    The adoption of hard constraints and weighted loss functions within PINNs confirmed the ability to balance empirical observations and governing physical laws, even in complex geometries where such integration is notoriously challenging.
    These results highlight the suitability of the proposed framework for realistic cultural heritage scenarios characterized by irregular shapes and heterogeneous materials.

    Another significant contribution of this work, especially given its multidisciplinary scope and broad range for applications, lies in its commitment to open and reproducible research across the full pipeline. This choice fosters transparency, and allows researchers and practitioners to adapt and extend the framework to new cultural heritage assets and application domains.

    Overall, the objectives outlined in the introduction have been achieved:
    \begin{itemize}
        \item The proposed framework introduces a systematic methodology for analyzing and processing 3D models through the 3D Model Module of the acquisition layer, enabling automatic preparation of data for PINNs, FEM simulations, ROM construction, and result visualization via $\textsf{xdmf}$ files.
        \item By exploiting PINNs, the framework effectively addresses both direct and inverse problems, combining physical knowledge with observational data for tasks such as parameter identification and the simulation of degradation phenomena.
        \item The integration of PINNs with ROMs enables the framework to efficiently identify asset-specific parameters and generate fast, reliable simulations during the online phase, supporting informed decision-making in cultural heritage conservation.
    \end{itemize}

    Despite these promising results, several challenges remain.
    The performance of the framework depends on the quality and consistency of input data, including both 3D models and sensor measurements.
    While automated mesh generation from Blender proved effective, the fidelity of simulations is influenced by the resolution and accuracy of geometric and physical inputs.
    Similarly, the performance of PINNs is sensitive to network architecture design and training dynamics, particularly in time-dependent and multi-physics problems.
    Nevertheless, the modular design of the proposed architecture, combined with the use of open-source libraries, allows for flexible experimentation, tuning, and future extensions.

    Future developments will focus on integrating strategies that infer physical dynamics directly from data in the absence of established models, as well as on defining dedicated workflows for different classes of cultural assets and historical buildings.
    Such extensions will further broaden the applicability of the framework, spanning domains from cultural heritage conservation to structural health monitoring and long-term risk assessment.

\section*{Acknowledgments}
The authors \textbf{CV}, \textbf{FP}, \textbf{DC}, and \textbf{GR} acknowledge the support provided by INdAM-GNCS. \textbf{FP} and \textbf{GR} also acknowledge the support of the European Union - NextGenerationEU, in the framework of the iNEST - Interconnected Nord-Est Innovation Ecosystem (iNEST ECS00000043 - CUP G93C22000610007) consortium and its CC5 Young Researchers initiative.

% \begin{footnotesize}
\bibliographystyle{unsrt}
\bibliography{biblio.bib}
% \end{footnotesize}

\end{document}